\documentclass{article}

\usepackage{PRIMEarxiv}

\usepackage[utf8]{inputenc} 
\usepackage[T1]{fontenc}    
\usepackage{url}            
\usepackage{booktabs}       
\usepackage{amsfonts}       
\usepackage{nicefrac}       
\usepackage{microtype}      
\usepackage{lipsum}
\usepackage{fancyhdr}       
\usepackage{graphicx}       
\graphicspath{{media/}}     

\usepackage{xcolor}         
\definecolor{citecolor}{HTML}{0071bc}
\usepackage[pagebackref=true,breaklinks=true,urlcolor=blue,colorlinks,citecolor=citecolor,bookmarks=false]{hyperref}

\usepackage{array}
\usepackage{makecell}     
\usepackage{caption}      
\usepackage{placeins}
\usepackage{adjustbox}

\usepackage{graphicx}
\usepackage{wrapfig}
\usepackage{tocloft}
\usepackage{enumitem}
\usepackage{longtable}
\usepackage{multirow}
\usepackage{nameref}
\usepackage[misc]{ifsym}
\usepackage{subcaption}
\usepackage{color} 
\usepackage{colortbl}
\usepackage[most]{tcolorbox}
\tcbuselibrary{listings,breakable}

\definecolor{drp-blue}{HTML}{1f77b4}
\definecolor{pretty-blue}{RGB}{0, 113, 188}
\definecolor{kaiming-green}{RGB}{57,181,74} 
\definecolor{mypurple}{RGB}{55,0,168} 
\definecolor{icmlblue}{rgb}{0,0.08,0.45} 
\definecolor{mygreen}{HTML}{4FC978}
\definecolor{linecolor1}{RGB}{246, 248, 239}
\definecolor{linecolor2}{RGB}{230, 234, 217}
\definecolor{linecolor3}{RGB}{211, 222, 190}
\definecolor{line-blue}{RGB}{243, 248, 252}
\definecolor{reconcolor}{HTML}{412F8A}
\definecolor{runpei-orange}{HTML}{F35F27}
\definecolor{runpei_blue}{HTML}{14294B}
\definecolor{datacolor}{HTML}{0009BF}
\definecolor{vitcolor}{HTML}{fc8e62}
\definecolor{cvprblue}{rgb}{0.21,0.49,0.74}
\definecolor{myblue}{rgb}{.39,.58,.93}

\definecolor{bg}{RGB}{248,248,248}

\definecolor{rank1}{HTML}{DCEFE2}   
\definecolor{rank2}{HTML}{E3F2E0}
\definecolor{rank3}{HTML}{EAF4DF}
\definecolor{rank4}{HTML}{F0F6DE}
\definecolor{rank5}{HTML}{F5F8DE}

\definecolor{rank6}{HTML}{FAF8DF}
\definecolor{rank7}{HTML}{FCF6E2}   

\definecolor{rank8}{HTML}{FCF1DF}
\definecolor{rank9}{HTML}{FBEBDC}
\definecolor{rank10}{HTML}{FAE4D8}

\definecolor{rank11}{HTML}{F8DDD4}
\definecolor{rank12}{HTML}{F5D3CF}
\definecolor{rank13}{HTML}{F1C8C8}
\definecolor{rank14}{HTML}{EABBBB} 

\definecolor{per}{HTML}{FCE4D3}
\definecolor{rea}{HTML}{FFF3CA}
\definecolor{nav}{HTML}{D9E1F4}
\definecolor{tra}{HTML}{E3F2D9}
\definecolor{con}{HTML}{D2F4F2}

\newcommand{\listcasestudyname}{\normalsize{List of Case Study Figures}}
\newlistof{casestudyfig}{csf}{\listcasestudyname}
\newcommand{\casestudyfigure}[4]{%
  \clearpage
  \begin{figure}[ht]
    \centering
    \refstepcounter{casestudyfig}%
    \addcontentsline{csf}{casestudyfig}{\protect\numberline{\thecasestudyfig}#2}%
    \includegraphics[width=0.85\textwidth]{#1}
    \caption{#3}
    \label{#4}
    \hyperlink{listofcasestudyfig}{Back to List of figures}
\end{figure}}
  
\title{ArchSIBench: Benchmarking the Architectural Spatial Intelligence of Vision-Language Models}

\author{%
  Qirui Shen\textsuperscript{1}\thanks{Equal contribution.} \quad
  Wenda Wang\textsuperscript{1}\footnotemark[1] \quad
  Jiachen Lu\textsuperscript{1} \quad
  Zilong Huang\textsuperscript{1}\\
  \textbf{Jin Bai}\textsuperscript{1} \quad
  \textbf{Lei He}\textsuperscript{1} \quad
  \textbf{Hongxuan Chen}\textsuperscript{1} \quad
  \textbf{Weixin Huang}\textsuperscript{1}\thanks{Corresponding author.}
  \\
  \textsuperscript{1}School of Architecture, Tsinghua University\\
  \texttt{\{shenqr22, wwd23, lu-jc21, huangzl22,}\\
  \texttt{bai-j24, helei23, hongxuan23\}@mails.tsinghua.edu.cn}\\
  \texttt{\{huangwx\}@tsinghua.edu.cn}\\
}

\begin{document}
\maketitle

\begin{abstract}
  Architectural spatial intelligence, the ability to recognize and infer architectural space, is fundamental to tasks such as robot navigation, embodied interaction, and 3D scene understanding and generation. Although extensive research has evaluated the basic spatial skills of Vision-Language Models (VLMs) such as relative orientation, distance comparison, and object counting, these tasks cover only the most elementary levels of spatial cognition and largely overlook higher-level cognition of architectural space, including layout understanding, circulation patterns, and functional zoning. In this work, we present {\bf ArchSIBench}, a {\bf Bench}mark for {\bf Arch}itectural {\bf S}patial {\bf I}ntelligence based on the perspectives from architecture, cognitive science, and psychology. ArchSIBench covers five core dimensions: perception, reasoning, navigation, transformation, and configuration, comprising 17 fine-grained subtasks. Through careful manual annotation by experts with architectural backgrounds, we construct 3,000 question-answer pairs to enable comprehensive evaluation of architectural spatial intelligence. Based on ArchSIBench, we evaluate various VLMs and find that the architectural spatial intelligence of most models shows significant differences from human baselines; additionally, models exhibit substantial variability across capability dimensions. Some state-of-the-art models can approach the level of human evaluators without architectural training. However, a clear gap remains compared to human evaluators with architectural training, particularly in spatial transformation and configuration reasoning. We believe that ArchSIBench will provide important insights and systematic resources for measuring and advancing the architectural spatial intelligence of VLMs. The dataset and code are available at \url{https://huggingface.co/datasets/ArchSIBench/ArchSIBench}.
\end{abstract}


\begin{figure}
  \centering
  \includegraphics[width=1\linewidth]{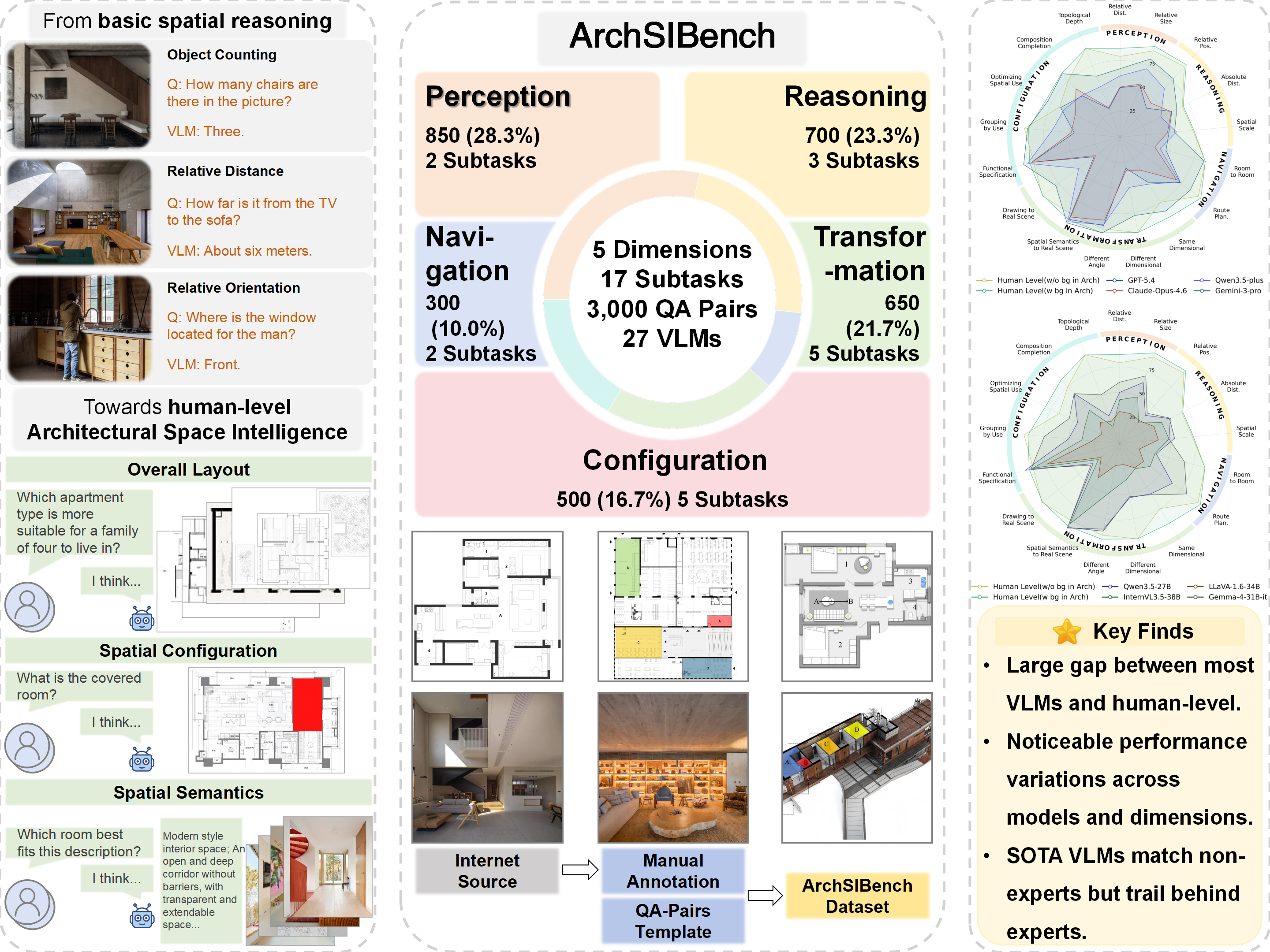}
  \caption{Overview of ArchSIBench.}
  \label{fig:Overview_of_ArchSIBench}
  \vspace{-1em}
\end{figure}

\section{Introduction}
\label{sec:intro}

Architectural spatial intelligence, the ability to recognize and infer the scale, layout, and configuration of architectural space, is a core component of human spatial intelligence. Unlike typical spaces or isolated indoor scenes, architectural space is inherently organized around human use: it constrains and guides people's movement and interaction through explicit geometric features (such as shape and scale) and implicit spatial relationships (such as circulation patterns and functional zoning). This human-centered spatial organization introduces latent structural and functional constraints that cannot be directly inferred from local geometric cues. Therefore, architectural spatial cognition is fundamentally more challenging than general spatial cognition. Across diverse architectural environments, ranging from historical monuments such as the Parthenon to modern buildings such as the Villa Savoye, as well as everyday residential and office spaces, humans can infer spatial structure, estimate scale, and form an understanding of overall layout from visual observation and cognitive experience~\cite{gardner2011frames,fox2010spatial,montello2014spatial,harvey2010space}, thereby supporting complex behaviors including navigation, interaction, spatial understanding, and design~\cite{newcombe2004spatial,meneghetti2022individual,newcombe2018three,tversky1999three,berkowitz2021spatial,montello2013functions,sutton2007spatial}. Such capabilities are equally central to tasks such as indoor navigation~\cite{yue2026spatial}, embodied intelligence~\cite{chen2025exploring}, and 3D scene understanding and generation~\cite{fu2024scene,ling2025scenethesis,yin2025floorplan,fang2025spatialgen,mao2025spatiallm}, which critically rely on architectural spatial intelligence. Despite the rapid progress of Vision-Language Models (VLMs)~\cite{bordes2024introduction,li2025benchmark} in these domains, it remains unclear whether they possess architectural spatial intelligence comparable to humans, or more stringently, to professional architects.

Recently, significant progress has been made in benchmarking the spatial intelligence of VLMs~\cite{zheng2025multimodal,liu2025spatial}. Several works primarily focus on basic spatial skills, including relative orientation, distance comparison, and object counting. While these tasks provide an important foundation for evaluating the basic spatial understanding of VLMs, they largely capture only elementary levels of spatial cognition. Spatial cognition in architectural space, especially from the perspective of architects, extends far beyond object-level and isolated-room-level relationships to include overall structure, layout organization, and functional configuration of the space. Architects can judge how the architectural space is divided and connected, how spatial scale affects behavior, how layout supports functional use~\cite{ching2023architecture,hillier2007space,hillier1989social}, and how observers transform spatial representations between different reference frames~\cite{cooper1990mental,bertoline1990visualization,sutton2007measuring}. These abilities, together with fundamental spatial cognitive abilities, form a broader notion of architectural spatial intelligence that is crucial in architectural design, spatial planning, and environmental cognition, yet remains largely absent from existing benchmarks for evaluating such capabilities of VLMs.

Motivated by the above considerations, we present ArchSIBench, a benchmark for architectural spatial intelligence grounded in architecture, cognitive science, and psychology. ArchSIBench models architectural spatial intelligence as a multi-level cognitive framework and systematically evaluates VLMs across five core dimensions: {\bf perception}, {\bf reasoning}, {\bf navigation}, {\bf transformation}, and {\bf configuration}. These dimensions are further decomposed into 17 fine-grained subtasks, covering diverse scenarios ranging from relative orientation estimation and distance measurement to 2D/3D conversion, multi-perspective spatial reasoning, spatial composition understanding, and functional analysis. Through careful manual annotation by experts with architectural backgrounds, we construct 3,000 high-quality question-answer pairs, sourced from architectural technical drawings (e.g., floor plans and sections), 3D representations (e.g., axonometric drawings and renderings), and real-scene images. We further establish dual human baselines comprising participants with and without architectural training, enabling a fine-grained comparison between VLMs and humans.

We evaluate \textbf{27} VLMs~\cite{hurst2024gpt,singh2025openai,anthropic_claude_opus4_5,anthropic_claude_opus4_6,qwen35blog,bai2025qwen3,google_gemini,wang2025internvl3,liu2024llavanext,google_gemma} on ArchSIBench and find a significant gap between their architectural spatial intelligence and human performance. While some of the most advanced models, such as Gemini-3-Pro~\cite{google_gemini} and Qwen3.5-Plus~\cite{qwen35blog}, approach the performance of human evaluators without architectural education backgrounds, a clear gap remains between them and human evaluators with architectural education backgrounds. We hope ArchSIBench can serve as a new benchmark for advancing research in this field by revealing the limitations and potential of VLMs in architectural spatial intelligence, and facilitating future progress in related work.

In summary, our main contributions are as follows: 

\begin{itemize}
\item We propose a multidisciplinary, professionally grounded taxonomy of architectural spatial intelligence, comprising five core dimensions: perception, reasoning, navigation, transformation, and configuration, with 17 fine-grained subtasks. 
\item We present {\bf ArchSIBench}, a carefully curated benchmark comprising 3,000 samples manually annotated by experts with architectural backgrounds, systematically spanning all dimensions and subtasks for evaluating VLMs on architectural spatial intelligence.
\item We evaluate 27 VLMs and establish fine-grained human baselines that distinguish trained architects from non-expert humans. The results reveal a significant gap between current VLMs and human performance, offering actionable insights for future model development.
\end{itemize}

\section{Related Work}

{\bf Taxonomy of Spatial Cognition:} Spatial cognition has long been a core issue in cognitive science, environmental psychology, and architecture~\cite{vasilyeva2012development}, with complementary taxonomies proposed from different disciplinary perspectives. For example, Newcombe et al.~\cite{newcombe2018three} suggest that spatial cognition is not a unified ability, but is composed of three systems with different evolutionary origins and neural foundations: navigation, object representation and transformation, and spatializing (as a symbolic tool). In addition, Newcombe proposes a classification method for spatial cognition~\cite{newcombe2004spatial}, which divides spatial cognition into ten aspects such as navigation-relevant cognition, allocentric frameworks, and inertial navigation; Tversky et al.~\cite{tversky1999three} argue that human spatial cognition consists of three types of schematic psychological representations: the space of navigation, the space around the body, and the space of the body. Different spaces adopt different reference frames and organizational principles depending on action demands. Research in environmental psychology and architecture extends the concept of spatial cognition from object-level manipulation to environment-level understanding. Some scholars believe that humans construct ``cognitive maps'' to represent space, supporting navigation and broader spatial reasoning~\cite{newcombe2004spatial,tolman1948cognitive,epstein2017cognitive}. This line of work highlights that spatial cognition extends from local geometric relations to global layout inference. In urban research, Kevin Lynch's seminal work \emph{The Image of the City} introduces five elements of cognitive maps: paths, edges, districts, nodes, and landmarks, thus constructing a structured way for human spatial cognition to map to larger built environments~\cite{lynch1964image}. In architecture, spatial cognition goes beyond object positioning and counting, encompassing holistic aspects such as configurational understanding. Theories such as Space Syntax~\cite{hillier2007space,hillier1989social,hillier1976space} emphasize that spatial cognition involves reasoning about integration and connectivity, which govern the interrelationships of space such as accessibility and visibility. This view underscores that spatial cognition requires an understanding of implicit structures, including spatial hierarchy, functional organization, and inter-space relationships. Despite substantial progress from diverse disciplinary perspectives in characterizing fundamental spatial abilities (e.g., distance, direction, and shape), aspects central to architectural space, such as layout understanding, circulation patterns, and functional zoning, remain comparatively underexplored in existing formulations of spatial intelligence.

{\bf Spatial Benchmarks for VLMs:} With the development of VLMs and growing demand for embodied intelligence and 3D scene generation, prior works have begun to systematically evaluate the spatial intelligence of VLMs. Existing benchmarks cover a range of tasks, including relative orientation, distance estimation, object counting, and path planning~\cite{szymanska2024space3d,majumdar2024openeqa,zhang2025open3d,du2024embspatial,azuma2022scanqa,ma2022sqa3d,yang2025thinking,ma20253dsrbench}. These works operationalize spatial cognition into standardized reasoning tasks, enabling standardized evaluation across models. For example, ScanQA~\cite{azuma2022scanqa} and SQA3D~\cite{ma2022sqa3d} introduce large-scale question answering datasets grounded in 3D indoor scenes, focusing on object attributes, spatial relations, and commonsense reasoning in reconstructed environments; VSI-Bench~\cite{yang2025thinking} provides a comprehensive evaluation for visual-spatial intelligence in dynamic 3D environments using egocentric video; 3DSRBench~\cite{ma20253dsrbench} evaluates core 3D spatial inference capabilities such as height, position, direction, and multi-target inference, and further examines robustness under uncommon camera viewpoints. Collectively, these studies have substantially advanced the development of spatial intelligence in VLMs. However, most of them adopt an object-centric formulation, assessing spatial intelligence primarily through object properties and inter-object relations. While this paradigm is effective for evaluating foundational geometric reasoning, it is less suited to capturing models' ability to understand the overall spatial structure. For example, existing benchmarks rarely involve  tasks such as judging the functional properties of space, understanding the correspondence between spatial combinations and usage functions, and spatial logical reasoning based on adjacency and complementary relationships. Yet such capabilities may be central to architectural spatial intelligence and constitute important prerequisites for architectural design and generation tasks.

{\bf Architectural Spatial Intelligence Benchmarks: }Research in architecture has long focused on how humans perceive, interpret, and use space. Architectural analysis and design education routinely relies on spatial cognition abilities such as layout recognition, scale judgment, accessibility analysis, and reasoning about functional organization. Yet despite the accumulation of rich spatial cognitive theories and evaluation methods in architecture, these advances have seen limited incorporation into benchmarks for VLMs. In recent years, several works have attempted to evaluate models using architecture-related tasks. Blueprint-Bench~\cite{petersson2025blueprint} defines the task of converting indoor photos of apartments into 2D floor plans with semantic information, and evaluates large language models, image generation models, and agent systems on a dataset containing photos and floor plans of 50 apartments; WAFFLE~\cite{ganon2025waffle} collects nearly 20,000 floor plans and associated metadata to evaluate models on tasks such as building type understanding, open-vocabulary floor plan segmentation, text-conditioned floor plan generation and structural-conditioned floor plan generation; AECV-Bench~\cite{kondratenko2026aecv} focuses on evaluating the intelligence level of multimodal models on AEC (Architectural Engineering and Construction) drawings. The benchmark includes 120 high-quality floor plans and 192 manually annotated question-answer pairs, covering tasks such as object counting (doors, windows, bedrooms, toilets), text extraction (OCR), instance counting, spatial reasoning, and comparative reasoning. These works are mainly targeted at specific vertical items and are task-oriented rather than structured, capability-oriented. Such designs are effective for measuring practical utility in particular applications, but remain limited by the absence of structured modeling of spatial intelligence itself. In contrast, ArchSIBench shifts the focus from task-specific performance to capability structure. By constructing an evaluation framework with explicit cognitive hierarchy, ArchSIBench aims to provide a unified and systematic evaluation of the architectural spatial intelligence of VLMs.

\section{ArchSIBench}

\subsection{Overview}

We present {\bf ArchSIBench}, a comprehensive benchmark for systematically evaluating the architectural spatial intelligence of VLMs. Rather than continuing to expand task diversity or dataset scale used in previous spatial reasoning benchmarks, we focus on constructing an evaluation framework with explicit cognitive hierarchies and adopt a capability-oriented assessment strategy. ArchSIBench includes 3,000 high-quality question-answer pairs based on architectural technical drawings (e.g., floor plans and sections), 3D representations (e.g., axonometric drawings and renderings), and real-scene images. All visual data are collected from open Internet sources and manually reviewed. All images and question-answer pairs are selected, processed, and reviewed by senior undergraduate students majoring in architecture. Through this process, we ensure sufficient dataset scale, broad thematic coverage, and high data quality.

\subsection{Task Set}
\label{ArchSIBench-Task-Set}

\begin{figure}[t]
    \centering
    
    \begin{minipage}[t]{0.66\linewidth}
        \centering
        \includegraphics[width=\linewidth]{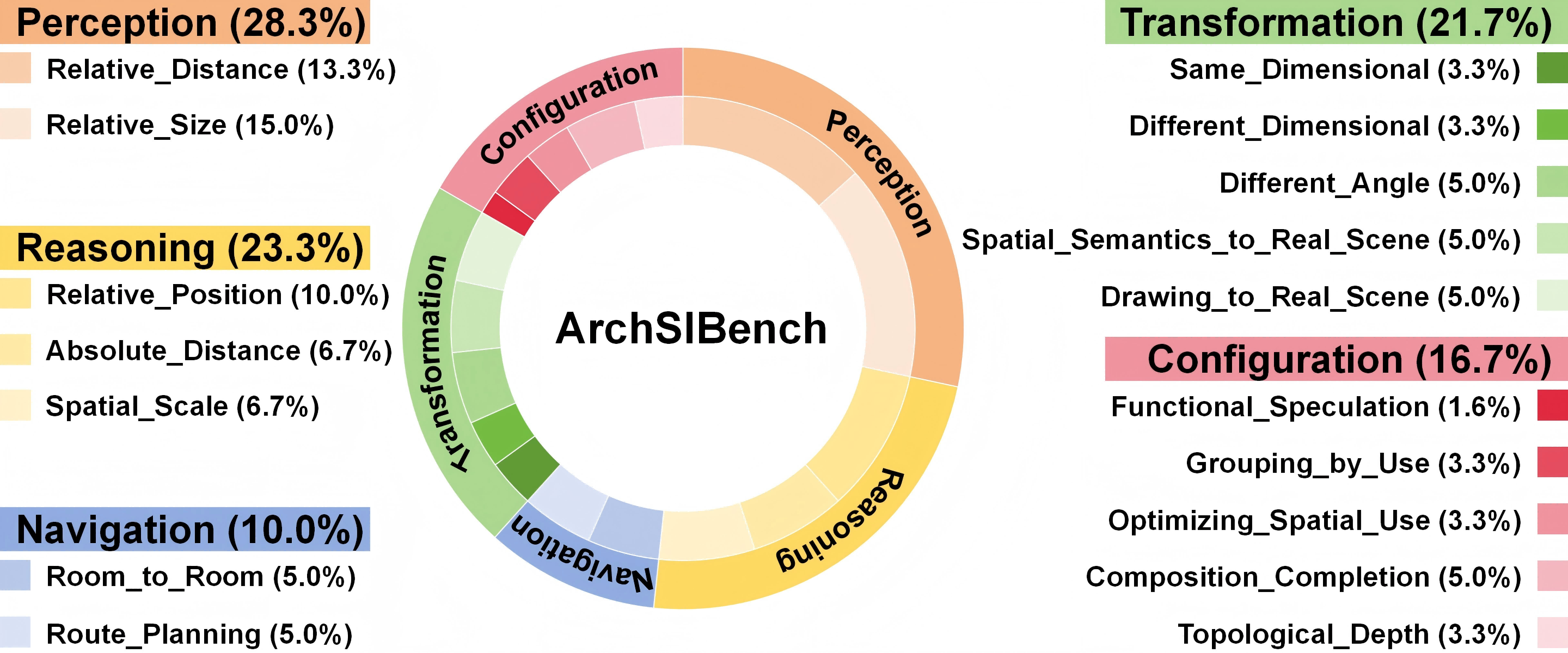}
        \caption{Distribution of data in ArchSIBench.}
        \label{fig:QA_Pairs_num}
    \end{minipage}
    \hfill
    \begin{minipage}[t]{0.33\linewidth}
        \centering
        \includegraphics[width=\linewidth]{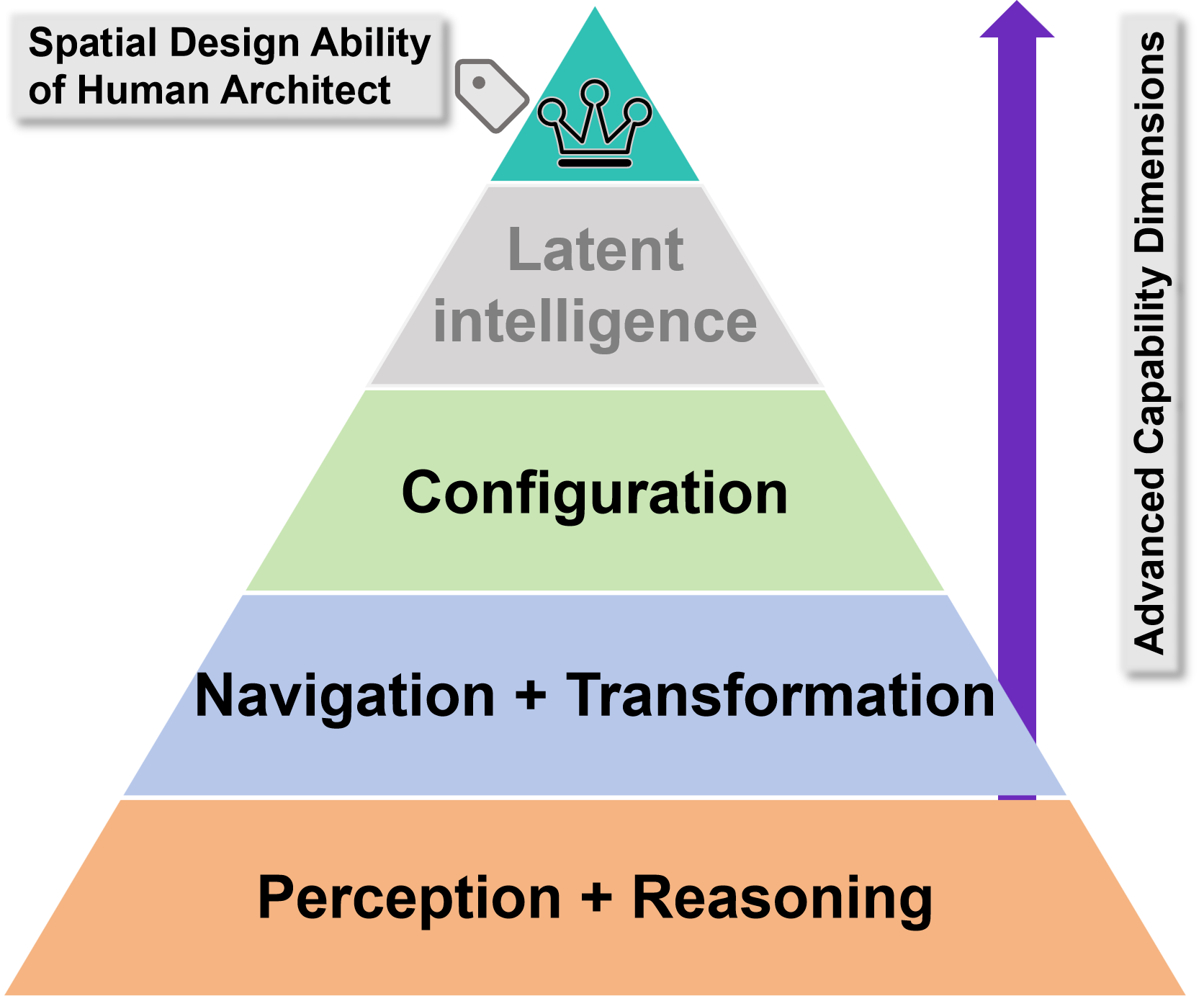}
        \caption{Pyramid-like structure of architectural spatial intelligence.}
        \label{fig:Pyramid}
    \end{minipage}
    \vspace{-1em}
\end{figure}

We organize ArchSIBench into five core dimensions: {\bf perception}, {\bf reasoning}, {\bf navigation}, {\bf transformation}, and {\bf configuration}. These dimensions are further decomposed into 17 fine-grained subtasks. An overview of ArchSIBench is shown in Figure~\ref{fig:Overview_of_ArchSIBench}, with the distribution of data shown in Figure~\ref{fig:QA_Pairs_num}. Guided by both interdisciplinary theory and practical experience with architectural tasks, we organize these dimensions into a pyramid-like structure, as shown in Figure~\ref{fig:Pyramid}, reflecting the hierarchical division and ability advancement from basic to higher-order and more specialized spatial intelligence. Notably, in ArchSIBench we focus on evaluating capabilities in the lower three levels of this pyramid. We consider that these levels are prerequisite abilities for architectural design and generation. Importantly, possessing these capabilities does not imply that VLMs can perform architectural design in the manner of human architects. Substantial latent dimensions remain between the configuration dimension and the final generation dimension, which require further investigation from architecture and cognitive science. In this section, we first provide a conceptual overview of the five core dimensions; detailed task definitions are shown in Appendix~\ref{sec:Detailed-Task-Design}.

{\bf Perception} emphasizes the initial understanding of space through intuitive spatial awareness, including basic spatial attributes such as the position of objects relative to the observer and relative positions between objects~\cite{vasilyeva2012development,tommasi2012psychology}, and approximate judgments of spatial scale~\cite{henry1993spatial}. Specifically, human spatial perception can be encoded through two reference frames: {\bf egocentric} and {\bf allocentric}~\cite{vasilyeva2012development,tommasi2012psychology}. This classification is also reflected in ArchSIBench: in some questions, we require models to perform viewpoint transformations, rather than relying solely on egocentric perception.

{\bf Reasoning} emphasizes the ability to infer spatial relationships, such as distance and relative orientation between objects, by integrating auxiliary cues including object size and position, thereby enabling deeper cognition of space~\cite{tverksy2018levels,freksa2005using}. In addition to integrating other objects to assist in judgment, humans also reason about space by relying on embodied references~\cite{tversky1999three}. It is easy for people to judge whether a certain bed is too small for their body or a space is too crowded for hosting parties. Similarly, in architectural design, ensuring that all aspects of the space conform to human scale is an essential consideration for comfort and usability. Accordingly, ArchSIBench includes tasks involving embodied spatial perception and human-scale reasoning as part of spatial understanding. We consider such abilities to be crucial for future tasks such as embodied intelligence and 3D scene generation suitable for human habitation.

{\bf Navigation} emphasizes the ability to identify feasible paths in space. It is among the most fundamental forms of spatial cognition shared by humans and many animals~\cite{newcombe2004spatial,newcombe2018three,tversky1999three,tommasi2012psychology,werner1997spatial}. Although navigation involves complex mechanisms and associated skills, at its core it concerns moving from one location to another~\cite{tverksy2018levels,chan2012objects}. In architectural space, target locations are often separated by structural elements such as walls, doors, and corridors, making direct straight-line movement infeasible. Therefore, in ArchSIBench, we particularly emphasize the ability to identify architectural structural elements in order to bypass obstacles and find practically feasible paths.

{\bf Transformation} emphasizes the ability to mentally transform perspectives and spatial representations across different modalities and viewpoints, including plan-section transformations, mappings between floor plans and real-scene images, and spatial imagination grounded in text or images. In cognitive science, transformation abilities are commonly reflected in capacities such as mental rotation, mental folding, and object manipulation~\cite{newcombe2018three,zacks2000mental,zacks2002parametric}. The tasks in architecture, such as understanding the correlation across representational views (e.g., from floor plans to sections) and across multimodal representations (e.g., from design intent to sketches or models), can also be viewed as advanced transformation tasks. Therefore, transformation is a core component of architectural spatial intelligence and plays an important role in architecture and related disciplines~\cite{sutton2007spatial}.

{\bf Configuration} emphasizes the cognitive ability to understand the global organization of architectural space, including reasoning about spatial attributes, understanding composition-function correspondences, and interpreting space through adjacency and complementary relationships. We consider the ability to understand spatial configuration as a more advanced form of spatial cognition, which also encompasses potential prerequisites for architectural design. The physical form of space comprises both shape and spatial configuration. Shape refers to the external geometric features, while spatial configuration refers to the relationships among internal elements~\cite{hasgul2015space}. In architecture, spatial configuration is particularly important because it can influence patterns of human behavior, social interaction, and collective activity~\cite{hillier2007space,hillier1989social,zerouati2020evaluating}. Accordingly, architecture has established tools such as {\bf Space Syntax}~\cite{hillier2007space,hillier1989social,hillier1976space} to analyze space in terms of spatial depth and connectivity patterns. In ArchSIBench, we include tasks designed to probe whether models can reason about space from a higher level and construct a more holistic understanding of overall spatial organization.

\subsection{Construction of ArchSIBench}
\label{sec:construction_of_datatset}

\begin{figure}
  \centering
  \includegraphics[width=1\linewidth]{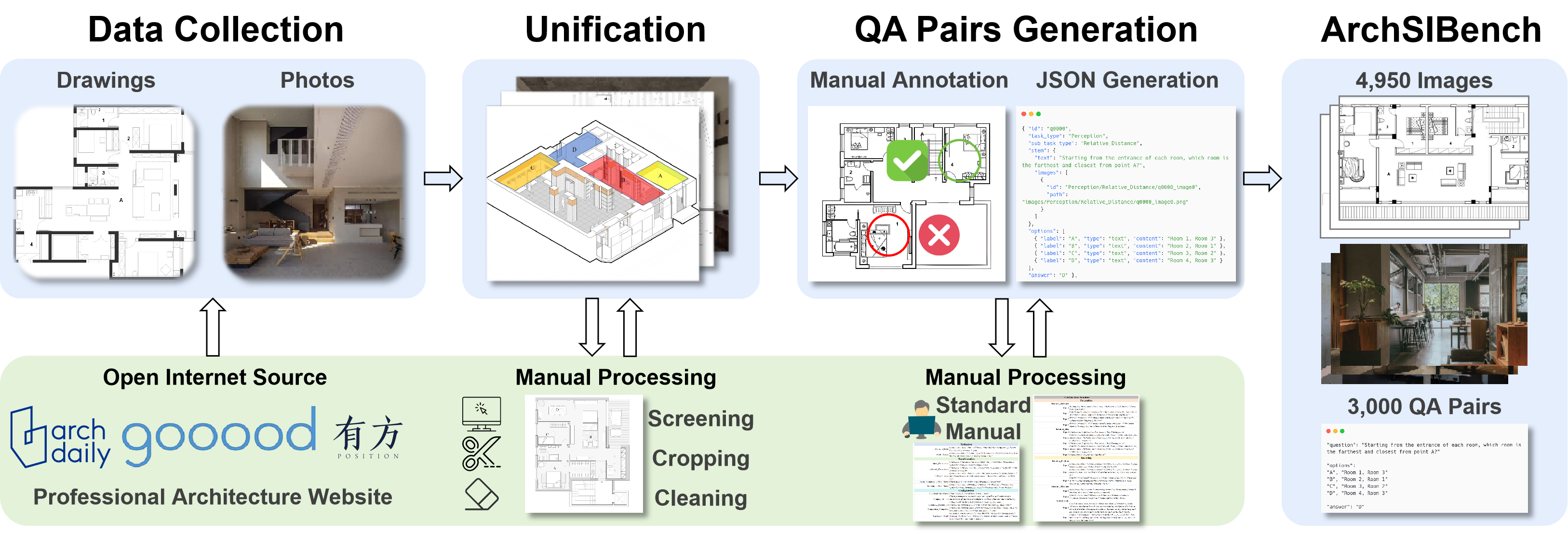}
  \caption{Dataset construction process.}
  \label{fig:Dataset_Construction}
  \vspace{-1em}
\end{figure}

{\bf Data Collection and Unification.} We collect architectural technical drawings (e.g., floor plans and sections), 3D representations (e.g., axonometric drawings and renderings), and real-scene images from open Internet sources. Specifically, we focus on collecting data from professional websites in the field of architecture such as \emph{Archdaily}~\cite{archdaily}, \emph{Goood}~\cite{gooood}, and \emph{Archiposition}~\cite{archiposition}, thereby ensuring that clear architectural semantic information is presented in the images (e.g., standard CAD legends, detailed interior perspectives). The images in the dataset cover diverse scenarios such as residential spaces, office spaces, and public spaces. For the initial images collected, we perform filtering and cleaning work. For example, for captions that conflict with the answer options, we obscure or eliminate them to avoid ambiguity. For questions related to embodied scale perception, we deliberately select images without human presence to avoid VLMs directly deriving answers based on human figures in the images, thereby forcing VLMs to engage in embodied imagination. For explicit visual cues that may leak answers (e.g., scales, size information, room labels), we eliminate them to minimize the possibility of VLMs obtaining answers directly through textual information rather than spatial cognition. Detailed examples are provided in Appendix~\ref{sec:Case-Study}.

{\bf Question-Answer Pairs Generation.} We recruit 10 senior undergraduate students majoring in architecture as annotators to construct question-answer pairs. All question-answer pairs in ArchSIBench are selected, processed, and reviewed by these annotators. For the five core dimensions and 17 subtasks, we develop a comprehensive instruction manual consisting of 28 exemplar templates covering all subtasks. We provide strict question templates and image annotation styles for each subtask. We also provide pre-job training for human volunteers to guide annotators in the construction process. All questions are multiple-choice and contain between 2 and 4 options with only 1 correct option, enabling standardized and consistent evaluation across models.

{\bf Human Quality Review.} Although we have adopted a fully manually annotated dataset construction method, and provided carefully designed guidance documents, the dataset may still contain ambiguities, noise, or errors due to limitations in data sources, annotator oversight, or inherent cognitive biases. To mitigate such issues, we implement a multi-stage manual verification protocol throughout the benchmark construction process. We divide the process into three stages, corresponding to 20\%, 50\%, and 100\% completion of the full dataset. In the first two stages, we focus on reviewing the outputs of each human annotator, including verification and correction of image quality, question and option content, and answer correctness and plausibility. In addition, during human baseline evaluation, participants are asked to report any items they consider problematic. These reported cases are further inspected, and when necessary, revised or re-annotated, with additional testing conducted to ensure consistency and reliability.

\section{Experiments on ArchSIBench}

\subsection{Evaluation Setup}
\label{sec:eval_setup}

{\bf Benchmark Models.} We conduct a comprehensive evaluation of 27 VLMs, covering diverse model families, parameter scales, and training methods. For proprietary models, we consider GPT-4o series~\cite{hurst2024gpt}, GPT-5 series~\cite{singh2025openai}, Claude-Opus-4 series~\cite{anthropic_claude_opus4_5,anthropic_claude_opus4_6}, Qwen3.5 series~\cite{qwen35blog}, Qwen3-VL series~\cite{bai2025qwen3}, and Gemini-3 series~\cite{google_gemini}. For open-source models, we consider the Qwen3.5 series~\cite{qwen35blog}, InternVL3.5 series~\cite{wang2025internvl3}, LLaVA-1.6 series~\cite{liu2024llavanext}, and Gemma-4 series~\cite{google_gemma}. All models are evaluated under a zero-shot setting and using a unified prompt. For questions involving multiple images, we adopt a standardized pipeline to merge the images of the question and options into one image, ensuring that the model receives a single image input regardless of the type of question and thereby avoiding performance variations caused by differences in multi-image processing capabilities. In the prompt, we explicitly instruct models to output only a single letter representing an option (such as A, B, C, or D), enabling automated evaluation of responses. For cases where the outputs do not follow the required format, we apply rule-based post-processing to extract the model's answer from the output. For open-source models, we deploy them using vLLM~\cite{kwon2023efficient} on two NVIDIA RTX PRO 6000 GPUs. For proprietary models, we perform inference via API calls.

{\bf Human Level Performance.} To compare the differences in architectural spatial intelligence between VLMs and humans, we establish two human baselines: the human baselines with and without architectural education backgrounds. We recruit 20 senior undergraduate students from science and engineering disciplines, including 10 from architecture-related majors (e.g., architecture, urban planning) and 10 from other majors. We do not adopt the method of extracting a subset from ArchSIBench for human evaluation, as such a method inevitably leads to distortion of the difficulty distribution of the subset due to differences in difficulty within the questions, thereby affecting the representativeness of results. We adopt the {\bf matrix sampling} strategy~\cite{childs2002matrix}. All 3,000 questions are included in human evaluation and randomly divided into 10 groups of 300 questions each, covering all task categories. Each human participant is randomly assigned one subset, ensuring that every question is answered exactly once across the cohort. All participants complete the evaluation through a unified web interface without access to external resources. Question order is randomized, and no strict time limit is imposed, to avoid a decrease in the accuracy of certain questions due to fatigue. The average completion time per participant is between 1 and 2 hours. Further evaluation details are provided in Appendix~\ref{sec:Evaluation-Details}.

\begin{table*}[t!]
\centering
\vspace{-3pt}
\caption{Performance of various models on \textbf{ArchSIBench}. The Rank column, from {\colorbox{rank1}{green}} to {\colorbox{rank14}{red}}, indicates the ranking of model performance from good to poor; the {\colorbox{rank1}{green}} cells in the remaining columns represent the scores of the best performing models in the task.}
\vspace{-5pt}
\setlength{\tabcolsep}{1.0pt}
\resizebox{1.0\linewidth}{!}{
\begin{tabular}{r@{\hspace{7.5pt}}ccccccccccccccccccc}
\toprule
\multirow{5}{*}{\begin{tabular}[l]{@{}l@{}}\textbf{Method}\end{tabular}} &
\multirow{5}{*}{\textbf{Rank}} & \multirow{5}{*}{\textbf{Avg.}} & 
\multicolumn{2}{c}{\multirow{2}{*}{\textbf{Perception}}} & \multicolumn{3}{c}{\multirow{2}{*}{\textbf{Reasoning}}} & \multicolumn{2}{c}{\multirow{2}{*}{\textbf{Navigation}}} & \multicolumn{5}{c}{\multirow{2}{*}{\textbf{Transformation}}} & \multicolumn{5}{c}{\multirow{2}{*}{\textbf{Configuration}}}\\ \\ \cmidrule{4-20}
& & & 
\begin{tabular}[c]{@{}c@{}}Rel. \\ Dist.\end{tabular} &
\begin{tabular}[c]{@{}c@{}}Rel. \\ Size\end{tabular} &
\begin{tabular}[c]{@{}c@{}}Rel. \\ Posi.\end{tabular} &
\begin{tabular}[c]{@{}c@{}}Abs. \\ Dist. \end{tabular} &
\begin{tabular}[c]{@{}c@{}}Spatial \\ Scale \end{tabular} &
\begin{tabular}[c]{@{}c@{}}Room \\ to Room \end{tabular} &
\begin{tabular}[c]{@{}c@{}}Route \\ Plan. \end{tabular} &
\begin{tabular}[c]{@{}c@{}}Same \\ Dimen.\end{tabular} &
\begin{tabular}[c]{@{}c@{}}Diff. \\ Dimen. \end{tabular} &
\begin{tabular}[c]{@{}c@{}}Diff. \\ Angle \end{tabular} &
\begin{tabular}[c]{@{}c@{}}Sem.\\ to Scene \end{tabular} &
\begin{tabular}[c]{@{}c@{}}Draw. \\ to Scene \end{tabular} &
\begin{tabular}[c]{@{}c@{}}Func. \\ Spec. \end{tabular} &
\begin{tabular}[c]{@{}c@{}}Group. \\ by Use \end{tabular} &
\begin{tabular}[c]{@{}c@{}}Optim. \\ Use \end{tabular} &
\begin{tabular}[c]{@{}c@{}}Comp. \\ Comple. \end{tabular} &
\begin{tabular}[c]{@{}c@{}}Topo. \\ Depth \end{tabular} \\
\midrule
\rowcolor{line-blue}\textbf{\emph{Baseline}} & & & & & & & & & & & & & & & & & & & \\
Human Level(w bg in Arch.) & 1 & 89.2 & 87.0 & 95.1 & 96.0 & 69.0 & 72.5 & 86.7 & 99.0 & 92.0 & 95.0 & 95.3 & 95.3 & 94.7 & 92.0 & 90.0 & 81.0 & 84.7 & 93.0 \\
\emph{Best} & - & 92.3 & - & - & - & - & - & - & - & - & - & - & - & - & - & - & - & - & - \\
\emph{Worst} & - & 86.3 & - & - & - & - & - & - & - & - & - & - & - & - & - & - & - & - & - \\
Human Level(w/o bg in Arch.) & 2 & 85.1 & 79.0 & 90.7 & 93.3 & 68.0 & 71.5 & 86.7 & 96.0 & 75.0 & 95.0 & 91.3 & 89.3 & 95.3 & 88.0 & 94.0 & 66.0 & 75.3 & 93.0 \\
\emph{Best} & - & 90.0 & - & - & - & - & - & - & - & - & - & - & - & - & - & - & - & - & - \\
\emph{Worst} & - & 79.0 & - & - & - & - & - & - & - & - & - & - & - & - & - & - & - & - & - \\
\midrule
\rowcolor{line-blue}\textbf{\emph{Proprietary Models}} & & & & & & & & & & & & & & & & & & & \\
GPT-4o & \cellcolor{rank12}12 & 49.1 & 42.0 & 56.4 & 37.7 & 49.5 & 	53.0 & 41.3 & 47.3 & 38.0 & 37.0 & 58.7 & 91.3 & 38.0 & 88.0 & 52.0 & 45.0 & 54.7 & 19.0 \\
GPT-4o-mini & \cellcolor{rank14}14 & 39.2 & 30.0 & 40.9 & 30.0 & 51.0 & 49.5 & 36.0 & 34.0 & 26.0 & 21.0 & 40.7 & 86.7 & 26.7 & 66.0 & 42.0 & 31.0 & 46.7 & 21.0 \\
GPT-5.2 & \cellcolor{rank7}7 & 53.5 & 44.8 & 54.9 & 41.0 & 51.5 & \cellcolor{rank1}57.5 & 40.7 & 62.7 & 47.0 & 39.0 & 92.0 & 96.0 & 42.7 & 92.0 & 43.0 & 55.0 & 55.3 & 24.0 \\
GPT-5.4 & \cellcolor{rank6}6 & 53.8 & 48.5 & 54.0 & 38.3 & 56.5 & 48.0 & 46.7 & 74.0 & 52.0 & 31.0 & 88.0 & 94.7 & 32.0 & 90.0 & 45.0 & 65.0 & 54.7 & 30.0 \\
GPT-5-mini & \cellcolor{rank5}5 & 62.8 & 56.2 & 68.4 & 59.3 & 58.5 & 51.0 & 53.3 & 74.7 & 48.0 & 46.0 & 87.3 & 97.3 & 51.3 & 90.0 & 70.0 & 62.0 & 66.7 & 37.0 \\
GPT-5-nano & \cellcolor{rank9}9 & 51.9 & 44.8 & 58.0 & 50.3 & 50.0 & \cellcolor{rank1}57.5 & 37.3 & 61.3 & 44.0 & 28.0 & 56.7 & 94.0 & 35.3 & 90.0 & 51.0 & 39.0 & 50.0 & 41.0 \\
Claude-Opus-4.5 & \cellcolor{rank11}11 & 49.1 & 43.3 & 49.1 & 39.0 & 59.5 & 49.5 & 31.3 & 60.0 & 33.0 & 26.0 & 66.0 & 91.3 & 32.7 & 90.0 & 52.0 & 50.0 & 61.3 & 24.0 \\
Claude-Opus-4.6 & \cellcolor{rank8}8 & 53.3 & 49.0 & 55.8 & 45.0 & 56.0 & 46.5 & 35.3 & 69.3 & 38.0 & 33.0 & 82.7 & 92.0 & 30.0 & 90.0 & 48.0 & 49.0 & 64.0 & 31.0 \\
Qwen3.5-plus & \cellcolor{rank4}4 & 63.2 & 60.0 & 71.8 & 64.3 & 44.5 & 55.5 & 40.7 & 73.3 & 54.0 & 50.0 & 87.3 & 97.3 & 50.7 & 94.0 & 67.0 & 61.0 & 55.3 & 53.0 \\
Qwen3-VL-plus & \cellcolor{rank10}10 & 49.7 & 47.2 & 56.7 & 36.3 & 58.0 & 53.0 & 27.3 & 67.3 & 35.0 & 35.0 & 44.0 & 94.7 & 31.3 & 86.0 & 57.0 & 46.0 & 50.7 & 29.0 \\
Qwen3-VL-flash & \cellcolor{rank13}13 & 48.4 & 40.8 & 51.6 & 36.0 & 59.0 & 55.5 & 47.3 & 52.0 & 48.0 & 41.0 & 58.7 & 96.0 & 23.3 & 92.0 & 46.0 & 43.0 & 44.7 & 13.0 \\
Gemini-3.1-pro & \cellcolor{rank3}3 & 77.2 & 67.3 & 83.8 & \cellcolor{rank1}77.7 & 58.5 & 55.0 & 84.7 & 86.7 & 77.0 & 75.0 & 94.7 & \cellcolor{rank1}98.7 & 74.7 & 92.0 & 92.0 & 75.0 & 84.0 & 59.0 \\
Gemini-3-pro & \cellcolor{rank1}1 & \cellcolor{rank1}77.9 & 67.0 & 82.0 & 77.3 & \cellcolor{rank1}62.0 & 53.5 & \cellcolor{rank1}85.3 & 86.0 & \cellcolor{rank1}85.0 & \cellcolor{rank1}78.0 & 95.3 & \cellcolor{rank1}98.7 & \cellcolor{rank1}78.7 & \cellcolor{rank1}98.0 & 91.0 & \cellcolor{rank1}79.0 & 84.0 & \cellcolor{rank1}64.0 \\
Gemini-3-flash & \cellcolor{rank2}2 & 77.6 & \cellcolor{rank1}67.5 & \cellcolor{rank1}86.0 & 76.3 & 60.0 & 53.5 & 84.0 & \cellcolor{rank1}89.3 & 82.0 & 72.0 & \cellcolor{rank1}96.0 & 97.3 & 74.7 & 94.0 & \cellcolor{rank1}95.0 & 76.0 & \cellcolor{rank1}84.7 & 56.0 \\
\rowcolor{line-blue}\textbf{\emph{Open-source Models}} & & & & & & & & & & & & & & & & & & & \\
Qwen3.5-27B & \cellcolor{rank3}3 & 55.9 & 50.0 & 64.0 & 37.3 & 54.5 & 54.0 & 37.3 & 64.7 & 58.0 & 54.0 & 85.3 & 97.3 & 32.6 & \cellcolor{rank1}96.0 & 50.0 & 46.0 & \cellcolor{rank1}66.0 & 29.0 \\
Qwen3.5-35B-A3B & \cellcolor{rank5}5 & 52.0 & 43.3 & 57.8 & 36.0 & 45.0 & 51.5 & 41.3 & 65.3 & 53.0 & 49.0 & 70.0 & 96.7 & \cellcolor{rank1}35.3 & \cellcolor{rank1}96.0 & 57.0 & 44.0 & 57.3 & 26.0 \\
Qwen3.5-122B-A10B & \cellcolor{rank4}4 & 53.6 & 46.8 & 63.3 & 35.6 & 50.0 & 52.5 & 38.0 & 71.3 & 56.0 & 53.0 & 62.7 & \cellcolor{rank1}98.0 & 28.7 & \cellcolor{rank1}96.0 & 54.0 & 54.0 & 48.0 & 39.0 \\
Qwen3.5-397B-A17B & \cellcolor{rank2}2 & 56.7 & 56.3 & 67.1 & 38.3 & 48.0 & \cellcolor{rank1}56.5 & 46.0 & 71.3 & 52.0 & 49.0 & 69.3 & 97.3 & \cellcolor{rank1}35.3 & \cellcolor{rank1}96.0 & 54.0 & 49.0 & 51.3 & 42.0 \\
InternVL3.5-14B & \cellcolor{rank7}7 & 46.3 & 44.8 & 51.1 & 37.0 & \cellcolor{rank1}60.0 & 49.0 & 39.3 & 44.7 & 37.0 & 47.0 & 44.7 & 91.3 & 27.3 & 82.0 & 39.0 & 33.0 & 36.0 & 28.0 \\
InternVL3.5-38B & \cellcolor{rank8}8 & 43.0 & 42.8 & 48.9 & 27.3 & 46.0 & 48.5 & 36.0 & 37.3 & 29.0 & 36.0 & 48.0 & 92.7 & 26.0 & 88.0 & 37.0 & 38.0 & 44.0 & 18.0 \\
InternVL3.5-20B-A4B & \cellcolor{rank10}10 & 37.9 & 37.8 & 42.0 & 30.7 & 46.5 & 45.0 & 28.7 & 36.7 & 31.0 & 26.0 & 28.7 & 83.3 & 24.7 & 80.0 & 36.0 & 20.0 & 30.0 & 20.0 \\
InternVL3.5-30B-A3B & \cellcolor{rank8}8 & 43.0 & 41.5 & 56.9 & 32.7 & 46.0 & 46.0 & 28.0 & 40.7 & 32.0 & 38.0 & 52.7 & 85.3 & 22.7 & 86.0 & 30.0 & 28.0 & 32.7 & 23.0 \\
LLaVA-1.6-Vicuna-7B & \cellcolor{rank13}13 & 28.0 & 30.5 & 23.3 & 23.3 & 37.0 & 35.0 & 27.3 & 21.3 & 30.0 & 23.0 & 22.7 & 26.7 & 25.3 & 74.0 & 31.0 & 24.0 & 26.7 & 28.0 \\
LLaVA-1.6-Vicuna-13B & \cellcolor{rank12}12 & 28.5 & 31.5 & 26.4 & 18.0 & 35.5 & 43.0 & 30.7 & 29.3 & 25.0 & 25.0 & 22.7 & 26.7 & 26.0 & 70.0 & 31.0 & 16.0 & 27.3 & 22.0 \\
LLaVA-1.6-34B & \cellcolor{rank11}11 & 30.1 & 31.3 & 31.1 & 29.7 & 37.5 & 38.0 & 20.0 & 29.3 & 28.0 & 21.0 & 22.0 & 39.3 & 22.7 & 74.0 & 27.0 & 20.0 & 28.0 & 23.0 \\
Gemma-4-26B-A4B-it & \cellcolor{rank5}5 & 52.0 & 47.0 & 55.3 & 29.7 & 57.5 & 51.0 & 53.3 & 62.0 & 53.0 & \cellcolor{rank1}58.0 & 61.3 & 96.0 & 31.3 & 78.0 & 56.0 & 47.0 & 55.3 & 25.0 \\
Gemma-4-31B-it & \cellcolor{rank1}1 & \cellcolor{rank1}62.5 & \cellcolor{rank1}60.0 & \cellcolor{rank1}70.9 & \cellcolor{rank1}40.0 & 55.5 & 48.5 & \cellcolor{rank1}66.0 & \cellcolor{rank1}82.0 & \cellcolor{rank1}59.0 & 53.0 & \cellcolor{rank1}87.3 & 96.7 & \cellcolor{rank1}35.3 & 90.0 & \cellcolor{rank1}74.0 & \cellcolor{rank1}55.0 & \cellcolor{rank1}66.0 & \cellcolor{rank1}51.0 \\
\end{tabular}
}
\vspace{-1em}
\label{tab:main_results}
\end{table*}
\subsection{Main Results}

Table~\ref{tab:main_results} shows the overall performance of various VLMs on ArchSIBench. We also present the performance of different VLMs in series in Appendix~\ref{sec:Detailed-Results-of-Different-Series-VLMs}. Qualitative examples are provided in Appendix~\ref{sec:Case-Study}. Our key observations are as follows:

{\bf Human Level Performance.} As expected, both groups of human evaluators achieve high scores on ArchSIBench. The average score for the architecture background group is {\bf 89.2}, and the average score for the non-architecture background group is {\bf 85.1}. These results indicate an appropriate level of difficulty of ArchSIBench. On the one hand, the human average score does not approach a ceiling, indicating that the tasks still remain cognitively challenging. On the other hand, the overall score remains at a high level, indicating that the benchmark is achievable for individuals with normal spatial intelligence, thus avoiding evaluation failure caused by excessive difficulty. We further observe that the architectural background group exhibits substantially lower variance compared to the non-architecture group (2.50 vs. 11.2), indicating that the dataset captures the consistency induced by domain-specific training rather than random response behavior. Based on our experience in architectural education and practice, we suggest that architectural training may provide a shared spatial analysis framework that enables participants to adopt more consistent cognitive strategies in spatial reasoning, thereby reducing performance variance.

\begin{figure}[t]
    \centering
    
    \begin{minipage}[t]{0.49\linewidth}
        \centering
        \includegraphics[width=\linewidth]{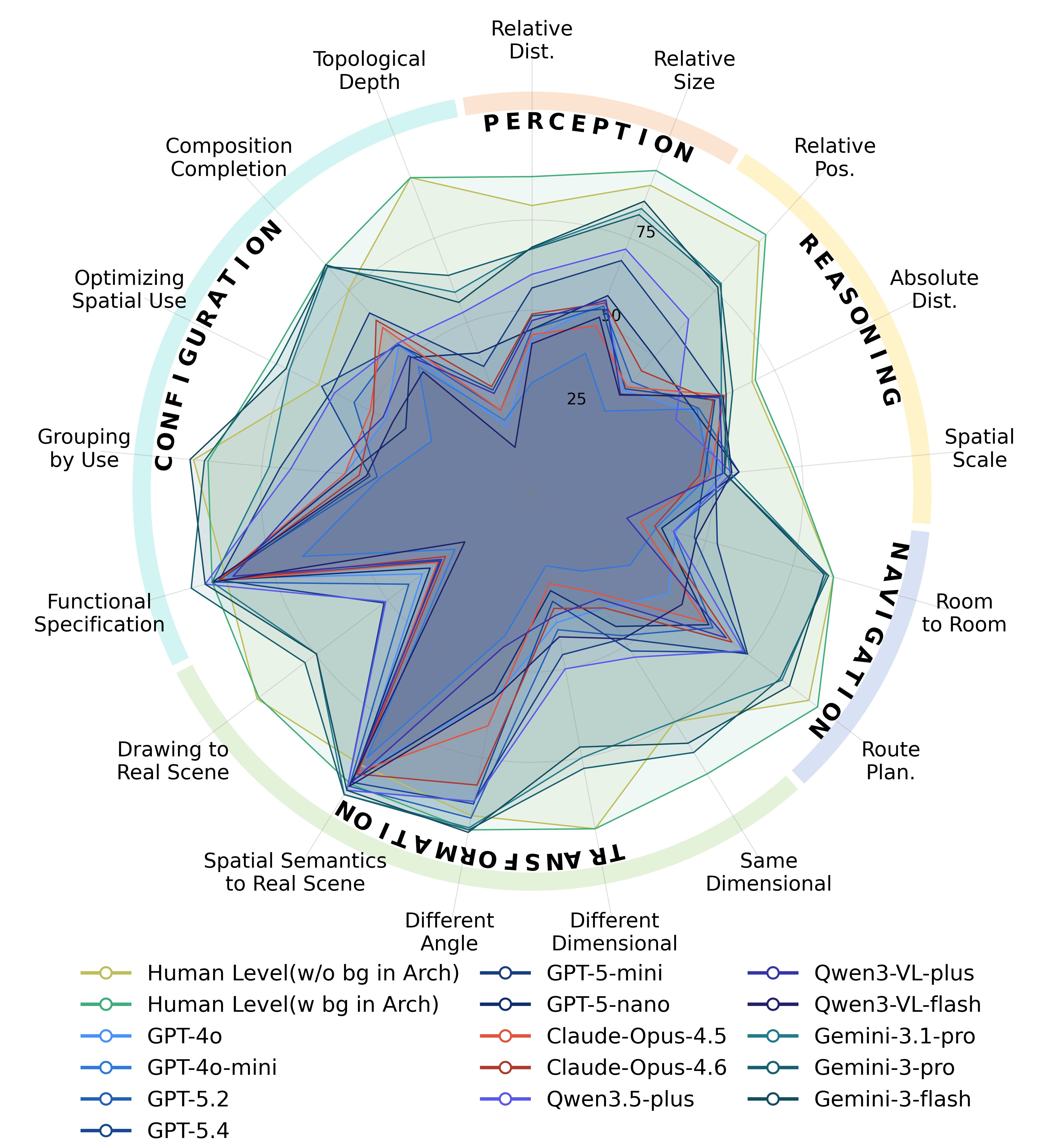}
        \caption{Performance of Proprietary VLMs on ArchSIBench.}
        \label{fig:proprietary VLMs}
    \end{minipage}
    \hfill
    \begin{minipage}[t]{0.49\linewidth}
        \centering
        \includegraphics[width=\linewidth]{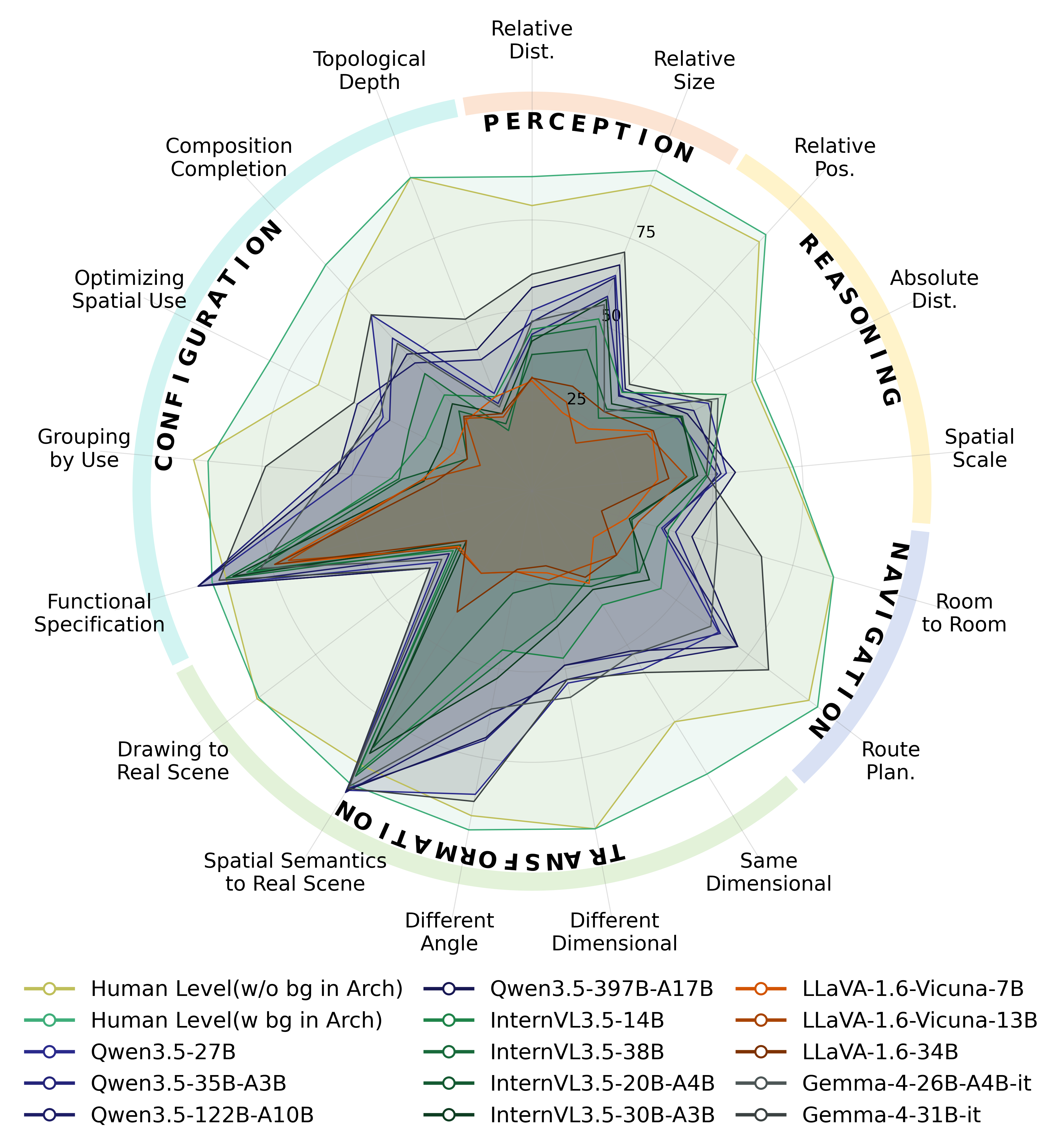}
        \caption{Performance of Open-source VLMs on ArchSIBench.}
        \label{fig:Open-source VLMs}
    \end{minipage}
    \vspace{-1em}
\end{figure}

{\bf Proprietary VLMs.} The overall performance of the proprietary VLMs is shown in Figure~\ref{fig:proprietary VLMs}. While most VLMs still exhibit a substantial performance gap compared to humans, we observe that the leading model family, Gemini-3, achieves highly competitive results. In terms of core dimensions, Gemini-3-Pro performs comparably to the human baseline of the non-architectural background group, and even matches it on transformation and configuration, while maintaining a clear lead over other models. In terms of subtasks, Gemini-3 series models match or exceed both human baselines on 3 of the 17 subtasks (Different Angle, Spatial Semantics to Real Scene, Functional Speculation), and approach the human baseline on 5 of 17 subtasks (Room to Room, Same Dimensional, Grouping by Use, Optimizing Spatial Use, Composition by Completion). Except for Gemini-3 series, we observe that other models exhibit a consistent ability gradient. Notably, this gradient is not smooth: performance across subtasks shows a highly non-uniform, discontinuous pattern, where models achieve near-human performance on certain subtasks while falling substantially behind on others. The majority of proprietary models perform relatively well on subtasks such as Spatial Semantics to Real Scene and Functional Speculation, while exhibiting substantially weaker performance on subtasks such as Different Dimensional, Drawing to Real Scene, and Topological Depth. We find that the subtasks with strong performance in VLMs mostly do not rely on global spatial structure modeling, but are closer to tasks that VLMs are better at, such as image recognition and text-image matching. Subtasks with weaker performance in VLMs often require the construction and maintenance of a cross-perspective, global ``world model'', or strong understanding of architectural elements (such as identifying walls, doors, and corridors).

{\bf Open-source VLMs.} The overall performance of the open-source VLMs is shown in Figure~\ref{fig:Open-source VLMs}. We find that on ArchSIBench, open-source models exhibit performance levels and capability hierarchies that are largely comparable to their proprietary counterparts within similar model tiers. In addition, we observe that increasing model scale does not lead to a significant improvement in performance on tasks in ArchSIBench. For instance, within the Qwen3.5 series, the 27B model achieves the best performance on many tasks, while the 35B and 122B models perform worse overall. Further scaling from 122B to 397B yields only marginal gains. We consider that this aspect reflects that the improvement of architectural spatial intelligence may rely more on specialized spatial representations, structured training data, or explicit geometric priors, rather than simply increasing parameter scale; Moreover, some of the current VLMs adopt a Mixture-of-Experts (MoE) architecture~\cite{jacobs1991adaptive,shazeer2017outrageously,fedus2022review}. In the inference process, a small subset of experts is activated, and these experts are usually naturally differentiated on massive texts (such as code and writing), and spatial related abilities may not be easily differentiated. We hope these findings can provide useful insights for the interpretability and optimization of VLM performance in spatial tasks.

\section{Limitations and Future Work}
\label{sec:limit}

Although ArchSIBench provides a systematic evaluation of architectural spatial intelligence of VLMs from multiple perspectives and levels, several directions remain for future work:

{\bf Improved evaluation dimensions.} As discussed in Sec.~\ref{ArchSIBench-Task-Set}, achieving human-level performance on recent tasks does not necessarily imply comparable capability in architectural design. There remain multiple dimensions between the configuration dimension and the generation dimension. Future work should integrate insights from cognitive science and architecture to define latent dimensions and develop more comprehensive evaluation frameworks.

{\bf Application of synthetic scenarios.} The available high-quality data from open Internet sources is insufficient for large-scale datasets. We consider introducing synthetic scenarios in the future. This aligns with architectural practice, where 3D modeling and rendering are widely used to present design schemes. Future work may incorporate recent advances in scene synthesis~\cite{raistrick2023infinite,raistrick2024infinigen} to expand the availability of high-quality data and enable more comprehensive evaluation.

\section{Conclusion}
\label{sec:conclusion}

We present {\bf ArchSIBench}, a benchmark for architectural spatial intelligence of VLMs based on the perspectives from architecture, cognitive science, and psychology. It consists of five core dimensions: perception, reasoning, navigation, transformation, and configuration, as well as 17 fine-grained subtypes, totaling 3,000 question-answer pairs manually annotated by experts with architectural backgrounds. Empirical evaluations reveal a substantial gap between existing VLMs and human baselines, and the internal differences are significant: VLMs perform better in tasks that do not rely on global spatial structure modeling and are closer to image recognition or image–text matching, but perform poorly in tasks that require building and maintaining a cross-perspective, global ``world model'' or strong understanding of architectural elements. Overall, ArchSIBench provides a useful benchmark for developing VLMs with powerful spatial understanding capabilities. We expect ArchSIBench to stimulate the community to build advancing VLMs toward expert-level spatial intelligence, with implications for architecture, embodied AI, and 3D scene understanding and generation.

\section*{Acknowledgments}
This work is supported by the National Natural Science Foundation of China [grant No. 52178019]. We would also like to thank the Beijing Institute of Architectural Design for providing valuable discussion opportunities and feedback on the evaluation dimensions and tasks of ArchSIBench, as well as all dataset annotators and human evaluators for their outstanding contributions.

\clearpage
\newpage

\bibliographystyle{unsrt}  
\bibliography{references}

@book{gardner2011frames,
  title={Frames of mind: The theory of multiple intelligences},
  author={Gardner, Howard},
  year={2011},
  publisher={Basic books}
}

@article{fox2010spatial,
  title={Spatial Intelligence: New Futures for Architecture},
  author={Fox, William L},
  journal={Places Journal},
  year={2010}
}

@article{montello2014spatial,
  title={Spatial cognition and architectural space: Research perspectives},
  author={Montello, Daniel R},
  journal={Architectural Design},
  volume={84},
  number={5},
  pages={74--79},
  year={2014},
  publisher={Wiley Online Library}
}

@article{harvey2010space,
  title={The space for culture and cognition},
  author={Harvey, Daina Cheyenne},
  journal={Poetics},
  volume={38},
  number={2},
  pages={185--204},
  year={2010},
  publisher={Elsevier}
}

@article{newcombe2004spatial,
  title={Spatial cognition},
  author={Newcombe, Nora S},
  journal={Memory and Cognitive Processes},
  volume={3},
  pages={113--163},
  year={2004},
  publisher={Stevens’ Handbook of Experimental Psychology, John Wiley, New York, ed}
}

@article{meneghetti2022individual,
  title={Individual differences in navigation: an introductory overview},
  author={Meneghetti, Chiara and Miola, Laura and Feraco, Tommaso and Muffato, Veronica and Miola},
  journal={Prime archives in psychology},
  volume={2},
  pages={3},
  year={2022},
  publisher={Vide Leaf}
}

@article{newcombe2018three,
  title={Three kinds of spatial cognition},
  author={Newcombe, Nora S},
  journal={Stevens' handbook of experimental psychology and cognitive neuroscience},
  volume={3},
  pages={1--31},
  year={2018},
  publisher={Wiley Online Library}
}

@article{tversky1999three,
  title={Three spaces of spatial cognition},
  author={Tversky, Barbara and Bauer Morrison, Julie and Franklin, Nancy and Bryant, David J},
  journal={The Professional Geographer},
  volume={51},
  number={4},
  pages={516--524},
  year={1999},
  publisher={Taylor \& Francis}
}

@article{berkowitz2021spatial,
  title={Spatial abilities for architecture: Cross sectional and longitudinal assessment with novel and existing spatial ability tests},
  author={Berkowitz, Michal and Gerber, Andri and Thurn, Christian M and Emo, Beatrix and Hoelscher, Christoph and Stern, Elsbeth},
  journal={Frontiers in psychology},
  volume={11},
  pages={609363},
  year={2021},
  publisher={Frontiers Media SA}
}

@article{montello2013functions,
  title={Functions and applications of spatial cognition.},
  author={Montello, Daniel R and Raubal, Martin},
  journal={Handbook of Spatial Cognition},
  year={2013},
  publisher={American Psychological Association}
}

@article{sutton2007spatial,
  title={Spatial cognition and its implications for design},
  author={Sutton, Ken J and Williams, Anthony P},
  journal={International Association of Societies of Design Research, Hong Kong, China},
  year={2007}
}

@article{yue2026spatial,
  title={Spatial-VLN: Zero-Shot Vision-and-Language Navigation With Explicit Spatial Perception and Exploration},
  author={Yue, Lu and Fan, Yue and Lian, Shiwei and Zhao, Yu and Yu, Jiaxin and Xie, Liang and Zhang, Feitian},
  journal={arXiv preprint arXiv:2601.12766},
  year={2026}
}

@article{chen2025exploring,
  title={Exploring embodied multimodal large models: Development, datasets, and future directions},
  author={Chen, Shoubin and Wu, Zehao and Zhang, Kai and Li, Chunyu and Zhang, Baiyang and Ma, Fei and Yu, Fei Richard and Li, Qingquan},
  journal={Information Fusion},
  volume={122},
  pages={103198},
  year={2025},
  publisher={Elsevier}
}

@article{fu2024scene,
  title={Scene-llm: Extending language model for 3d visual understanding and reasoning},
  author={Fu, Rao and Liu, Jingyu and Chen, Xilun and Nie, Yixin and Xiong, Wenhan},
  journal={arXiv preprint arXiv:2403.11401},
  year={2024}
}

@article{ling2025scenethesis,
  title={Scenethesis: A language and vision agentic framework for 3d scene generation},
  author={Ling, Lu and Lin, Chen-Hsuan and Lin, Tsung-Yi and Ding, Yifan and Zeng, Yu and Sheng, Yichen and Ge, Yunhao and Liu, Ming-Yu and Bera, Aniket and Li, Zhaoshuo},
  journal={arXiv preprint arXiv:2505.02836},
  year={2025}
}

@article{yin2025floorplan,
  title={FloorPlan-DeepSeek (FPDS): A multimodal approach to floorplan generation using vector-based next room prediction},
  author={Yin, Jun and Zeng, Pengyu and Zhong, Jing and Li, Peilin and Zhang, Miao and Luo, Ran and Lu, Shuai},
  journal={arXiv preprint arXiv:2506.21562},
  year={2025}
}

@article{fang2025spatialgen,
  title={Spatialgen: Layout-guided 3d indoor scene generation},
  author={Fang, Chuan and Li, Heng and Liang, Yixun and Zheng, Jia and Mao, Yongsen and Liu, Yuan and Tang, Rui and Zhou, Zihan and Tan, Ping},
  journal={arXiv preprint arXiv:2509.14981},
  volume={3},
  year={2025}
}

@article{mao2025spatiallm,
  title={Spatiallm: Training large language models for structured indoor modeling},
  author={Mao, Yongsen and Zhong, Junhao and Fang, Chuan and Zheng, Jia and Tang, Rui and Zhu, Hao and Tan, Ping and Zhou, Zihan},
  journal={arXiv preprint arXiv:2506.07491},
  year={2025}
}

@article{bordes2024introduction,
  title={An introduction to vision-language modeling},
  author={Bordes, Florian and Pang, Richard Yuanzhe and Ajay, Anurag and Li, Alexander C and Bardes, Adrien and Petryk, Suzanne and Ma{\~n}as, Oscar and Lin, Zhiqiu and Mahmoud, Anas and Jayaraman, Bargav and others},
  journal={arXiv preprint arXiv:2405.17247},
  year={2024}
}

@article{li2025benchmark,
  title={Benchmark evaluations, applications, and challenges of large vision language models: A survey},
  author={Li, Zongxia and Wu, Xiyang and Du, Hongyang and Nghiem, Huy and Shi, Guangyao},
  journal={arXiv preprint arXiv:2501.02189},
  volume={1},
  pages={1},
  year={2025}
}

@article{zheng2025multimodal,
  title={Multimodal spatial reasoning in the large model era: A survey and benchmarks},
  author={Zheng, Xu and Dongfang, Zihao and Jiang, Lutao and Zheng, Boyuan and Guo, Yulong and Zhang, Zhenquan and Albanese, Giuliano and Yang, Runyi and Ma, Mengjiao and Zhang, Zixin and others},
  journal={arXiv preprint arXiv:2510.25760},
  year={2025}
}

@article{liu2025spatial,
  title={Spatial Reasoning in Multimodal Large Language Models: A Survey of Tasks, Benchmarks and Methods},
  author={Liu, Weichen and Xue, Qiyao and Wang, Haoming and Yin, Xiangyu and Yang, Boyuan and Gao, Wei},
  journal={arXiv preprint arXiv:2511.15722},
  year={2025}
}

@book{ching2023architecture,
  title={Architecture: Form, space, and order},
  author={Ching, Francis DK},
  year={2023},
  publisher={John Wiley \& Sons}
}

@book{hillier2007space,
  title={Space is the machine: a configurational theory of architecture},
  author={Hillier, Bill},
  year={2007},
  publisher={Space Syntax}
}

@book{hillier1989social,
  title={The social logic of space},
  author={Hillier, Bill and Hanson, Julienne},
  year={1989},
  publisher={Cambridge university press}
}

@article{cooper1990mental,
  title={Mental representation of three-dimensional objects in visual problem solving and recognition.},
  author={Cooper, Lynn A},
  journal={Journal of Experimental Psychology: Learning, Memory, and Cognition},
  volume={16},
  number={6},
  pages={1097},
  year={1990},
  publisher={American Psychological Association}
}

@article{bertoline1990visualization,
  title={A visualization and orthographic drawing test using the Macintosh computer},
  author={Bertoline, Gary R and Miller, Daniel C},
  journal={Engineering Design Graphics Journal},
  volume={54},
  number={1},
  pages={1--7},
  year={1990}
}

@article{sutton2007measuring,
  title={Measuring 3-D understanding on the Web and in the laboratory},
  author={Sutton, Ken and Heathcote, Andrew and Bore, Miles},
  journal={Behavior Research Methods},
  volume={39},
  number={4},
  pages={926--939},
  year={2007},
  publisher={Springer}
}

@article{hurst2024gpt,
  title={GPT-4o System Card},
  author={Hurst, Aaron and Lerer, Adam and Goucher, Adam P and Perelman, Adam and Ramesh, Aditya and Clark, Aidan and Ostrow, AJ and Welihinda, Akila and Hayes, Alan and Radford, Alec and others},
  journal={arXiv preprint arXiv:2410.21276},
  year={2024}
}

@article{singh2025openai,
  title={Openai gpt-5 system card},
  author={Singh, Aaditya and Fry, Adam and Perelman, Adam and Tart, Adam and Ganesh, Adi and El-Kishky, Ahmed and McLaughlin, Aidan and Low, Aiden and Ostrow, AJ and Ananthram, Akhila and others},
  journal={arXiv preprint arXiv:2601.03267},
  year={2025}
}

@misc{anthropic_claude_opus4_5,
  author       = {{Anthropic}},
  title        = {Introducing Claude Opus 4.5},
  year         = {2025},
  howpublished = {https://www.anthropic.com/news/claude-opus-4-5},
  url          = {https://www.anthropic.com/news/claude-opus-4-5},
  urldate      = {April 2026}
}

@misc{anthropic_claude_opus4_6,
  author       = {{Anthropic}},
  title        = {Introducing Claude Opus 4.6},
  year         = {2026},
  howpublished = {https://www.anthropic.com/news/claude-opus-4-6},
  url          = {https://www.anthropic.com/news/claude-opus-4-6},
  urldate      = {April 2026}
}

@misc{qwen35blog,
    title = {Qwen3.5: Accelerating Productivity with Native Multimodal Agents},
    url = {https://qwen.ai/blog?id=qwen3.5},
    author = {Qwen Team},
    month = {February},
    year = {2026}
}

@article{bai2025qwen3,
  title={Qwen3-vl technical report},
  author={Bai, Shuai and Cai, Yuxuan and Chen, Ruizhe and Chen, Keqin and Chen, Xionghui and Cheng, Zesen and Deng, Lianghao and Ding, Wei and Gao, Chang and Ge, Chunjiang and others},
  journal={arXiv preprint arXiv:2511.21631},
  year={2025}
}

@misc{google_gemini,
  author  = {{Google DeepMind}},
  title   = {Gemini 3: Our most intelligent AI model that brings any idea to life},
  year    = {2026},
  howpublished = {https://deepmind.google/models/gemini/},
  url     = {https://deepmind.google/models/gemini/},
  urldate = {April 2026}
}

@article{wang2025internvl3,
  title={Internvl3. 5: Advancing open-source multimodal models in versatility, reasoning, and efficiency},
  author={Wang, Weiyun and Gao, Zhangwei and Gu, Lixin and Pu, Hengjun and Cui, Long and Wei, Xingguang and Liu, Zhaoyang and Jing, Linglin and Ye, Shenglong and Shao, Jie and others},
  journal={arXiv preprint arXiv:2508.18265},
  year={2025}
}

@misc{liu2024llavanext,
    title={LLaVA-NeXT: Improved reasoning, OCR, and world knowledge},
    url={https://llava-vl.github.io/blog/2024-01-30-llava-next/},
    author={Liu, Haotian and Li, Chunyuan and Li, Yuheng and Li, Bo and Zhang, Yuanhan and Shen, Sheng and Lee, Yong Jae},
    month={January},
    year={2024}
}

@misc{google_gemma,
  author  = {{Google DeepMind}},
  title   = {Gemma: Our most capable open models},
  year    = {2026},
  howpublished = {https://deepmind.google/models/gemma/},
  url     = {https://deepmind.google/models/gemma/},
  urldate = {April 2026}
}

@article{vasilyeva2012development,
  title={Development of spatial cognition},
  author={Vasilyeva, Marina and Lourenco, Stella F},
  journal={Wiley Interdisciplinary Reviews: Cognitive Science},
  volume={3},
  number={3},
  pages={349--362},
  year={2012},
  publisher={Wiley Online Library}
}

@article{tolman1948cognitive,
  title={Cognitive maps in rats and men.},
  author={Tolman, Edward C},
  journal={Psychological review},
  volume={55},
  number={4},
  pages={189},
  year={1948},
  publisher={American Psychological Association}
}

@article{epstein2017cognitive,
  title={The cognitive map in humans: spatial navigation and beyond},
  author={Epstein, Russell A and Patai, Eva Zita and Julian, Joshua B and Spiers, Hugo J},
  journal={Nature neuroscience},
  volume={20},
  number={11},
  pages={1504--1513},
  year={2017},
  publisher={Nature Publishing Group US New York}
}

@book{lynch1964image,
  title={The image of the city},
  author={Lynch, Kevin},
  year={1964},
  publisher={MIT press}
}

@article{hillier1976space,
  title={Space syntax},
  author={Hillier, Bill and Leaman, Adrian and Stansall, Paul and Bedford, Michael},
  journal={Environment and Planning B: Planning and design},
  volume={3},
  number={2},
  pages={147--185},
  year={1976},
  publisher={SAGE Publications Sage UK: London, England}
}

@inproceedings{szymanska2024space3d,
  title={Space3d-bench: Spatial 3d question answering benchmark},
  author={Szyma{\'n}ska, Emilia and Dusmanu, Mihai and Buurlage, Jan-Willem and Rad, Mahdi and Pollefeys, Marc},
  booktitle={European Conference on Computer Vision},
  pages={68--85},
  year={2024},
  organization={Springer}
}

@inproceedings{majumdar2024openeqa,
  title={Openeqa: Embodied question answering in the era of foundation models},
  author={Majumdar, Arjun and Ajay, Anurag and Zhang, Xiaohan and Putta, Pranav and Yenamandra, Sriram and Henaff, Mikael and Silwal, Sneha and Mcvay, Paul and Maksymets, Oleksandr and Arnaud, Sergio and others},
  booktitle={Proceedings of the IEEE/CVF Conference on Computer Vision and Pattern Recognition},
  pages={16488--16498},
  year={2024}
}

@article{zhang2025open3d,
  title={Open3D-VQA: A Benchmark for Comprehensive Spatial Reasoning with Multimodal Large Language Model in Open Space},
  author={Zhang, Weichen and Zhou, Zile and Zeng, Xin and Liu, Xuchen and Fang, Jianjie and Gao, Chen and Li, Yong and Cui, Jinqiang and Chen, Xinlei and Zhang, Xiao-Ping},
  journal={arXiv preprint arXiv:2503.11094},
  year={2025}
}

@inproceedings{du2024embspatial,
  title={Embspatial-bench: Benchmarking spatial understanding for embodied tasks with large vision-language models},
  author={Du, Mengfei and Wu, Binhao and Li, Zejun and Huang, Xuan-Jing and Wei, Zhongyu},
  booktitle={Proceedings of the 62nd Annual Meeting of the Association for Computational Linguistics (Volume 2: Short Papers)},
  pages={346--355},
  year={2024}
}

@inproceedings{azuma2022scanqa,
  title={Scanqa: 3d question answering for spatial scene understanding},
  author={Azuma, Daichi and Miyanishi, Taiki and Kurita, Shuhei and Kawanabe, Motoaki},
  booktitle={Proceedings of the IEEE/CVF Conference on Computer Vision and Pattern Recognition},
  pages={19129--19139},
  year={2022}
}

@article{ma2022sqa3d,
  title={Sqa3d: Situated question answering in 3d scenes},
  author={Ma, Xiaojian and Yong, Silong and Zheng, Zilong and Li, Qing and Liang, Yitao and Zhu, Song-Chun and Huang, Siyuan},
  journal={arXiv preprint arXiv:2210.07474},
  year={2022}
}

@inproceedings{yang2025thinking,
  title={Thinking in space: How multimodal large language models see, remember, and recall spaces},
  author={Yang, Jihan and Yang, Shusheng and Gupta, Anjali W and Han, Rilyn and Fei-Fei, Li and Xie, Saining},
  booktitle={Proceedings of the Computer Vision and Pattern Recognition Conference},
  pages={10632--10643},
  year={2025}
}

@inproceedings{ma20253dsrbench,
  title={3dsrbench: A comprehensive 3d spatial reasoning benchmark},
  author={Ma, Wufei and Chen, Haoyu and Zhang, Guofeng and Chou, Yu-Cheng and Chen, Jieneng and de Melo, Celso and Yuille, Alan},
  booktitle={Proceedings of the IEEE/CVF International Conference on Computer Vision},
  pages={6924--6934},
  year={2025}
}

@article{petersson2025blueprint,
  title={Blueprint-Bench: Comparing spatial intelligence of LLMs, agents and image models},
  author={Petersson, Lukas and Backlund, Axel and Wennst{\"o}m, Axel and Petersson, Hanna and Sharrock, Callum and Dabiri, Arash},
  journal={arXiv preprint arXiv:2509.25229},
  year={2025}
}

@inproceedings{ganon2025waffle,
  title={Waffle: Multimodal floorplan understanding in the wild},
  author={Ganon, Keren and Alper, Morris and Mikulinsky, Rachel and Averbuch-Elor, Hadar},
  booktitle={2025 IEEE/CVF Winter Conference on Applications of Computer Vision (WACV)},
  pages={1488--1497},
  year={2025},
  organization={IEEE}
}

@article{kondratenko2026aecv,
  title={AECV-Bench: Benchmarking Multimodal Models on Architectural and Engineering Drawings Understanding},
  author={Kondratenko, Aleksei and Birhane, Mussie and Hsain, Houssame E and Maciocci, Guido},
  journal={arXiv preprint arXiv:2601.04819},
  year={2026}
}

@article{tommasi2012psychology,
  title={Psychology of spatial cognition},
  author={Tommasi, Luca and Laeng, Bruno},
  journal={Wiley Interdisciplinary Reviews: Cognitive Science},
  volume={3},
  number={6},
  pages={565--580},
  year={2012},
  publisher={Wiley Online Library}
}

@inproceedings{henry1993spatial,
  title={Spatial perception in virtual environments: Evaluating an architectural application},
  author={Henry, Daniel and Furness, Tom},
  booktitle={Proceedings of IEEE Virtual Reality Annual International Symposium},
  pages={33--40},
  year={1993},
  organization={IEEE}
}

@incollection{tverksy2018levels,
  title={Levels and structure of spatial knowledge},
  author={Tverksy, Barbara},
  booktitle={Cognitive mapping},
  pages={24--43},
  year={2018},
  publisher={Routledge}
}

@inproceedings{freksa2005using,
  title={Using orientation information for qualitative spatial reasoning},
  author={Freksa, Christian},
  booktitle={Theories and Methods of Spatio-Temporal Reasoning in Geographic Space: International Conference GIS—From Space to Territory: Theories and Methods of Spatio-Temporal Reasoning Pisa, Italy, September 21--23, 1992 Proceedings},
  pages={162--178},
  year={2005},
  organization={Springer}
}

@incollection{werner1997spatial,
  title={Spatial cognition: The role of landmark, route, and survey knowledge in human and robot navigation1},
  author={Werner, Steffen and Krieg-Br{\"u}ckner, Bernd and Mallot, Hanspeter A and Schweizer, Karin and Freksa, Christian},
  booktitle={Informatik’97 Informatik als Innovationsmotor: 27. Jahrestagung der Gesellschaft f{\"u}r Informatik Aachen, 24.--26. September 1997},
  pages={41--50},
  year={1997},
  publisher={Springer}
}

@article{chan2012objects,
  title={From objects to landmarks: the function of visual location information in spatial navigation},
  author={Chan, Edgar and Baumann, Oliver and Bellgrove, Mark A and Mattingley, Jason B},
  journal={Frontiers in psychology},
  volume={3},
  pages={304},
  year={2012},
  publisher={Frontiers Research Foundation}
}

@article{zacks2000mental,
  title={Mental spatial transformations of objects and perspective},
  author={Zacks, Jeffrey M and Mires, JON and Tversky, Barbara and Hazeltine, Eliot},
  journal={Spatial Cognition and Computation},
  volume={2},
  number={4},
  pages={315--332},
  year={2000},
  publisher={Springer}
}

@article{zacks2002parametric,
  title={A parametric study of mental spatial transformations of bodies},
  author={Zacks, Jeffrey M and Ollinger, John M and Sheridan, Margaret A and Tversky, Barbara},
  journal={Neuroimage},
  volume={16},
  number={4},
  pages={857--872},
  year={2002},
  publisher={Elsevier}
}

@article{hasgul2015space,
  title={Space as configuration: Patterns of space and culture},
  author={Hasg{\"u}l, Esin},
  journal={Proceedings of the ARCHTHEO},
  volume={2015},
  pages={9th},
  year={2015}
}

@article{zerouati2020evaluating,
  title={Evaluating the impact of mass housings' in-between spaces' spatial configuration on users' social interaction},
  author={Zerouati, Wiem and Bellal, Tahar},
  journal={Frontiers of Architectural Research},
  volume={9},
  number={1},
  pages={34--53},
  year={2020},
  publisher={Elsevier}
}

@misc{archdaily,
howpublished = {https://www.archdaily.com/},
title = {{archdaily}},
url = {https://www.archdaily.com/},
urldate = {05/03/26}
}

@misc{gooood,
howpublished = {https://www.gooood.cn/},
title = {{gooood}},
url = {https://www.gooood.cn/},
urldate = {05/03/26}
}

@misc{archiposition,
howpublished = {https://www.archiposition.com/},
title = {{archiposition}},
url = {https://www.archiposition.com/},
urldate = {05/03/26}
}

@inproceedings{kwon2023efficient,
  title={Efficient memory management for large language model serving with pagedattention},
  author={Kwon, Woosuk and Li, Zhuohan and Zhuang, Siyuan and Sheng, Ying and Zheng, Lianmin and Yu, Cody Hao and Gonzalez, Joseph and Zhang, Hao and Stoica, Ion},
  booktitle={Proceedings of the 29th Symposium on Operating Systems Principles},
  pages={611--626},
  year={2023}
}

@article{childs2002matrix,
  title={Matrix sampling of items in large-scale assessments},
  author={Childs, Ruth A and Jaciw, Andrew P},
  journal={Practical Assessment, Research, and Evaluation},
  volume={8},
  number={1},
  year={2002},
  publisher={University of Massachusetts Amherst Libraries}
}

@article{jacobs1991adaptive,
  title={Adaptive mixtures of local experts},
  author={Jacobs, Robert A and Jordan, Michael I and Nowlan, Steven J and Hinton, Geoffrey E},
  journal={Neural computation},
  volume={3},
  number={1},
  pages={79--87},
  year={1991},
  publisher={MIT Press}
}

@article{shazeer2017outrageously,
  title={Outrageously large neural networks: The sparsely-gated mixture-of-experts layer},
  author={Shazeer, Noam and Mirhoseini, Azalia and Maziarz, Krzysztof and Davis, Andy and Le, Quoc and Hinton, Geoffrey and Dean, Jeff},
  journal={arXiv preprint arXiv:1701.06538},
  year={2017}
}

@article{fedus2022review,
  title={A review of sparse expert models in deep learning},
  author={Fedus, William and Dean, Jeff and Zoph, Barret},
  journal={arXiv preprint arXiv:2209.01667},
  year={2022}
}

@inproceedings{raistrick2023infinite,
  title={Infinite photorealistic worlds using procedural generation},
  author={Raistrick, Alexander and Lipson, Lahav and Ma, Zeyu and Mei, Lingjie and Wang, Mingzhe and Zuo, Yiming and Kayan, Karhan and Wen, Hongyu and Han, Beining and Wang, Yihan and others},
  booktitle={Proceedings of the IEEE/CVF Conference on Computer Vision and Pattern Recognition},
  pages={12630--12641},
  year={2023}
}

@inproceedings{raistrick2024infinigen,
  title={Infinigen indoors: Photorealistic indoor scenes using procedural generation},
  author={Raistrick, Alexander and Mei, Lingjie and Kayan, Karhan and Yan, David and Zuo, Yiming and Han, Beining and Wen, Hongyu and Parakh, Meenal and Alexandropoulos, Stamatis and Lipson, Lahav and others},
  booktitle={Proceedings of the IEEE/CVF Conference on Computer Vision and Pattern Recognition},
  pages={21783--21794},
  year={2024}
}

\clearpage
\newpage

\appendix

\section{Detailed Task Design}
\label{sec:Detailed-Task-Design}

\begin{figure}[b]
  \centering
  \includegraphics[width=1\linewidth]{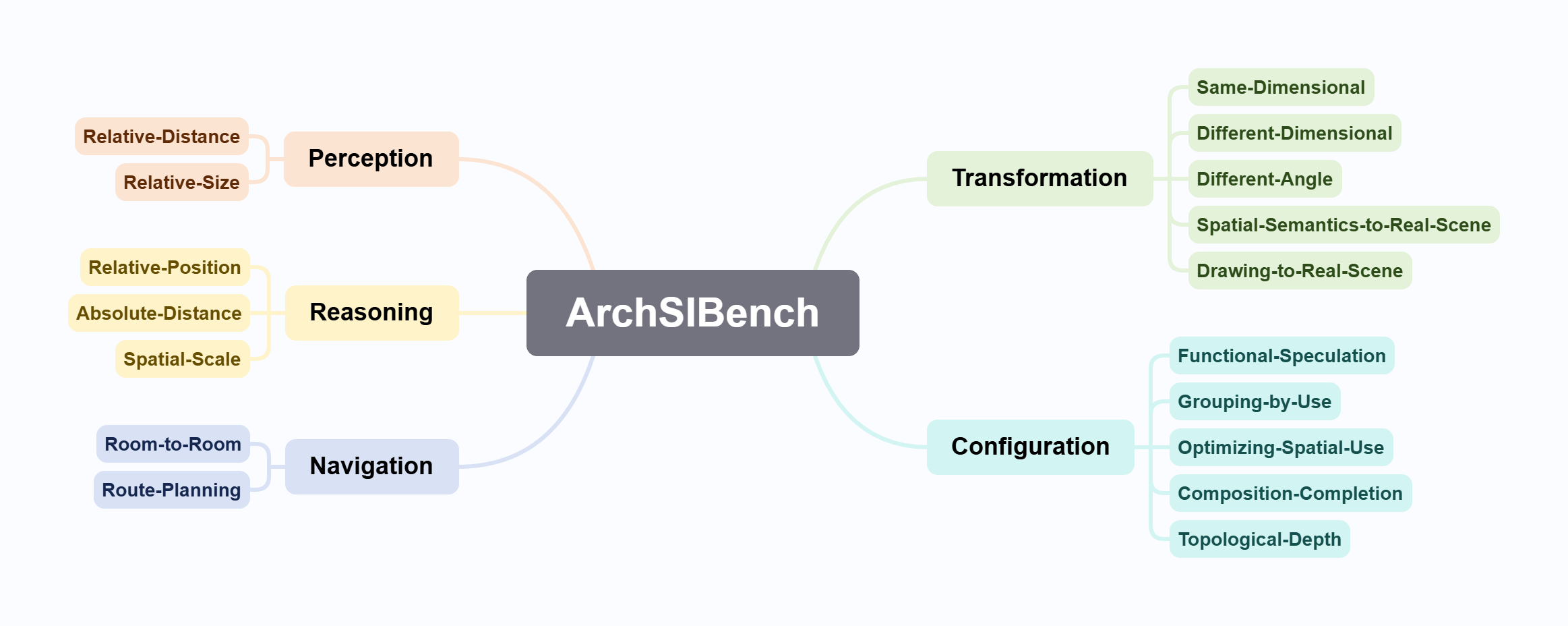}
  \caption{Overview of ArchSIBench Tasks.}
  \label{fig:ArchSIBench}
\end{figure}

ArchSIBench aims to construct a benchmark framework with explicit cognitive hierarchy to provide a unified and systematic assessment of the architectural spatial intelligence of VLMs. For this purpose, the benchmark covers five core dimensions: perception, reasoning, navigation, transformation, and configuration, and is further decomposed into 17 subtasks, some of which have multiple fine-grained question types within them. An overview of tasks in ArchSIBench is shown in Figure~\ref{fig:ArchSIBench}. Each dimension or subtask targets a distinct form of spatial intelligence task and incorporates challenging test cases grounded in both theoretical definitions and practical experience. In this section, we introduce the definition of each task category and discuss its practical significance for downstream applications.

\subsection{Perception}

{\bf Perception} focuses on evaluating the ability of VLMs to form an intuitive understanding of space, including fundamental spatial attributes such as the position of objects relative to the observer, the relative positions between objects, and the approximate size of the space. Here, \emph{intuitive understanding} means that models are not required to provide exact values for distance between objects and spatial size; instead, the emphasis is placed on relative comparisons, such as comparative distance (closer or farther) and comparative size (larger or smaller). Intuition-based relative perception is crucial for applications such as real-time embodied intelligence: models only need to evaluate qualitative relationships such as distance and size, without outputting exact numerical values, thereby allowing for rapid spatial comprehension and response.

\begin{itemize}[left=5pt]
\item {\bf Relative-Distance} Given a floor plan or real-scene image, estimate the relative distance (closer or farther) between two spaces or objects.
\item {\bf Relative-Size} Given a floor plan or real-scene image, estimate the relative size relationship (larger or smaller) between two spaces.
\end{itemize}

\subsection{Reasoning}

{\bf Reasoning} focuses on evaluating the ability of VLMs to go beyond perceptual intuition and infer the exact distance, scale, and relative position between objects, combined with auxiliary information such as the size and orientation of other objects. Distinct from the perception dimension, which emphasizes coarse relational judgments, the reasoning dimension allows for approximate quantification through reference objects and contextual reasoning. This process also includes spatial reasoning based on human scale and embodied experience, such as determining whether the space is crowded or meets usage needs. Based on reasoning ability, models are expected to transition from coarse intuitive perception to more fine-grained spatial understanding, thereby supporting more complex analysis and decision-making tasks.

\begin{itemize}[left=5pt]
\item {\bf Relative-Position} Given a floor plan or real-scene image, infer the relative orientation relationship between two spaces or objects.
\item {\bf Absolute-Distance} Given a floor plan or real-scene image, estimate the exact distance between two spaces or objects.
\item {\bf Spatial-Scale} Given a real-scene image, infer whether the space can meet certain embodied action requirements.
\end{itemize}

\subsection{Navigation}

{\bf Navigation} focuses on evaluating the ability of VLMs to understand structures in complex spaces and determine feasible paths, namely the ability to move from one location to another. This ability not only relies on a comprehensive understanding of spatial layout, but also requires VLMs to identify the constraint relationships formed by building structural elements such as walls, doors, and corridors. In the architectural space, we focus on practically feasible paths rather than geometric shortest paths, meaning that the model needs to select and plan paths within structural constraints. Navigation is crucial for robotic mobility, indoor localization, and path planning.

\begin{itemize}[left=5pt]
\item {\bf Room-to-Room} Given a floor plan, select the shortest or longest path between two spaces.
\item {\bf Route-Planning} Given a floor plan, select a feasible path sequence from a given starting point to a given endpoint.
\end{itemize}

\subsection{Transformation}

{\bf Transformation} focuses on evaluating the ability of VLMs to map across different spatial representations and perform spatial imagination, including perspective transformation, correspondence between 2D and 3D representations, and spatial reconstruction based on existing information. Transformation refers to the ability of VLMs to go beyond the current perspective and mentally reconstruct spatial representations, rather than relying solely on a single visual input for judgment. This type of ability is manifested in cognitive science as mental rotation and mental folding, and in architecture as an understanding of the relationships between different drawings and forms of expression. Models with strong transformation abilities can establish consistent representations across multi-view and multimodal information, thereby supporting more complex spatial understanding and reasoning tasks.

\begin{itemize}[left=5pt]
\item {\bf Same-Dimensional} Given two 2D drawings with different perspectives (e.g., floor plan and section view), identify the corresponding position of a given point from one drawing in the other.
\item {\bf Different-Dimensional} Given two drawings of different dimensions (e.g., floor plan to axonometric view), identify the corresponding position of a given point from one drawing in the other.
\item {\bf Different-Angle} Given a set of real-scene images from parallel human viewpoints at different angles, select the photo most likely to be taken in the same space as a given reference image.
\item {\bf Spatial-Semantics-to-Real-Scene} Given a text describing the spatial information of a scene, select the real-scene image that best matches the description.
\item {\bf Drawing-to-Real-Scene} Given a floor plan and a preset perspective, select the most likely real-scene image to be presented in the space from the preset perspective.
\end{itemize}

\subsection{Configuration}

{\bf Configuration} focuses on evaluating the ability of VLMs to understand the overall structure and organization of space, including the combination relationships among spaces, functional zoning, and underlying functional logic of usage patterns. This capability is particularly important in architecture, containing potential capabilities for architectural design and generation, as spatial configuration directly shapes human behavior and spatial experience. In practical applications, understanding spatial configuration not only supports analysis of existing environments, but also provides a foundation for design generation and optimization, enabling models to reason about spatial structure and function at a higher level.

\begin{itemize}[left=5pt]
\item {\bf Functional-Speculation} Given a real-scene image, infer the function based on the overall configuration of the corresponding space in the image.
\item {\bf Grouping-by-Use} Given a floor plan, determine the functional zoning of a given space based on usage logic.
\item {\bf Optimizing-Spatial-Use} Given a set of floor plans and a description of spatial usage requirements, determine which floor plan represents the spatial combination that can more effectively meet the usage requirements.
\item {\bf Composition-Completion} Given a floor plan with partially occluded regions, infer the most appropriate spatial completion based on the missing areas and the existing layout.
\item {\bf Topological-Depth} Given a floor plan, determine the topological depth of the specified spaces.
\end{itemize}

\section{Detailed Results of Different Series VLMs}
\label{sec:Detailed-Results-of-Different-Series-VLMs}

\begin{figure}[t]
    \centering

    \begin{subfigure}[t]{0.49\linewidth}
        \centering
        \includegraphics[width=\linewidth]{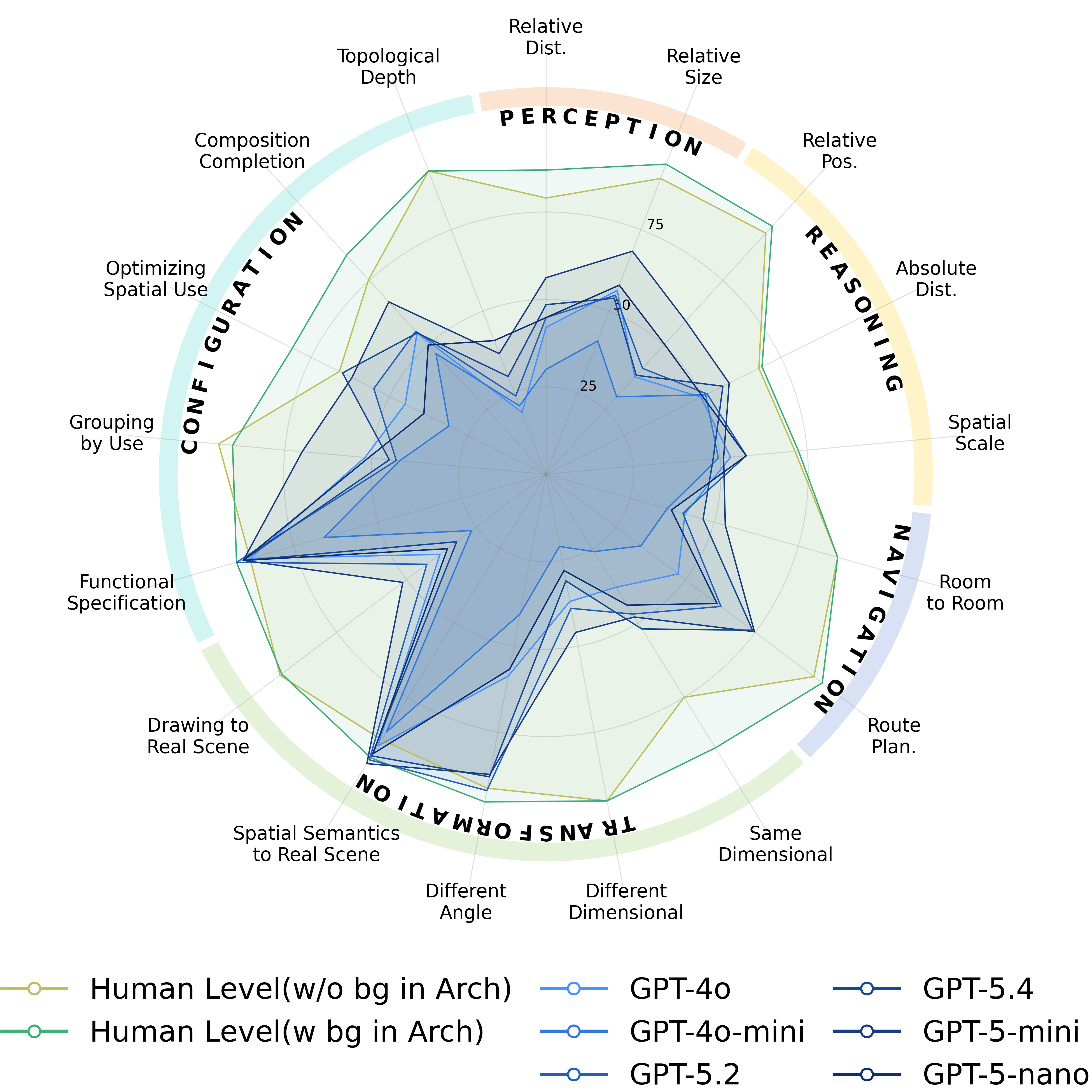}
        \caption{GPT family}
        \label{fig:GPT}
    \end{subfigure}
    \hfill
    \begin{subfigure}[t]{0.49\linewidth}
        \centering
        \includegraphics[width=\linewidth]{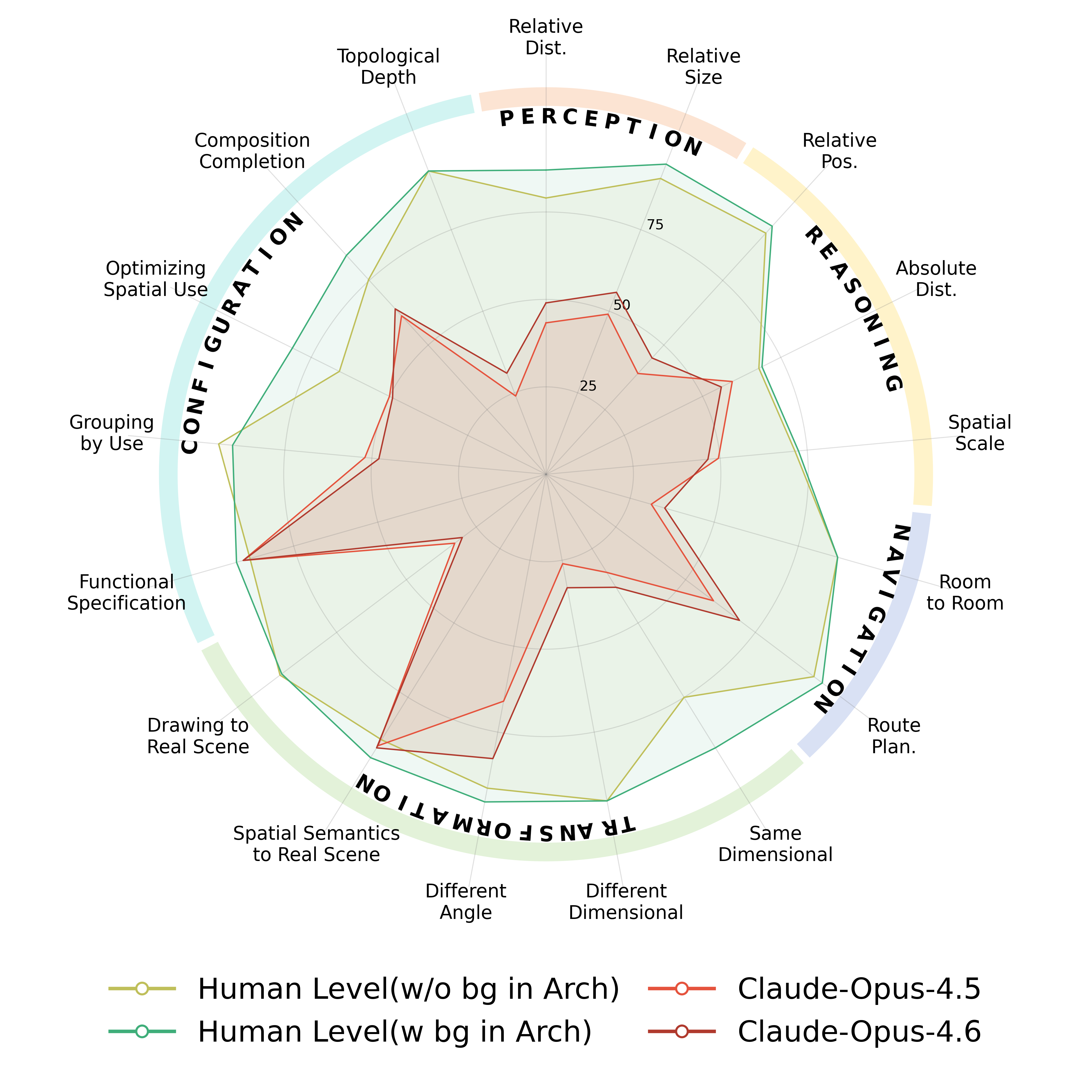}
        \caption{Claude-Opus-4 family}
        \label{fig:Claude}
    \end{subfigure}

    \begin{subfigure}[t]{0.49\linewidth}
        \centering
        \includegraphics[width=\linewidth]{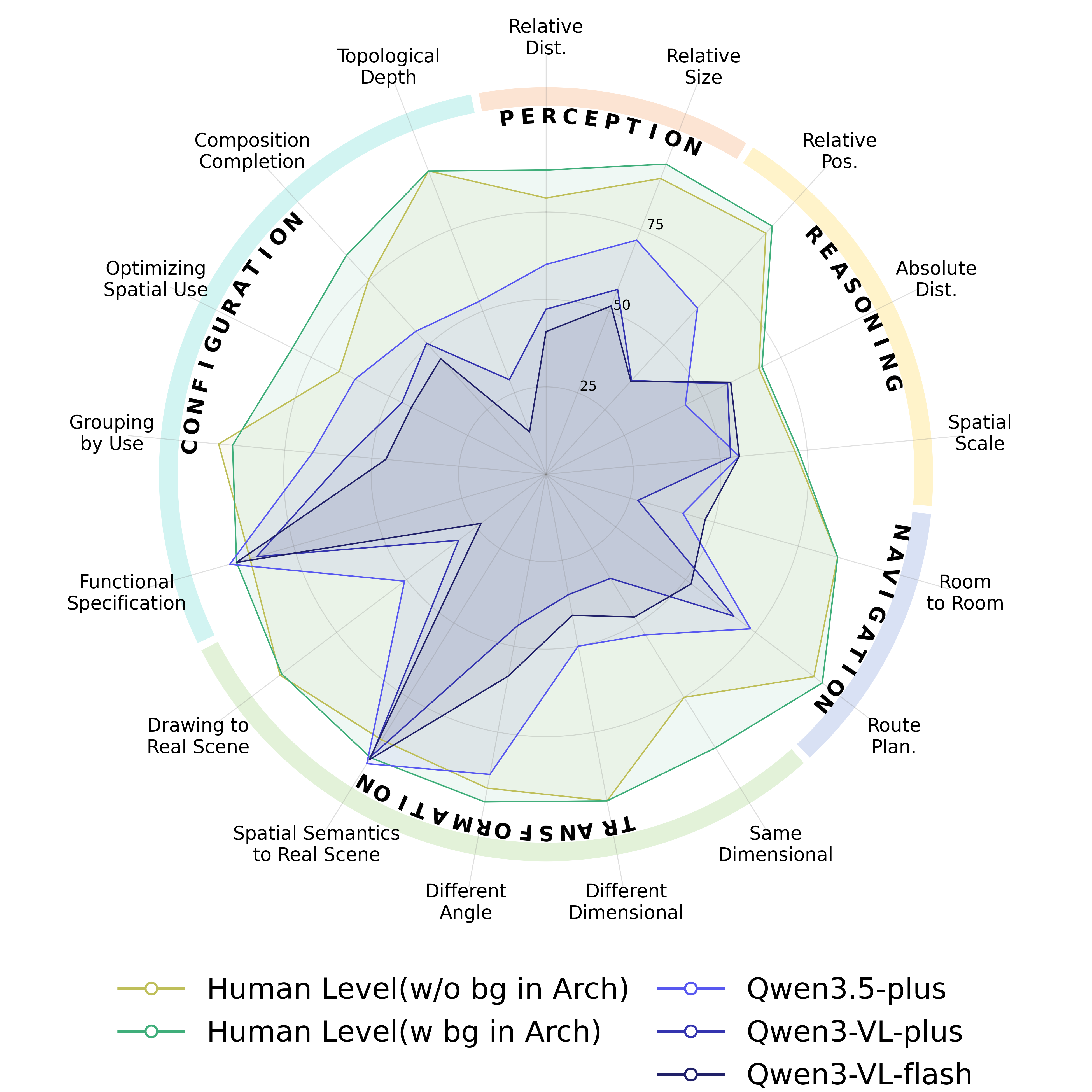}
        \caption{Qwen proprietary family}
        \label{fig:Qwen_Closed}
    \end{subfigure}
    \hfill
    \begin{subfigure}[t]{0.49\linewidth}
        \centering
        \includegraphics[width=\linewidth]{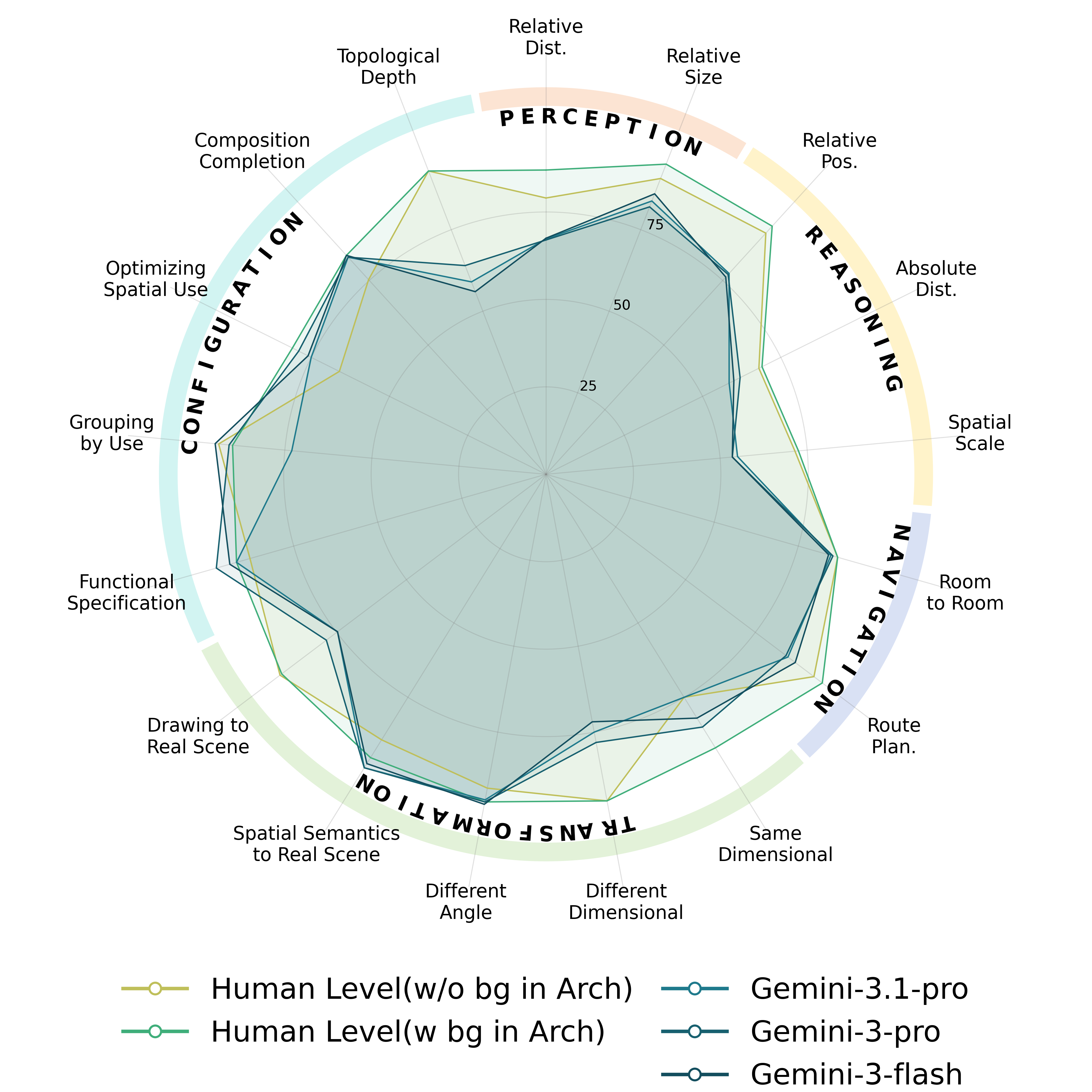}
        \caption{Gemini-3 family}
        \label{fig:Gemini}
    \end{subfigure}

    \caption{Performance of Proprietary VLMs.}
    \label{fig:proprietary_models}
\end{figure}

\begin{figure}[t]
    \centering

    \begin{subfigure}[t]{0.49\linewidth}
        \centering
        \includegraphics[width=\linewidth]{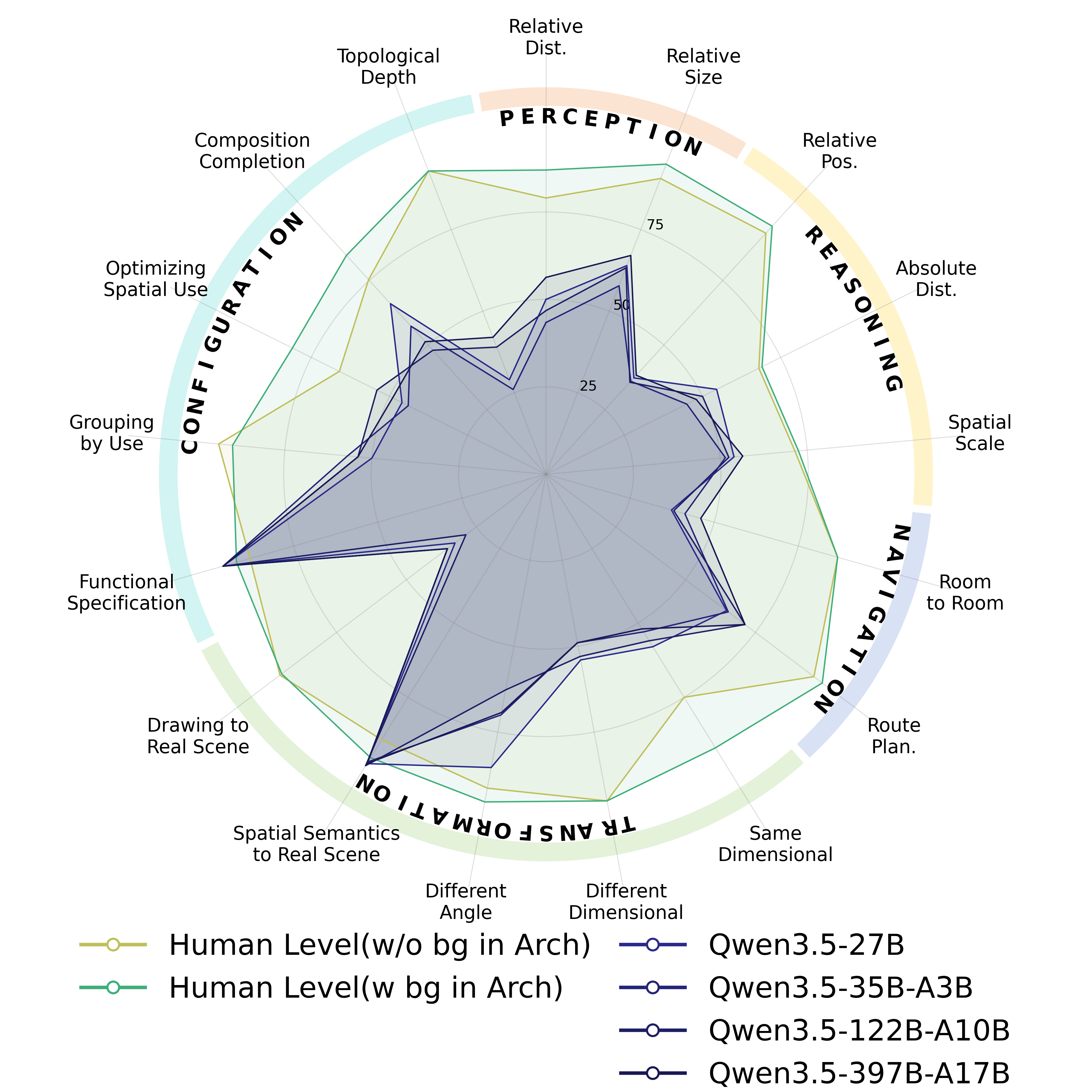}
        \caption{Qwen open-source family}
        \label{fig:Qwen3.5}
    \end{subfigure}
    \hfill
    \begin{subfigure}[t]{0.49\linewidth}
        \centering
        \includegraphics[width=\linewidth]{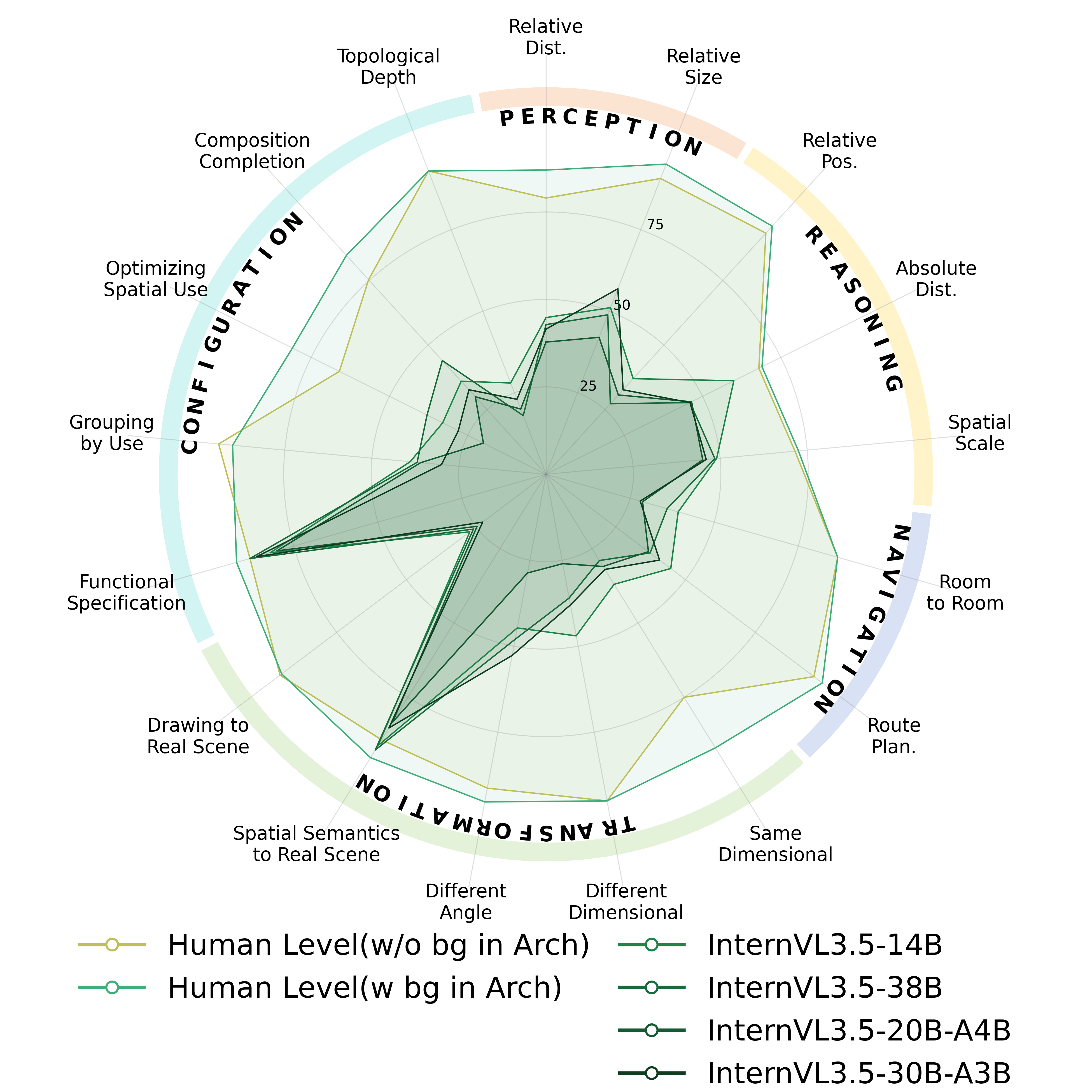}
        \caption{InternVL3.5 family}
        \label{fig:InternVL3.5}
    \end{subfigure}

    \begin{subfigure}[t]{0.49\linewidth}
        \centering
        \includegraphics[width=\linewidth]{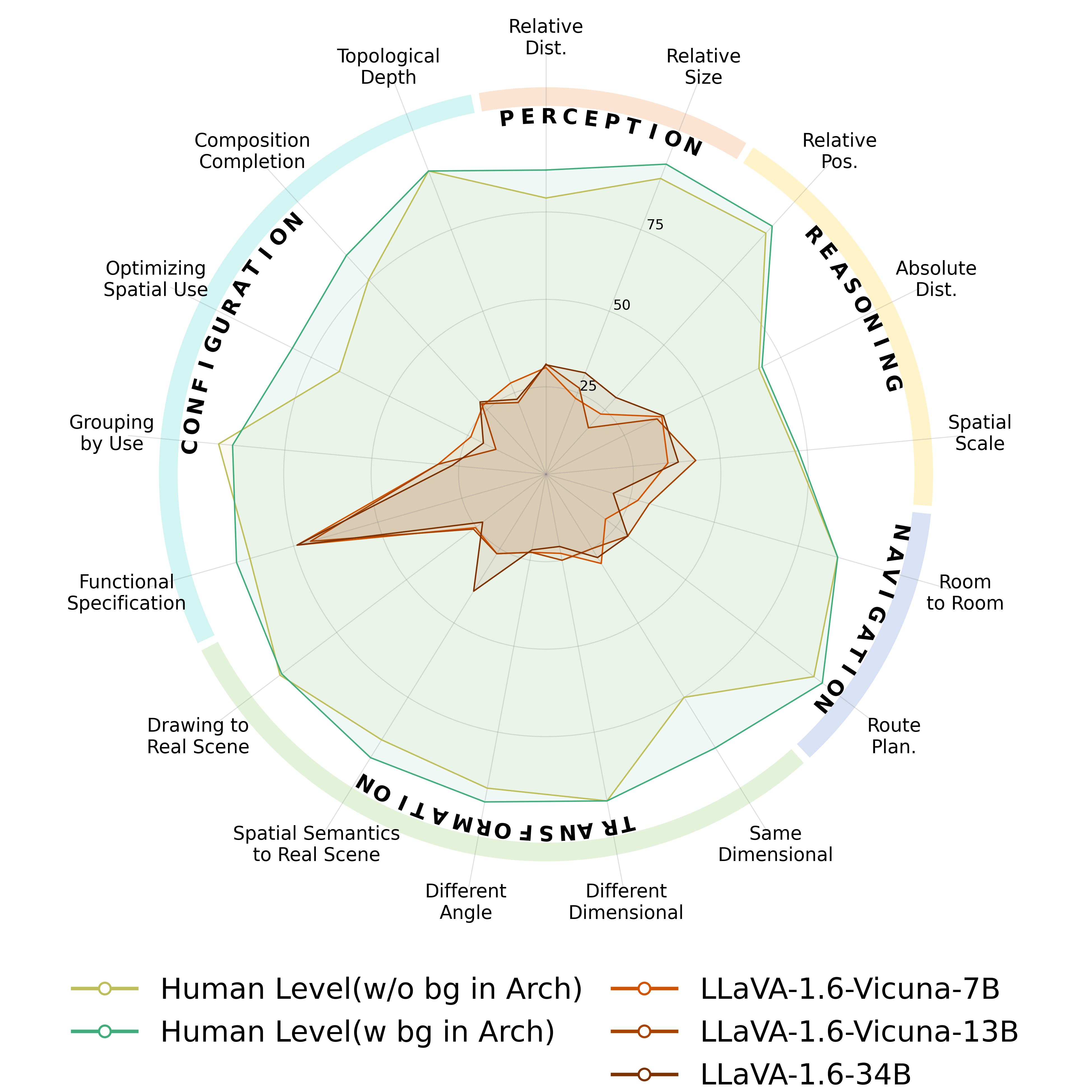}
        \caption{LLaVA-1.6 family}
        \label{fig:LLaVA-1.6}
    \end{subfigure}
    \hfill
    \begin{subfigure}[t]{0.49\linewidth}
        \centering
        \includegraphics[width=\linewidth]{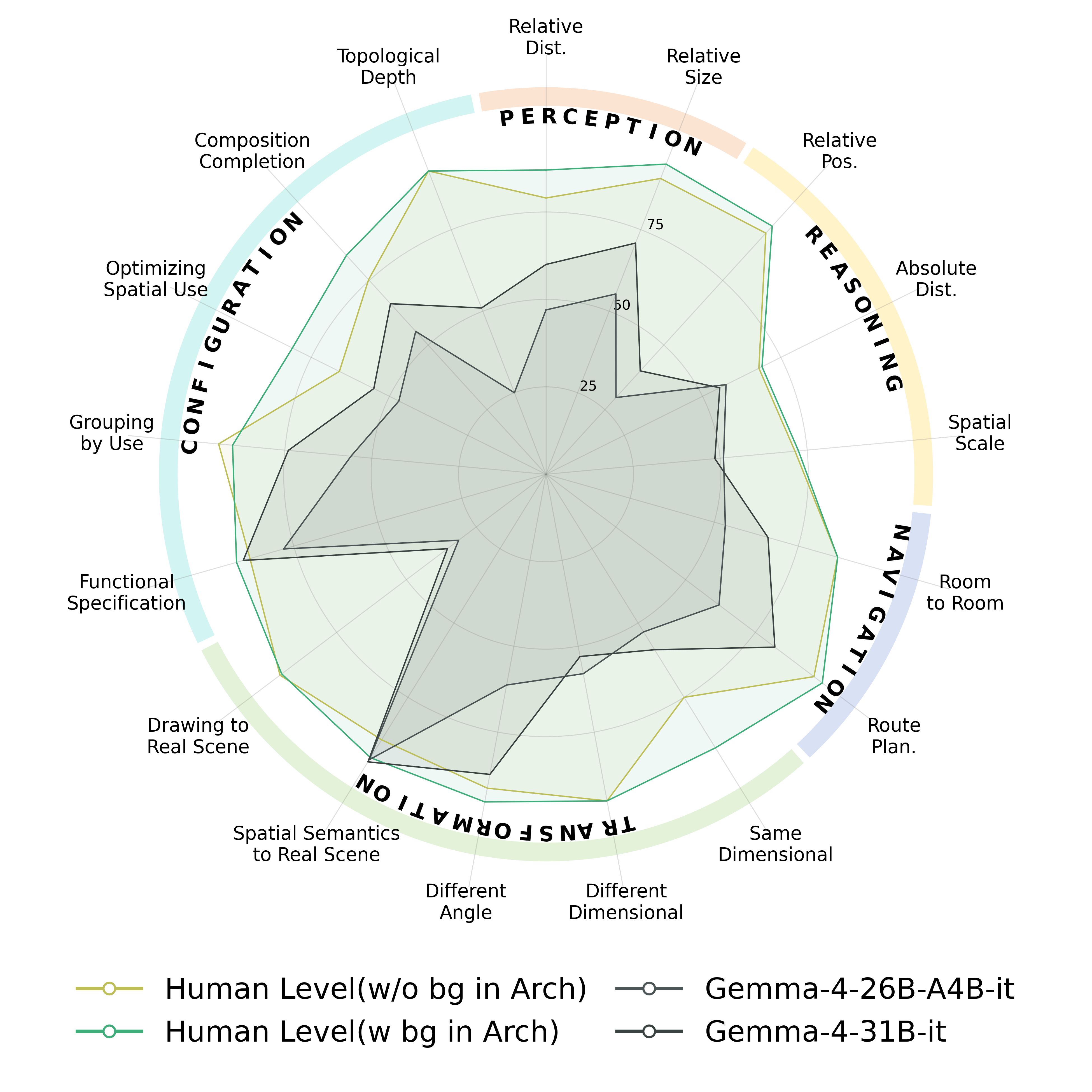}
        \caption{Gemma-4 family}
        \label{fig:Gemma-4}
    \end{subfigure}

    \caption{Performance of Open-Source VLMs.}
    \label{fig:open_source_models}
\end{figure}

We present the performance of different VLM series, as shown in Figure~\ref{fig:proprietary_models} and Figure~\ref{fig:open_source_models}.

From the radar visualizations, we observe that this non-smooth, ``zigzag'' performance pattern is consistently shared across different model families. Both proprietary and open-source VLMs exhibit similar non-uniform distributions across subtasks, with near-human performance on some tasks but substantial deficits on others, suggesting a common structural limitation rather than model-specific variance. This pattern contrasts with human evaluators, especially those with architectural training who show high accuracy and low variance, indicating stable, global spatial reasoning strategies. In contrast, VLMs rely more on task-specific heuristics, resulting in limited generalization across spatial reasoning tasks. Overall, this suggests that current VLMs lack a unified ``world model'', and that progress may require advances beyond scaling, such as structured spatial representations, geometric priors, and training for globally consistent reasoning.

\section{Evaluation Details}
\label{sec:Evaluation-Details}

\begin{table*}[t!]
\centering
\vspace{-5pt}
\caption{\textbf{Question templates for tasks in ArchSIBench. }We replace the angle brackets \texttt{<>} part in the templates to instantiate benchmark questions.}
\vspace{-5pt}

\renewcommand{\arraystretch}{1.3}
\setlength{\tabcolsep}{5pt}

\resizebox{0.85\linewidth}{!}{
\begin{tabular}{r|p{13.7cm}}
\toprule
\textbf{Task} & \textbf{Question Template} \\
\midrule

\rowcolor{per}
\multicolumn{2}{c}{\textbf{Perception}} \\

Relative Distance & - \\
\hline

Type 1 &
\textit{Starting from the entrance of each room, which room is the farthest and closest from <point A>?} \\
\hline

Type 2 &
\textit{Considering the actual accessible paths in the building, starting from the entrance of each room, which room is the farthest and closest from <point A>?} \\
\hline

Type 3 &
\textit{From the perspective of the photographer, which object in the image is at the <shortest> (or <longest>) distance?} \\
\hline

Type 4 &
\textit{From the perspective of the woman wearing white and yellow clothes, which object in the image is at the <shortest> (or <longest>) distance?} \\
\hline

Relative Size & - \\
\hline

Type 1 &
\textit{Which are the largest and smallest rooms in the following picture? } \\
\hline

Type 2 &
\textit{Consider the enclosed space composed of walls, doors, windows, entrances, etc., and determine which letter in the following picture represents the largest and smallest room, respectively.} \\
\hline

Type 3 &
\textit{Which are the largest and smallest rooms in the following picture?} \\
\hline

Type 4 &
\textit{Consider the enclosed space composed of walls, doors, windows, entrances, etc., and determine which letter in the following picture represents the largest and smallest room, respectively.} \\
\hline

Type 5 &
\textit{Which room is <larger> (or <smaller>)?} \\

\midrule

\rowcolor{rea}
\multicolumn{2}{c}{\textbf{Reasoning}} \\

Relative Position & - \\
\hline

Type 1 &
\textit{If you are a robot, when you are located at Point A facing Point B in the diagram, which direction is the <room 1> (or <room 2/3/4...>)?} \\
\hline

Type 2 &
\textit{If you are a robot, when you are located at Point A facing Point B in the diagram, which direction is the <exit> (or <bed/window/stairs...>) of the room you are in.} \\
\hline

Type 3 &
\textit{Consider the real-world 3D locations and orientations of the objects. Which side of the <woman in black> (or other character) is facing towards the <green plants> (or other item)?} \\
\hline

Absolute Distance & - \\
\hline

Type 1 &
\textit{Estimate the distance from the entrance of Room A to the entrance of Room B based on furniture and other reference materials.} \\
\hline

Type 2 &
\textit{Consider the real-world 3D locations. What is the distance between the <wooden checkered steps> and <cyan sofa> (or other item)?} \\
\hline

Spatial Scale & - \\
\hline

Type 1 &
\textit{Based on spatial scale, determine whether the following behavior is possible: <You are an adult male and you want to stand up straight on the bed on the left> (or other descriptions, such as: You are a ten year old boy, and you want to stand straight on the bottom bunk of this bunk bed?).} \\
\hline

Type 2 &
\textit{What kind of feeling may <three> (or <one/four/five...>) people feel when they are in the <dining space> (or <living room/study space...>) shown in the picture at the same time?} \\

\midrule

\rowcolor{nav}
\multicolumn{2}{c}{\textbf{Navigation}} \\

Room to Room &
\textit{If you were a robot, what would be the <shortest> (or <longest>) path from Point A to Point B in the diagram?} \\
\hline

Route Planning &
\textit{If you are a robot, which path is feasible for you to reach Point B from Point A in the diagram (where you are standing facing north)?} \\

\midrule

\rowcolor{tra}
\multicolumn{2}{c}{\textbf{Transformation}} \\

Same Dimensional &
\textit{Which point in the section (the image below) may correspond to the position of <point X> in the plan view (the image above)?} \\
\hline

Different Dimensional &
\textit{Which point in the plan view may correspond to the position of <point X> in the axonometric diagram?} \\
\hline

Different Angle &
\textit{If you were a robot, when you were in the space shown in the picture below and turned your head horizontally, which space in the picture would you most likely see?} \\
\hline

Spatial Semantics to Real Scene &
\textit{Which image is most similar to the following text description?} \\
\hline

Drawing to Real Scene &
\textit{If you were a robot, which image would you most likely see when you are at <point X> in the following picture and looking in the direction of the arrow?} \\

\midrule

\rowcolor{con}
\multicolumn{2}{c}{\textbf{Configuration}} \\

Functional Speculation &
\textit{What are people most likely to do in this space?} \\
\hline

Grouping by Use &
\textit{Which grouping most reasonably separates <public> and <private> (or another set of opposing descriptions, such as <indoor> and <outdoor>, <noisy> and <quiet>) space in the following floor plan?} \\
\hline

Optimizing Spatial Use &
\textit{Which of the following layouts better satisfies the following requirement?} \\
\hline

Composition Completion &
\textit{Which of the following space options best completes the RED missing area in the floor plan based on functional logic and spatial adjacency?} \\
\hline

Topological Depth &
\textit{For point A, which room has the <smallest> (or <largest>) topological depth (Topology depth can be simplified as the number of intermediate rooms that need to be passed from one room to another)?} \\

\bottomrule
\end{tabular}
}
\vspace{-5pt}
\label{tab:question_templates}
\end{table*}
\subsection{Question Templates}

ArchSIBench contains 28 different types of question. We construct a corresponding question template for each type of question. Depending on the requirements of each task, the text enclosed in angle brackets (``<>'') within a template can be replaced with specific values to efficiently generate questions in bulk. The templates for all question types are summarized in Table~\ref{tab:question_templates}.

In ArchSIBench, question-answer pairs are stored as an image directory and a corresponding questions.json file. In addition to the images, the JSON entry for each question-answer pair contains information such as question ID, task\_type, sub\_task\_type, stem, options, and answer. An example JSON entry is shown below:

\begin{tcolorbox}[
enhanced,
breakable,
title=JSON Template,
colback=black!3,
colframe=black!70,
listing only,
listing engine=listings
]
\{\\
    "id": "q0000",\\
    "task\_type": "Perception",\\
    "sub\_task\_type": "Relative\_Distance",\\
    "stem": \{\\
        "text": "Starting from the entrance of each room, which room is the farthest and closest from point A?",\\
        "images": [\\
        \{
        "id": "Perception/Relative\_Distance/q0000\_image0",\\
        "path": "images/Perception/Relative\_Distance/q0000\_image0.png"
        \}
        ]
    \},\\
    "options": [\\
    \{
        "label": "A",
        "type": "text",
        "content": "Room 1, Room 3"
    \},\\
    \{
        "label": "B",
        "type": "text",
        "content": "Room 2, Room 1"
    \},\\
    \{
        "label": "C",
        "type": "text",
        "content": "Room 3, Room 2"
    \},\\
    \{
        "label": "D",
        "type": "text",
        "content": "Room 4, Room 3"
    \}
    ],\\
    "answer": "D"\\
\}
\end{tcolorbox}

\subsection{Unified Prompt}

To ensure input consistency and reproducibility of results, we adopt a unified prompt template for all questions:

\begin{tcolorbox}[
enhanced,
breakable,
title=Prompt Template,
colback=black!3,
colframe=black!70,
listing only,
listing engine=listings
]
"prompt":\\
"You are a spatial cognitive assistant. Based on the image and question, provide your answer. Always ground your answer in the visual evidence; do not hallucinate unseen objects. If uncertain, pick the most plausible option.never refuse or reply `insufficient information.'\\
\\
Question: Starting from the entrance of each room, which room is the farthest and closest from point A? A single image is provided representing the question's scene.\\
\\
Options:\\
A: Room 1, Room 3\\
B: Room 2, Room 1\\
C: Room 3, Room 2\\
D: Room 4, Room 3\\
\\
Answer the most appropriate option label (A/B/C/D). Your output must consist of ONLY one letter: A, B, C, or D (if applicable. Some questions only have two or three options, and you can only choose the answer you believe to be correct from those). DO NOT output ANY additional information.",\\
\\
"images": [base64]
\end{tcolorbox}

\begin{figure}[t]
    \centering
    
    \begin{minipage}[t]{0.48\linewidth}
        \centering
        \includegraphics[width=\linewidth]{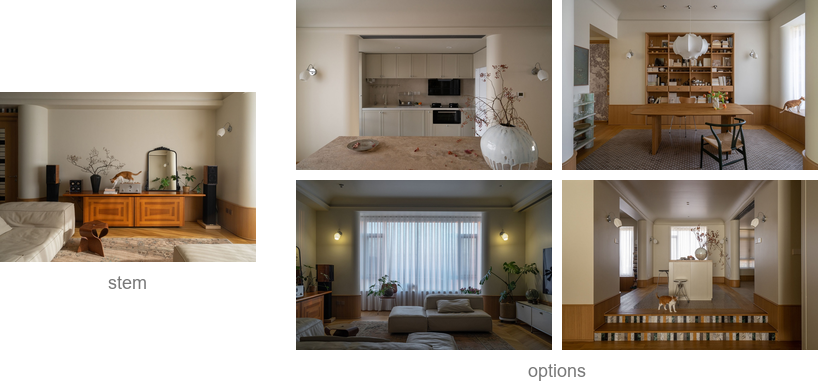}
        \caption{Merged image example 1.}
        \label{fig:Merged_Image_1}
    \end{minipage}
    \hfill
    \begin{minipage}[t]{0.48\linewidth}
        \centering
        \includegraphics[width=\linewidth]{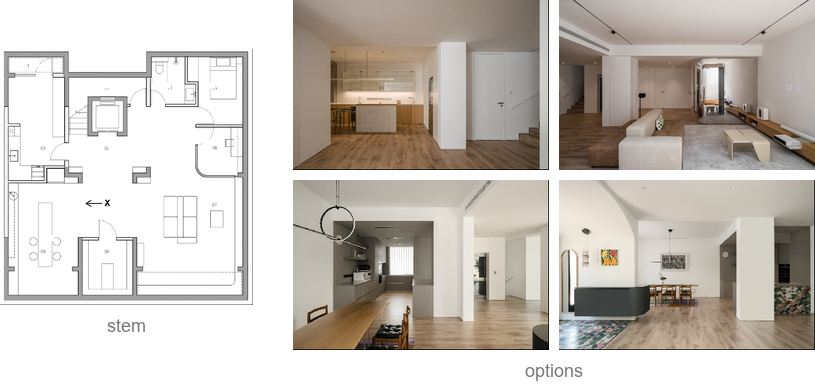}
        \caption{Merged image example 2.}
        \label{fig:Merged_Image_2}
    \end{minipage}
\end{figure}

In addition, to avoid model capability fluctuations caused by differences in multi-image processing capabilities, for questions involving multiple images, we use a unified pipeline to merge the question and option images into a single image. The example of the merged image is shown in Figure~\ref{fig:Merged_Image_1} and Figure~\ref{fig:Merged_Image_2}.

\subsection{Human Evaluation Setup}

\begin{figure}[t]
  \centering

  \begin{subfigure}{\linewidth}
    \centering
    \includegraphics[width=\linewidth]{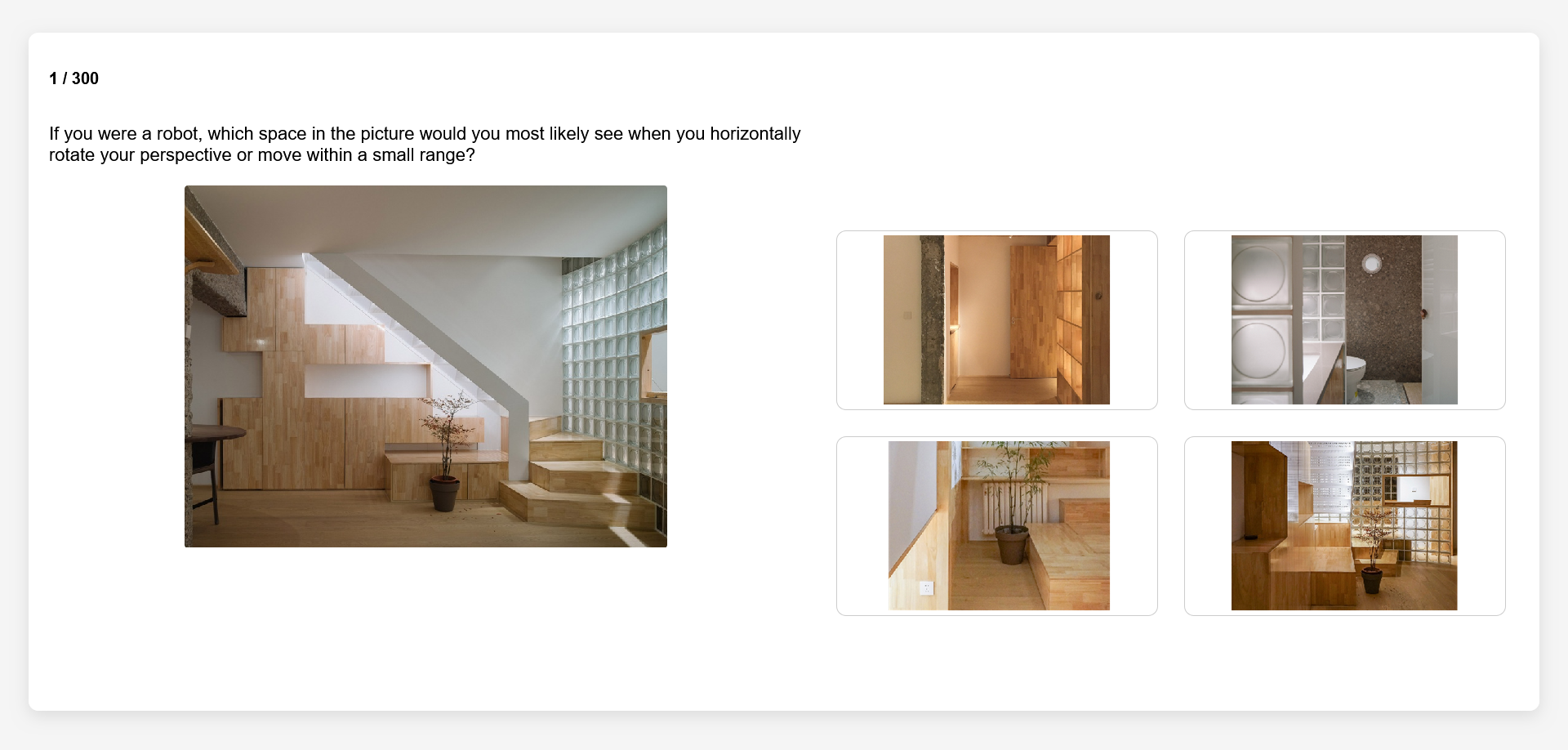}
    \label{fig:Interface_1}
  \end{subfigure}

  \begin{subfigure}{\linewidth}
    \centering
    \includegraphics[width=\linewidth]{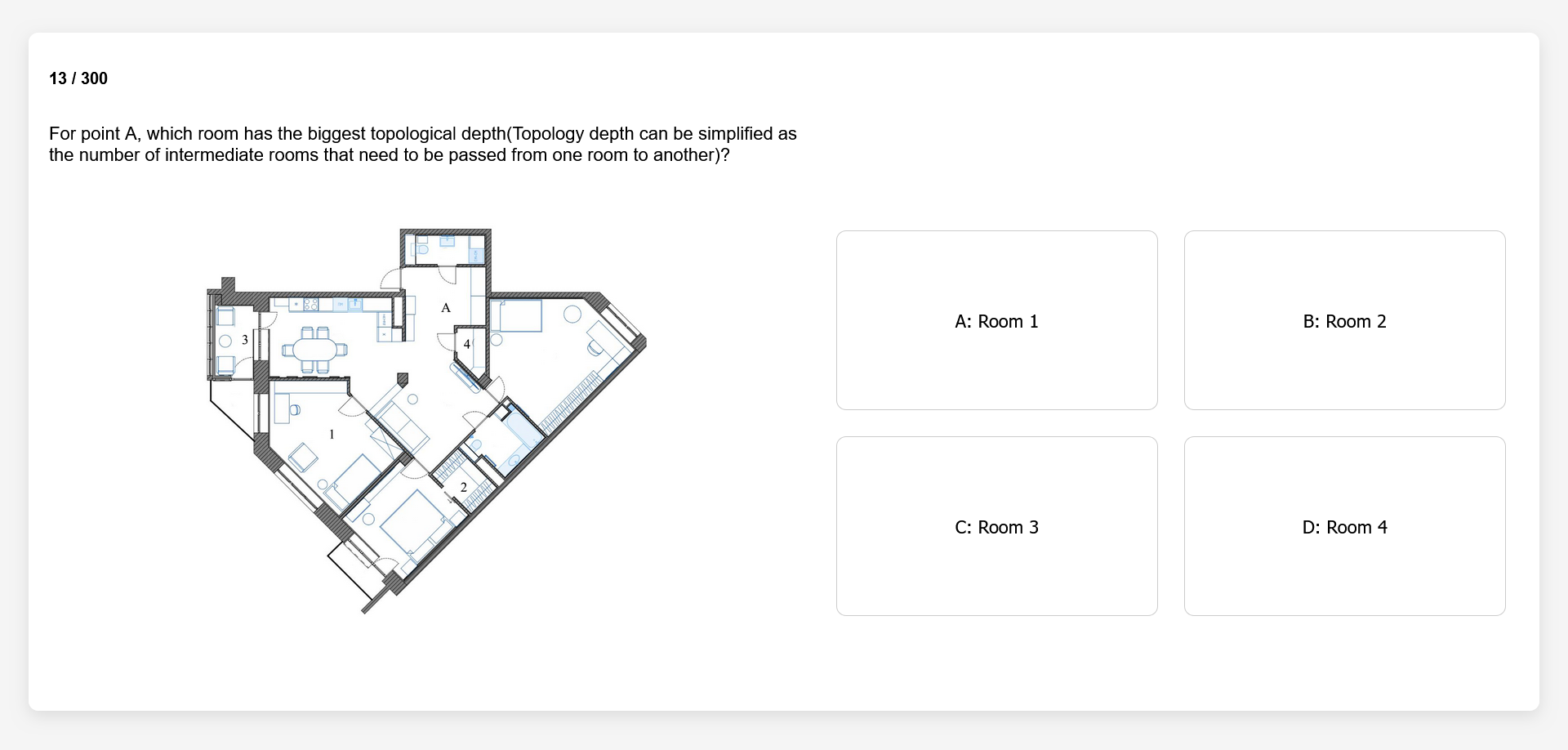}
    \label{fig:Interface_2}
  \end{subfigure}

  \caption{Human evaluation interfaces.}
  \label{fig:Human_evaluation_interfaces}
\end{figure}

Human baseline performance is a key reference point in our work. To quantify the extent to which VLMs can understand architectural space like humans, we establish two human baselines: the human baseline with architectural education backgrounds and the human baseline without architectural education backgrounds. Examples of the human evaluation interface are shown in Figure~\ref{fig:Human_evaluation_interfaces}.

We further analyze the results from the two human baselines. Figure~\ref{fig:Result_HumanVSGemini3} compares the performance of both human groups with that of the best-performing VLM on ArchSIBench. Our main findings are as follows:

\begin{figure}
  \centering
  \includegraphics[width=0.6\linewidth]{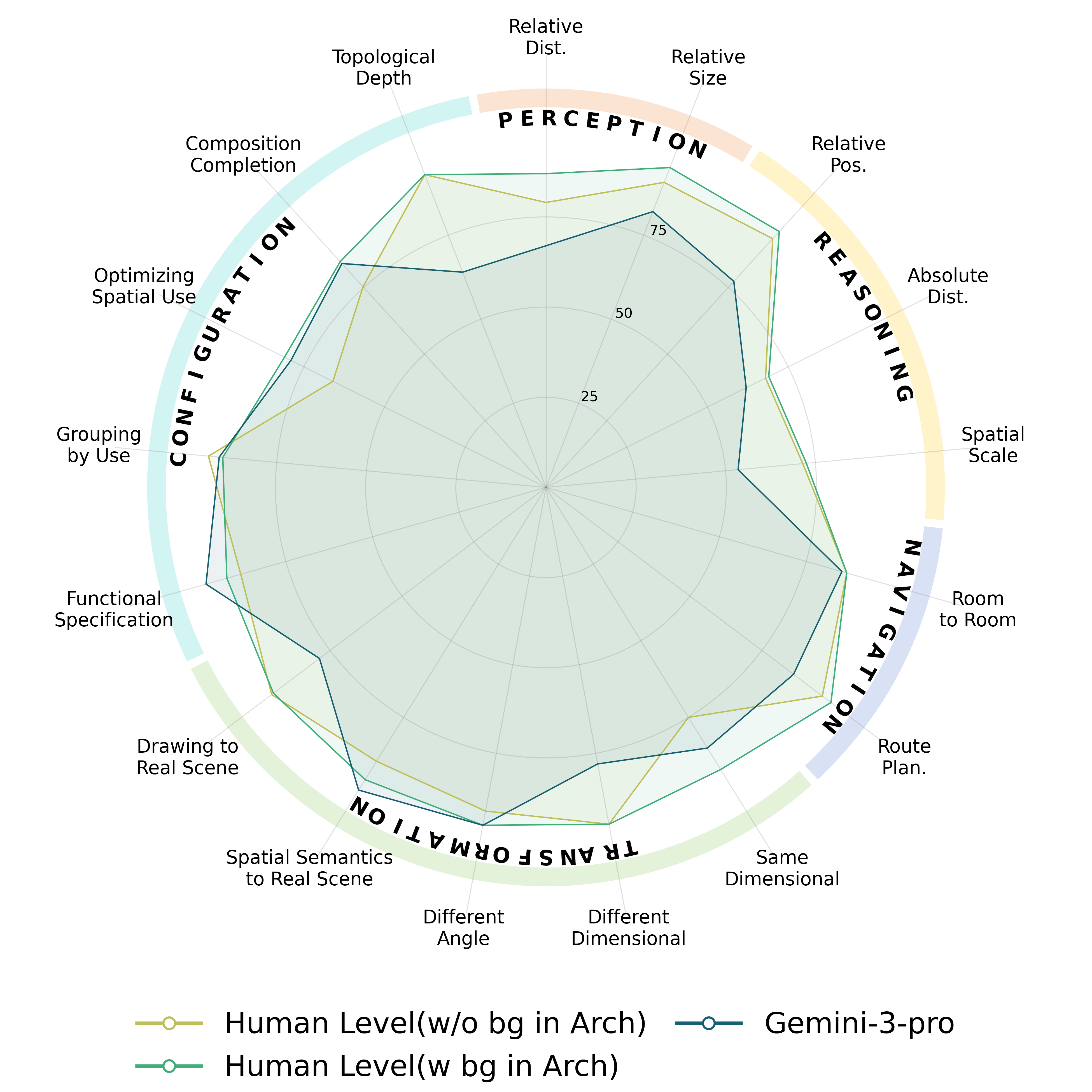}
  \caption{Performance of two human baselines with the best-performing VLM on ArchSIBench.}
  \label{fig:Result_HumanVSGemini3}
\end{figure}

{\bf Perception} dimension shows a noticeable gap between the two human groups, whereas {\bf Reasoning} dimension exhibits minimal difference, with both groups achieving relatively lower scores. These results suggest that perception-based tasks are more aligned with spatial intuition, suggesting that professional training may enhance students' intuitive understanding of distance and scale. In particular, the Absolute-Distance task requires metric estimation of spatial distances. However, humans are generally more adept at encoding relative relationships than absolute measurements, and in the absence of explicit scale references, absolute distance estimation remains inherently uncertain. The Spatial-Scale task involves reasoning about human actions and plausible occupancy in real scenes, which is highly experience-dependent and may admit multiple valid interpretations; in contrast, architectural training typically emphasizes scale reasoning in floor plans or sectional drawing representations rather than real-world embodied scenes.

{\bf Navigation} dimension shows little difference between the two human groups, with both achieving relatively high performance. We interpret this as evidence that navigation is a highly universal human spatial cognition ability that does not strongly depend on professional training.

{\bf Transformation} and {\bf Configuration} dimensions exhibit more varied patterns. Within ArchSIBench, the Same-Dimensional task requires establishing correspondences between floor plans and sectional drawings, where professional training provides a clear advantage. In contrast, Different-Dimensional and Drawing-to-Real-Scene tasks resemble map-reading style reasoning, which may not rely heavily on professional training. Tasks such as Optimizing-Spatial-Use and Composition-Completion involve spatial configuration understanding and thus benefit significantly from professional architectural training. The Topological-Depth task can often be solved through intuitive reasoning, while Grouping-by-Use yields consistently high scores in both groups, likely due to reliance on general everyday commonsense knowledge.



\section{Error Analysis}
\label{sec:Errors-Analysis}

We select two mid-tier models from the Claude and GPT series: Claude-Opus-4.5 and GPT-5.2, and ask them to answer questions and output their thinking process. We compare and analyze their outputs, and summarize several common categories of errors in this section. Detailed case studies are presented in Appendix~\ref{sec:Case-Study}.

\begin{itemize}[left=5pt]
\item {\bf Visual Perception Error.} This refers to fundamental perceptual errors occurring during the processing of visual inputs, including incorrect judgments or omissions regarding object existence, position, color, shape, and spatial layout. Such errors do not involve explicit reasoning and instead arise from inaccurate or incomplete interpretation of visual evidence.
\item {\bf Relational Reasoning Error.} This refers to the model's failures in correctly establishing or computing quantitative or qualitative relationships among spatial entities, such as distance, relative size, topological connectivity, or path length. Such errors typically occur during multi-step accumulation, comparison, or spatial inference processes based on reference objects.
\item {\bf Architectural Element Understanding Error.} This refers to the model's failure to correctly identify or interpret the meanings of various elements in architectural drawings, such as stair orientation, room functions, furniture categories, spatial enclosure relationships, or plan symbols, thereby leading to subsequent errors in spatial understanding.
\item {\bf Viewpoint Transformation Error.} This refers to the model's failures in maintaining spatial consistency when mapping across different spatial representations (e.g., floor plans, sections, and axonometric views) or during mental rotation and viewpoint transformation, often due to the lack of a cross-perspective, global ``world model''. Typical errors include incorrect cross-view point mapping, misjudgment of spatial continuity under local viewpoint changes, and failures in egocentric-allocentric perspective transformation.
\item {\bf Embodied Scale Reasoning Error.} This refers to the model's failure to consistently integrate human scale, furniture scale, and spatial scale, resulting in incorrect judgments of embodied interaction-related properties such as passability, spatial capacity, or spatial compatibility.
\item {\bf Logical Reasoning Error.} This refers to inconsistent reasoning under semantic or commonsense constraints, including violations of spatial functional plausibility, usage logic, or behavioral constraints. Such errors do not directly arise from visual perception or spatial computation, but from an overall incoherence in the reasoning process or deviations from typical human-like logic.
\item {\bf Semantic Perception Error.} This refers to the model's failures in correctly aligning textual semantics with visual evidence, such as misinterpreting key semantic cues or incorrectly weighting important features, thereby leading to subsequent spatial reasoning errors. Such errors do not necessarily involve explicit reasoning and instead arise from incorrect interpretation or omission of the textual semantics themselves.
\end{itemize}

\section{Case Study}
\label{sec:Case-Study}

\captionsetup[figure]{labelformat=simple, labelsep=colon, name=Figure}
\renewcommand{\thefigure}{G\arabic{figure}}
\setcounter{figure}{0}

The appendix presents qualitative analysis of Claude-Opus-4.5 and GPT-5.2, including an analysis of 28 examples, to illustrate more examples of ArchSIBench to elaborate on specific task arrangements, question and answer settings, image settings, and error cases. 

\hypertarget{listofcasestudyfig}{}
\listofcasestudyfig

\casestudyfigure{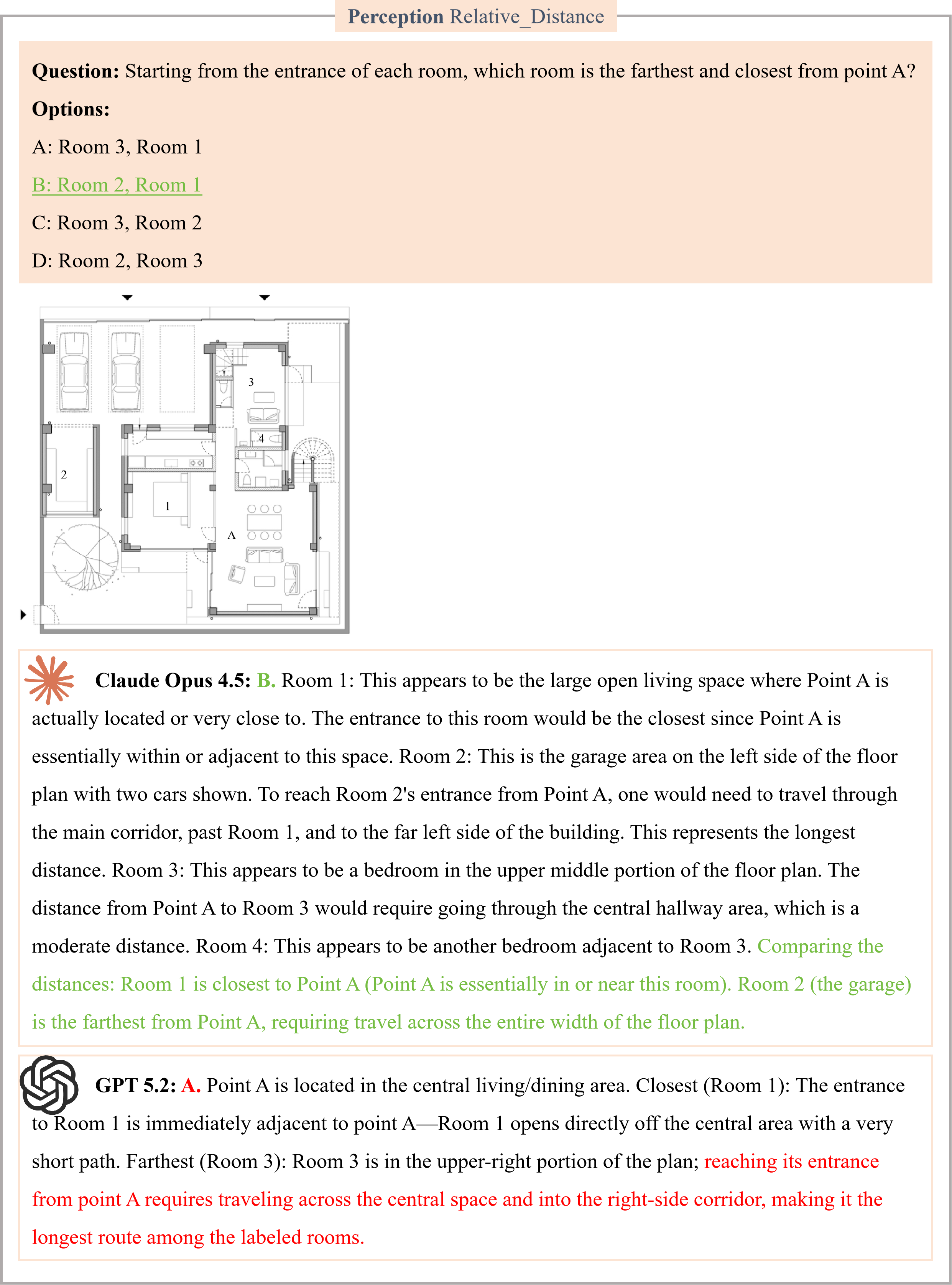}%
  {Case 1}%
  {Relational Reasoning Error: GPT-5.2 failed to correctly aggregate and compare the actual traversable path lengths from point A to the entrances of different rooms.}%
  {fig:case_study_1}

\casestudyfigure{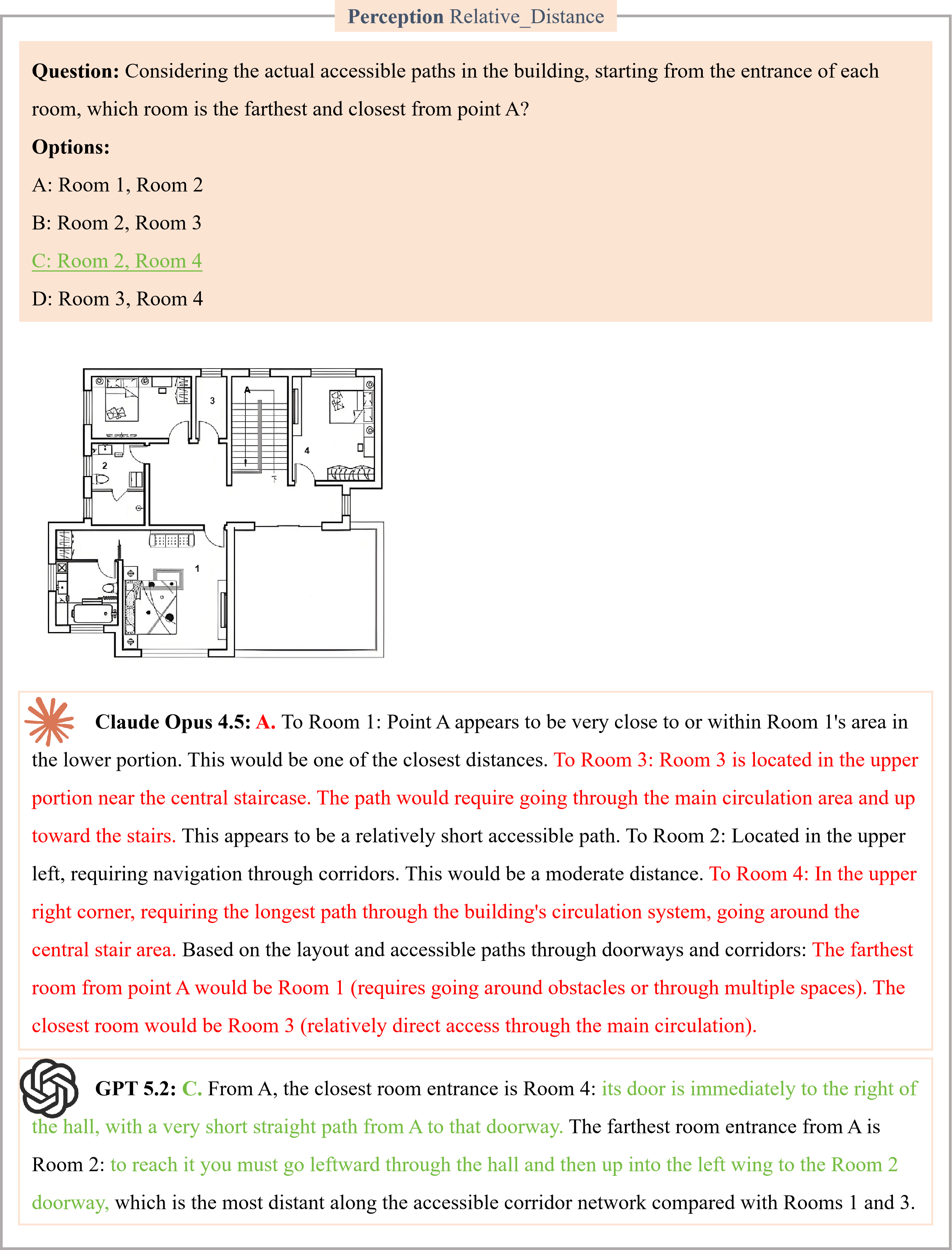}%
  {Case 2}%
  {Architectural Element Understanding Error: Claude-Opus-4.5 failed to correctly interpret the traversal direction of the staircase in the image, leading to an incorrect inference regarding the access routes to Rooms 3 and 4.}%
  {fig:case_study_2}

\casestudyfigure{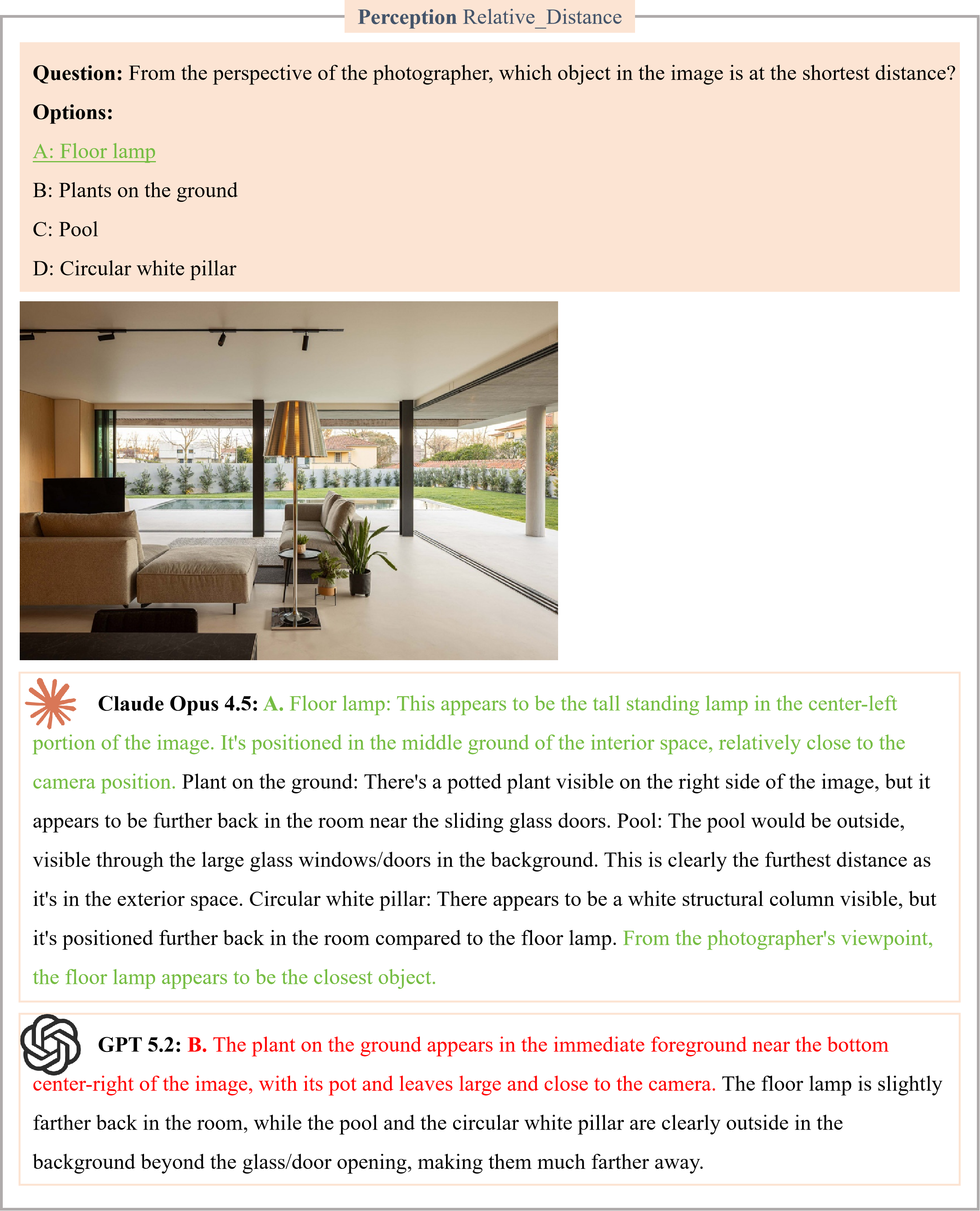}%
  {Case 3}%
  {Visual Perception Error: GPT-5.2 failed to correctly perceive foreground-background spatial relationships, leading to an incorrect judgment of the relative positions of the plant and the floor lamp.}%
  {fig:case_study_3}

\casestudyfigure{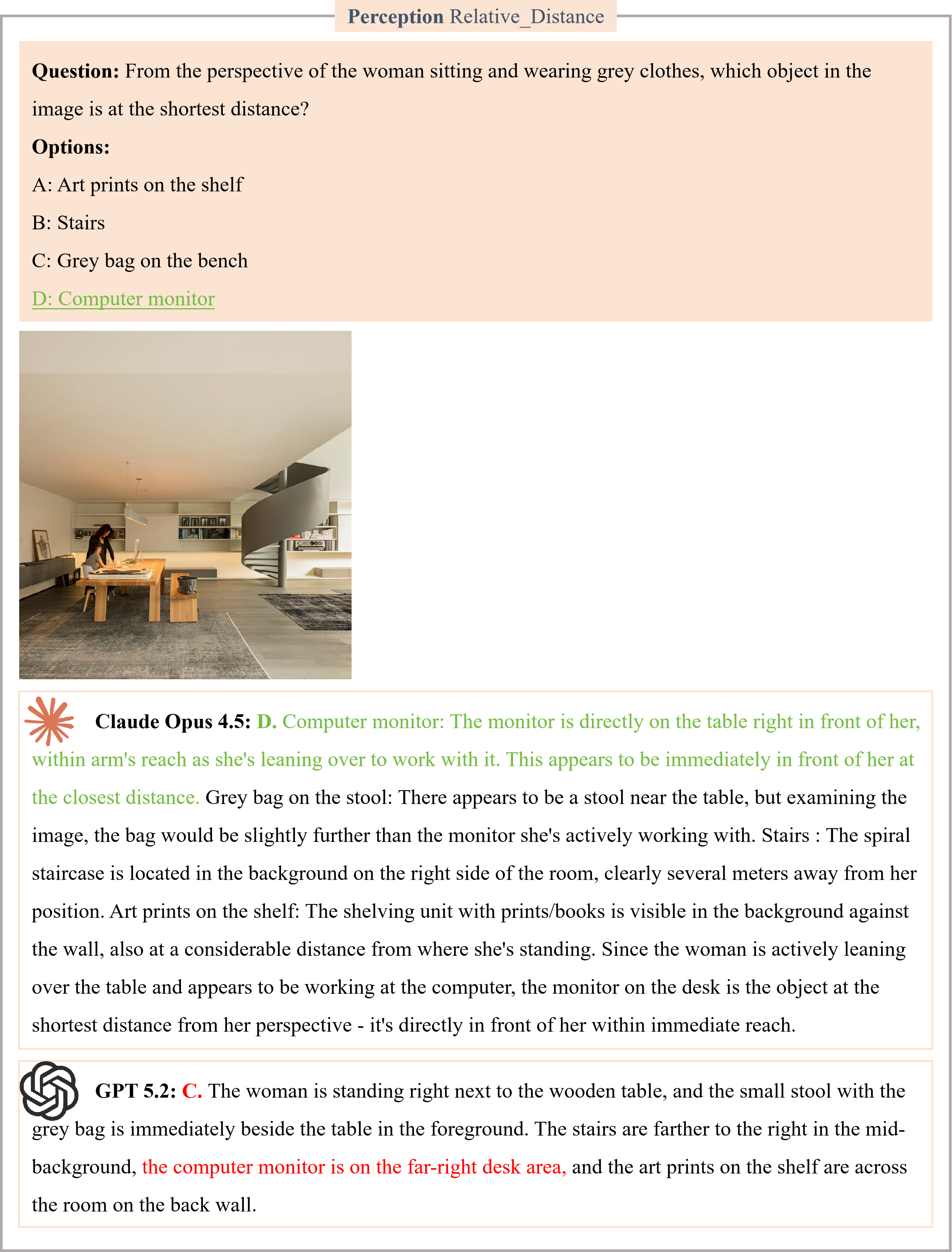}%
  {Case 4}%
  {Visual Perception Error: GPT-5.2 failed to correctly identify the target person specified in the question and made an inaccurate judgment regarding the precise location of the computer monitor.}%
  {fig:case_study_4}

\casestudyfigure{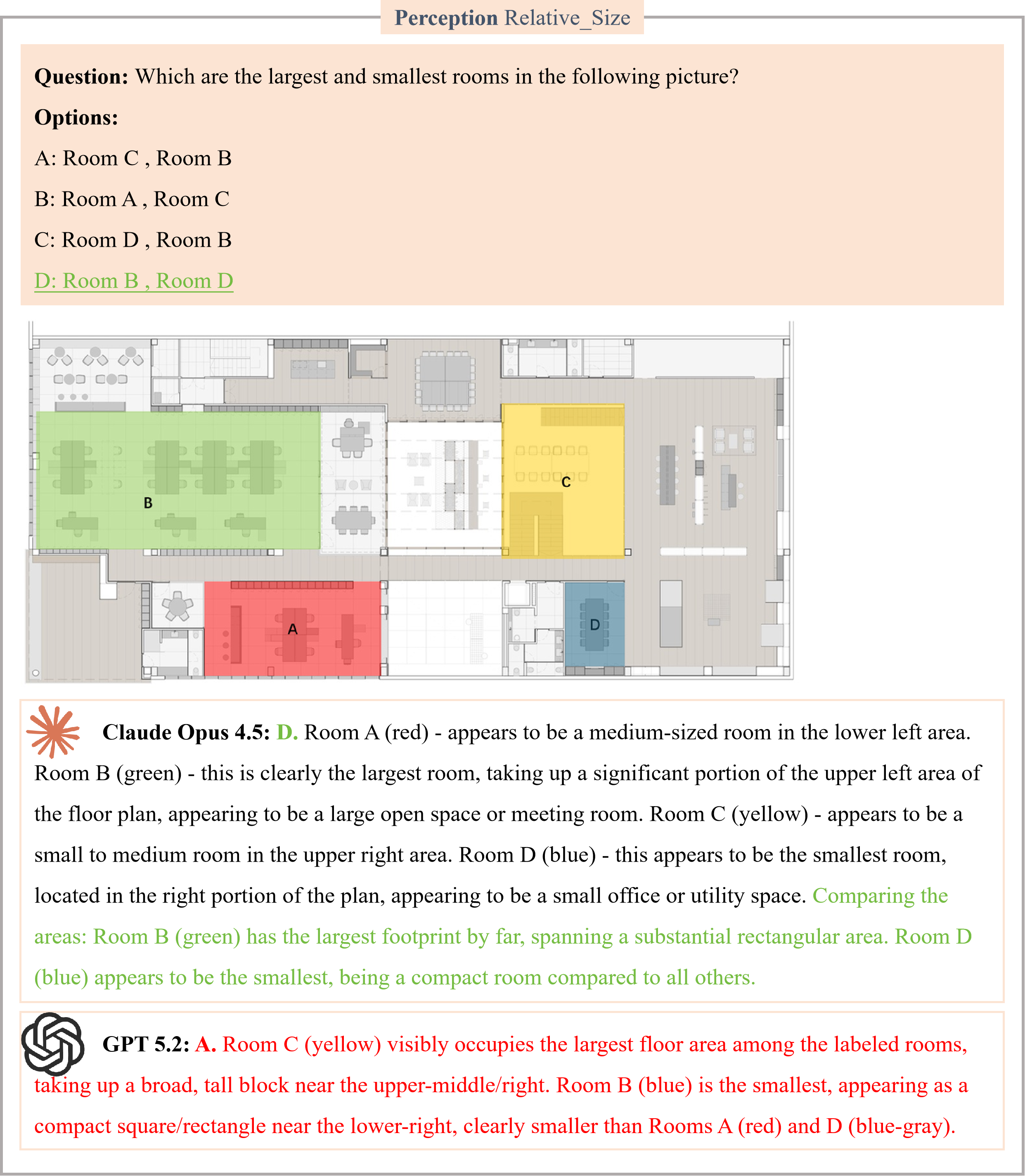}%
  {Case 5}%
  {Visual Perception Error and Relational Reasoning Error: GPT-5.2 failed to correctly recognize or distinguish Room B (which should be green rather than blue), and incorrectly identified Room C as the largest room.}%
  {fig:case_study_5}

\casestudyfigure{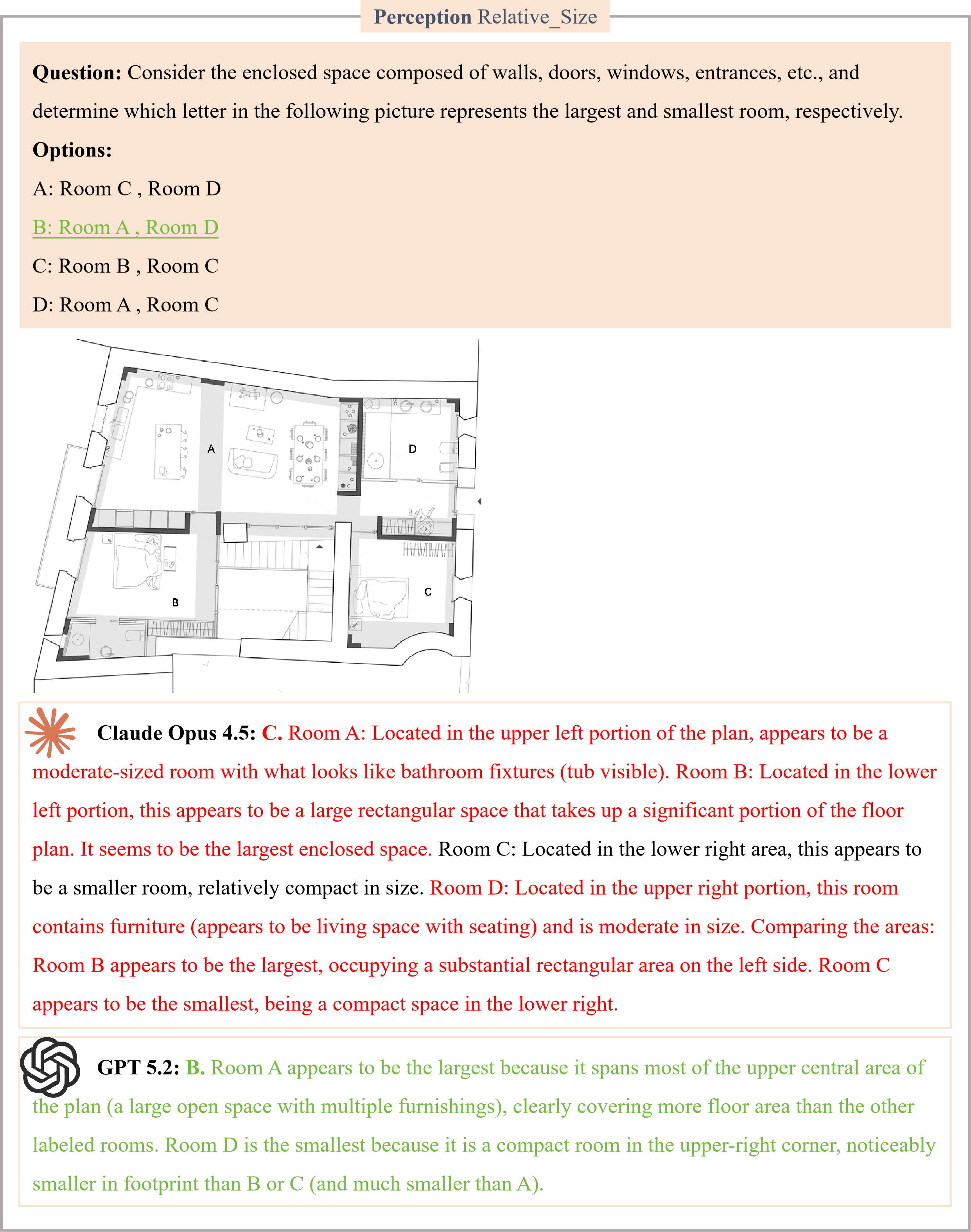}%
  {Case 6}%
  {Architectural Element Understanding Error and Relational Reasoning Error: Claude-Opus-4.5 incorrectly identified the furniture types and room attributes in Rooms A and D, and failed to accurately capture the containment and enclosure relationships between rooms, leading to incorrect judgments of their relative sizes.}%
  {fig:case_study_6}

\casestudyfigure{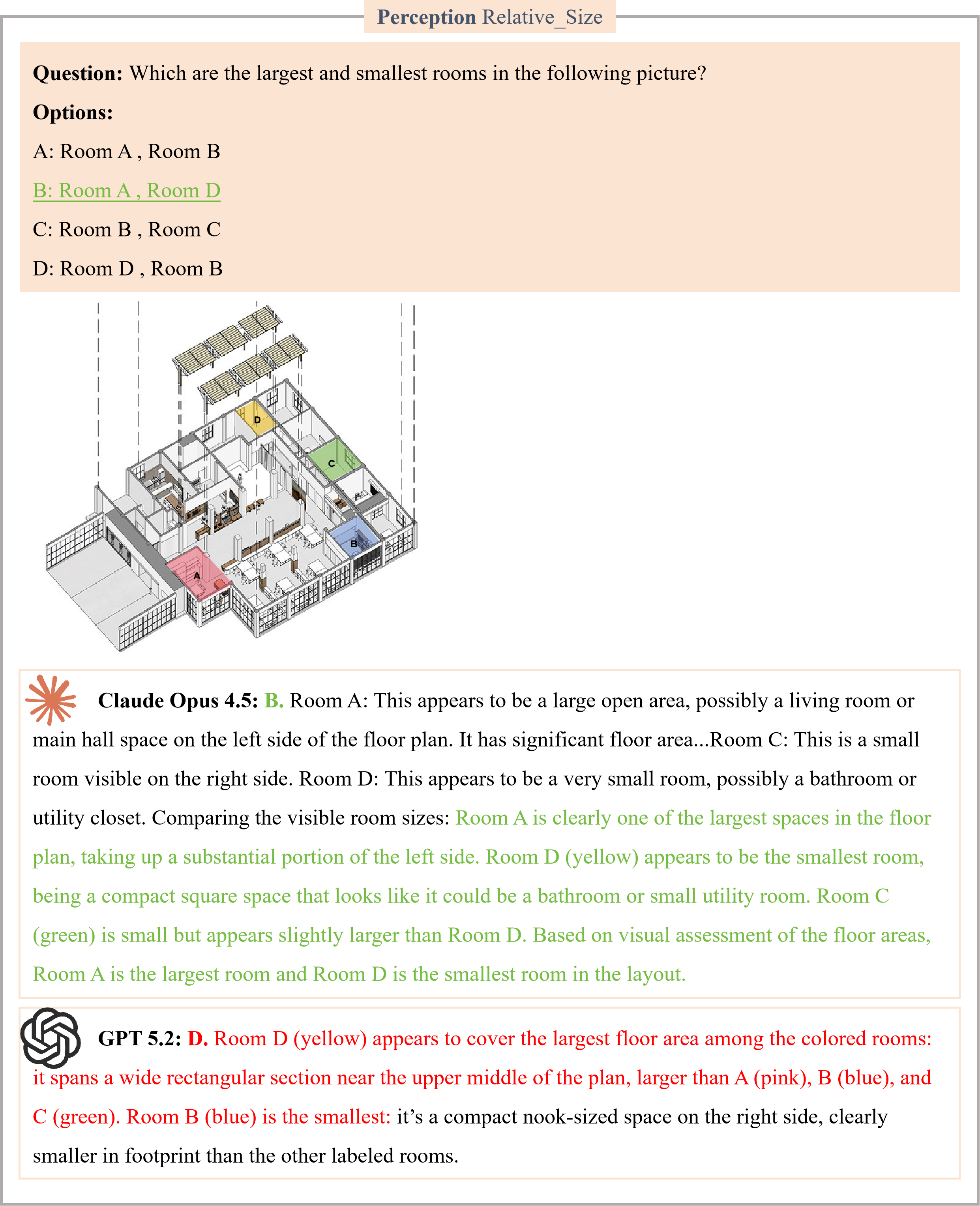}%
  {Case 7}%
  {Relational Reasoning Error: GPT-5.2 incorrectly inferred the relative size relationships among the rooms.}%
  {fig:case_study_7}

\casestudyfigure{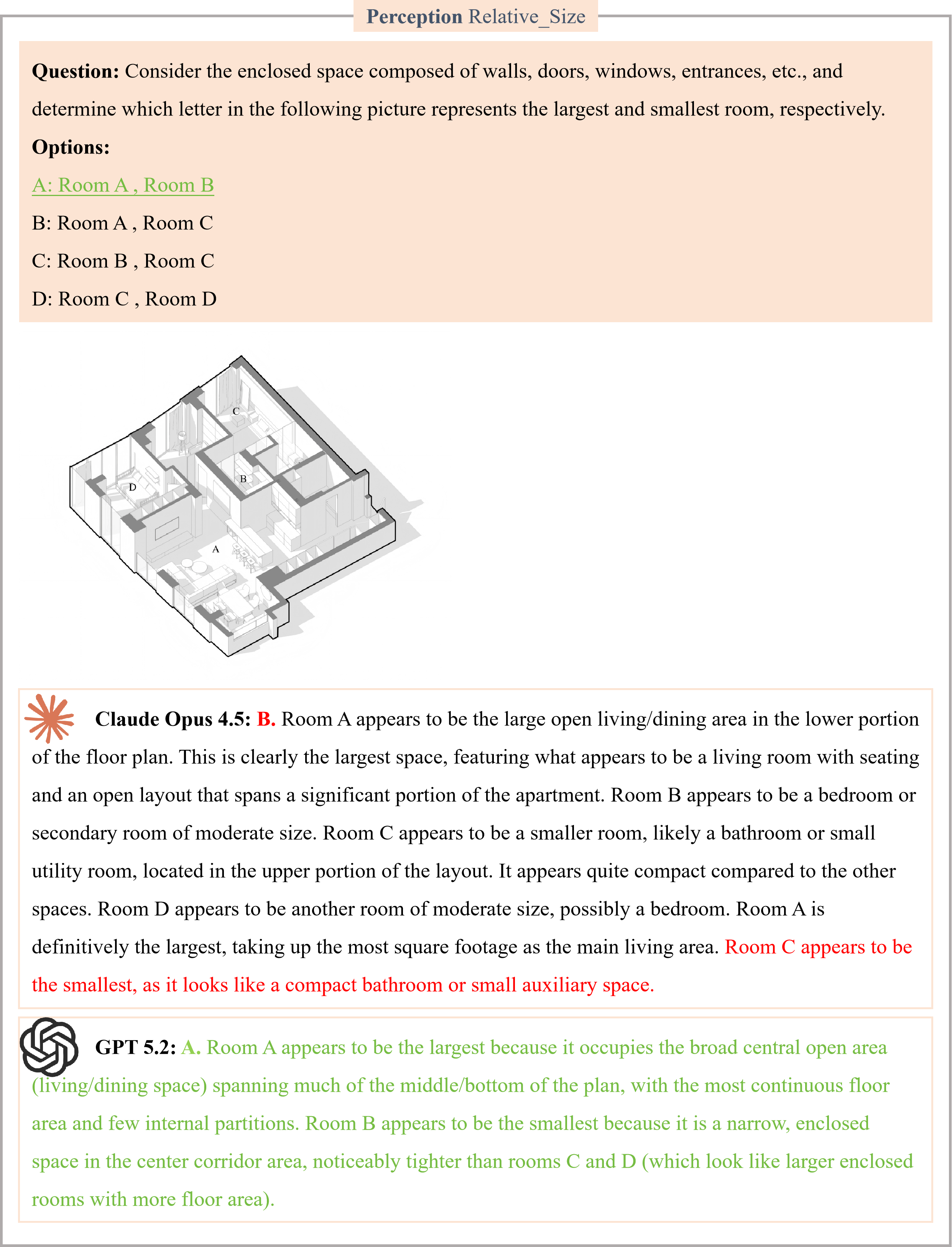}%
  {Case 8}%
  {Architectural Element Understanding Error and Relational Reasoning Error: Claude-Opus-4.5 incorrectly identified the furniture types and room attributes in Rooms B and C, and also misjudged the relative size relationships among the rooms.}%
  {fig:case_study_8}

\casestudyfigure{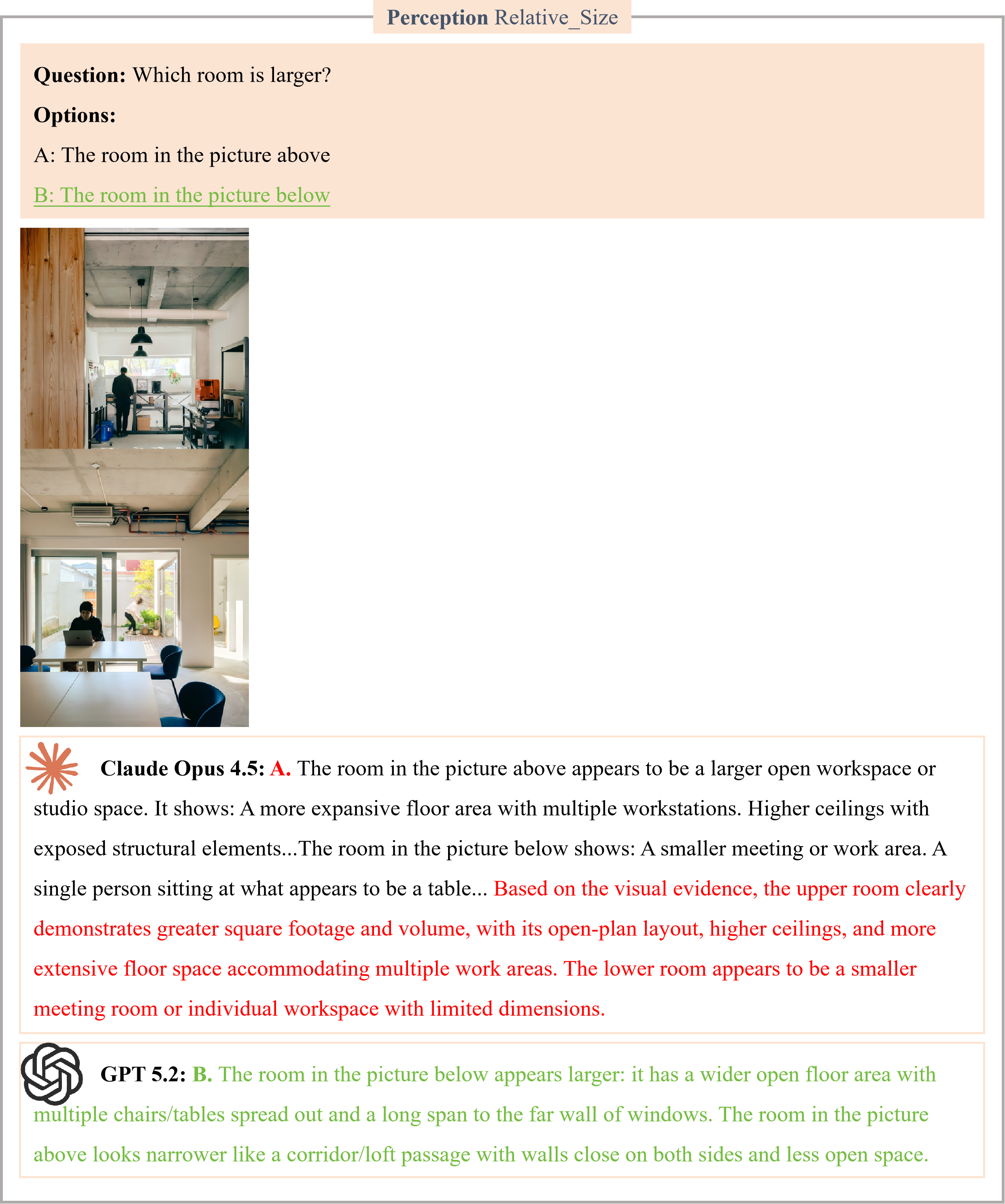}%
  {Case 9}%
  {Visual Perception Error and Relational Reasoning Error: Claude-Opus-4.5 failed to compare the actual sizes of rooms across two independent images based solely on visual evidence.}%
  {fig:case_study_9}

\casestudyfigure{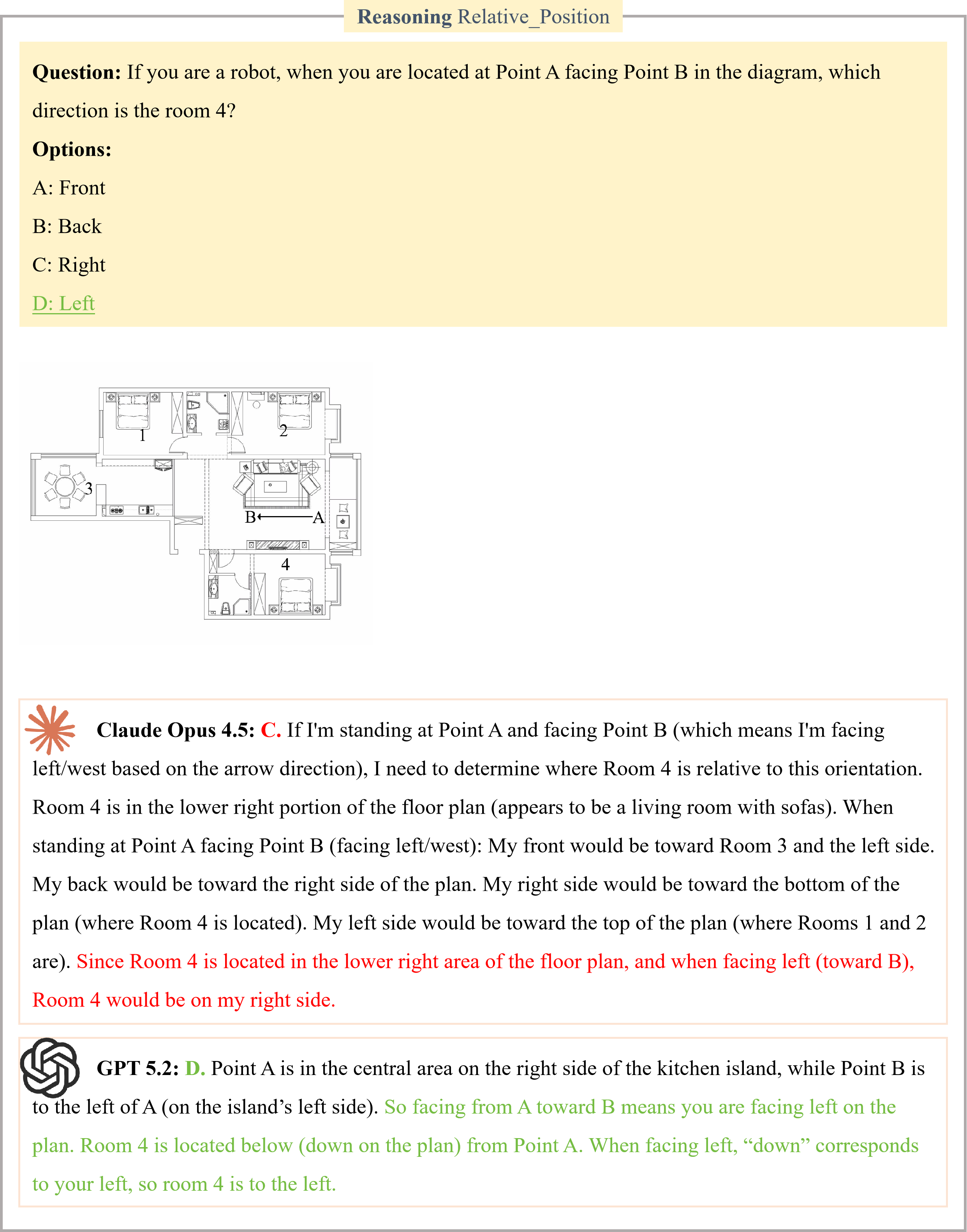}%
  {Case 10}%
  {Viewpoint Transformation Error and Relational Reasoning Error: Claude-Opus-4.5 decomposed the mental rotation process into intermediate steps but still produced an incorrect result, failing to correctly establish a local coordinate system and consequently reversing left-right judgments.}%
  {fig:case_study_10}

\casestudyfigure{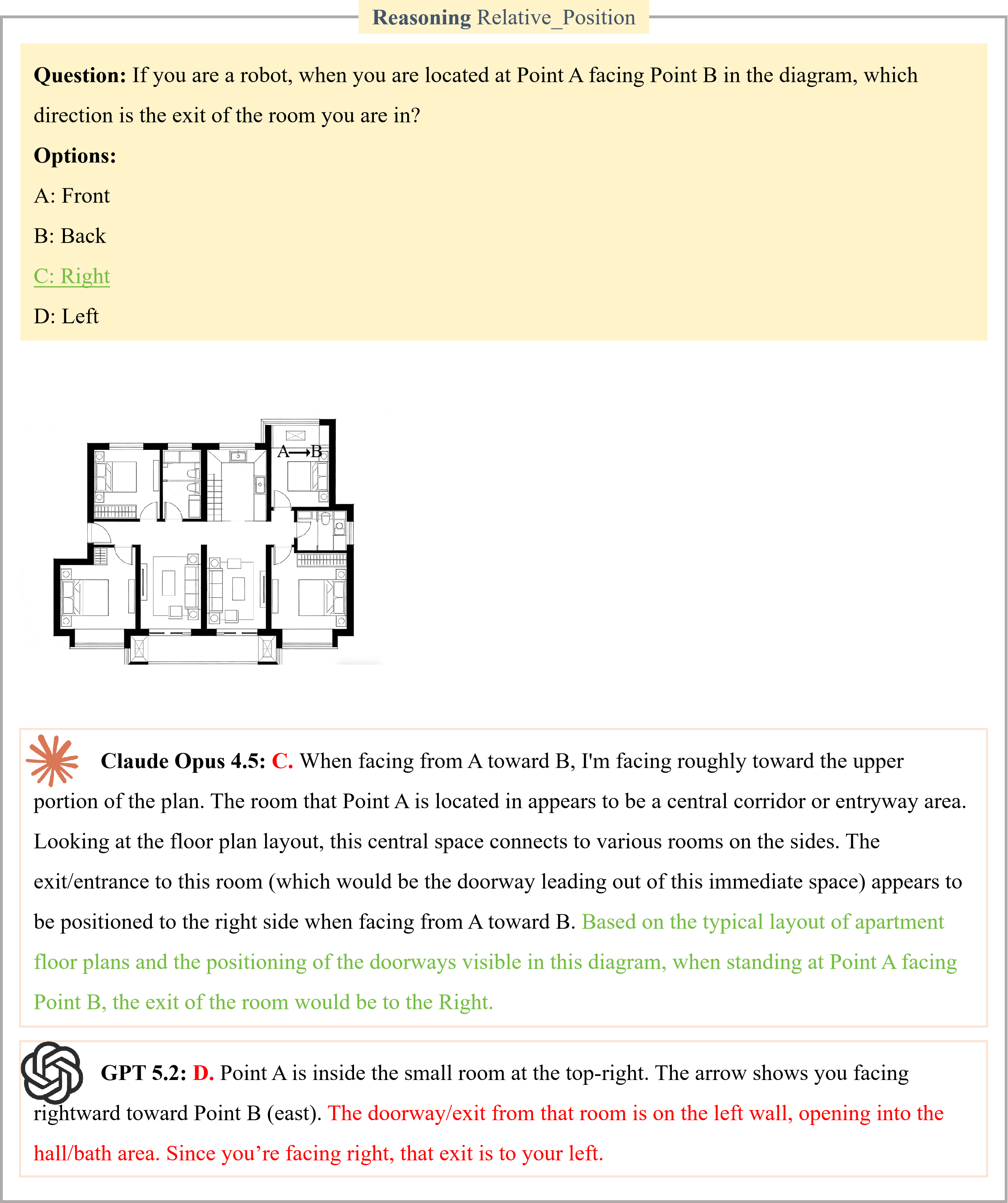}%
  {Case 11}%
  {Viewpoint Transformation Error and Relational Reasoning Error: GPT-5.2 decomposed the mental rotation process into intermediate steps but still produced an incorrect result, failing to correctly establish a local coordinate system and consequently reversing left-right judgments.}%
  {fig:case_study_11}

\casestudyfigure{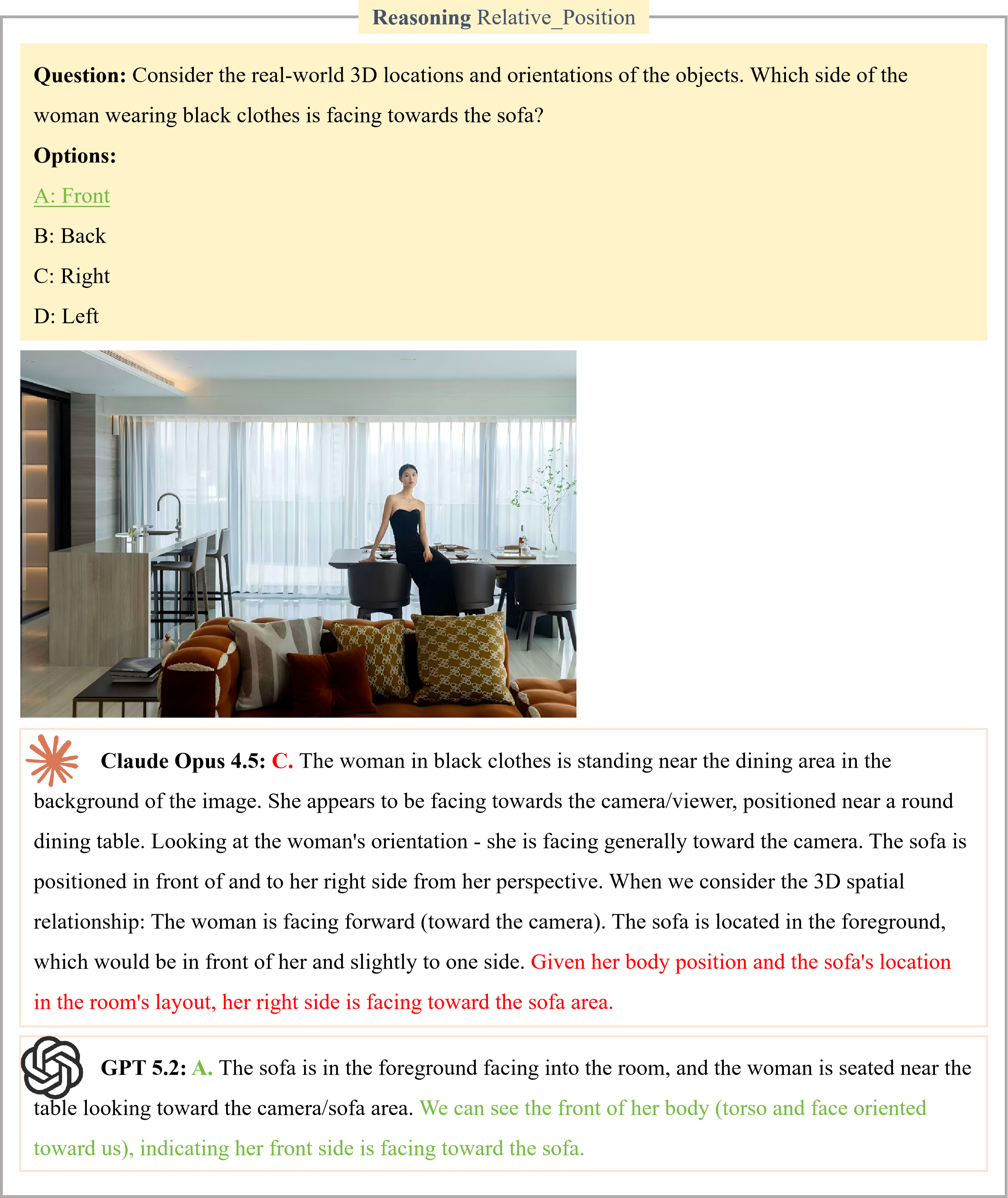}%
  {Case 12}%
  {Visual Perception Error: Claude-Opus-4.5 overemphasized minor variations in the depicted human pose, leading to an incorrect inference of the overall body orientation.}%
  {fig:case_study_12}

\casestudyfigure{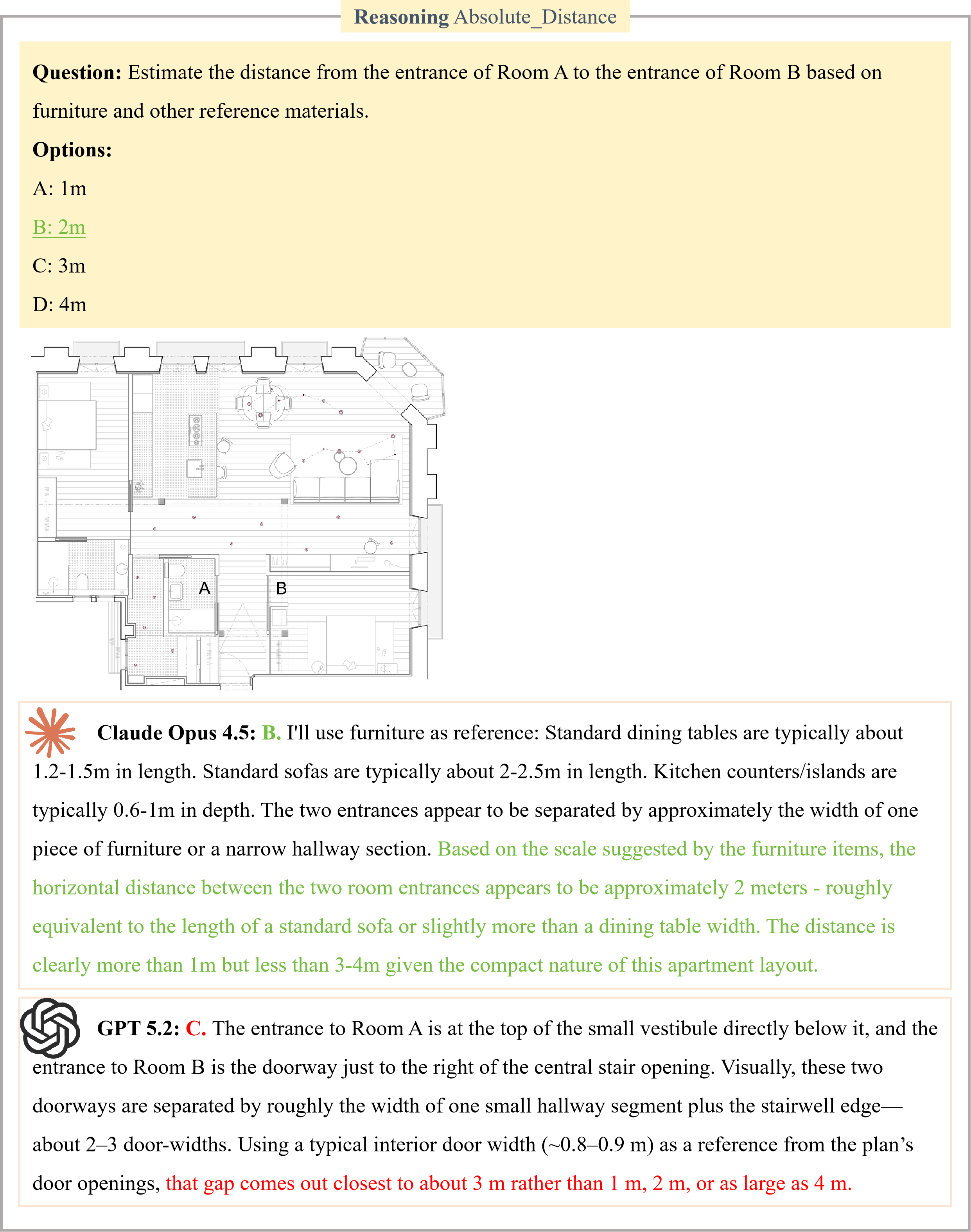}%
  {Case 13}%
  {Relational Reasoning Error: GPT-5.2 appropriately leveraged doorway dimensions as a reference scale, but introduced errors when applying this reference to precise distance counting and aggregation, resulting in incorrect distance estimation.}%
  {fig:case_study_13}

\casestudyfigure{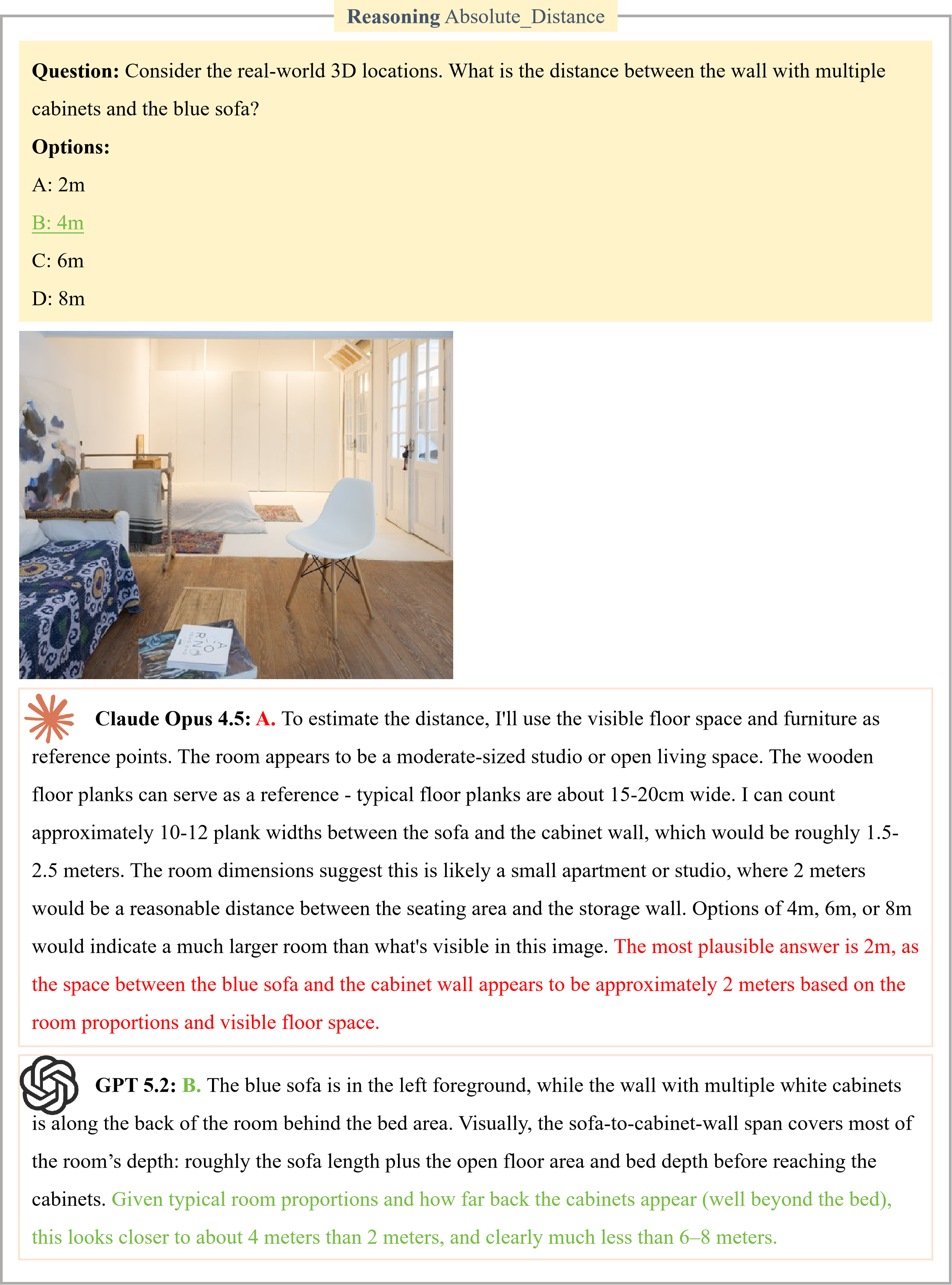}%
  {Case 14}%
  {Relational Reasoning Error: Claude-Opus-4.5 selected small-scale objects such as wooden floor planks as reference units instead of larger, more reliable anchors like beds or sofas, leading to significant deviations in spatial distance estimation.}%
  {fig:case_study_14}

\casestudyfigure{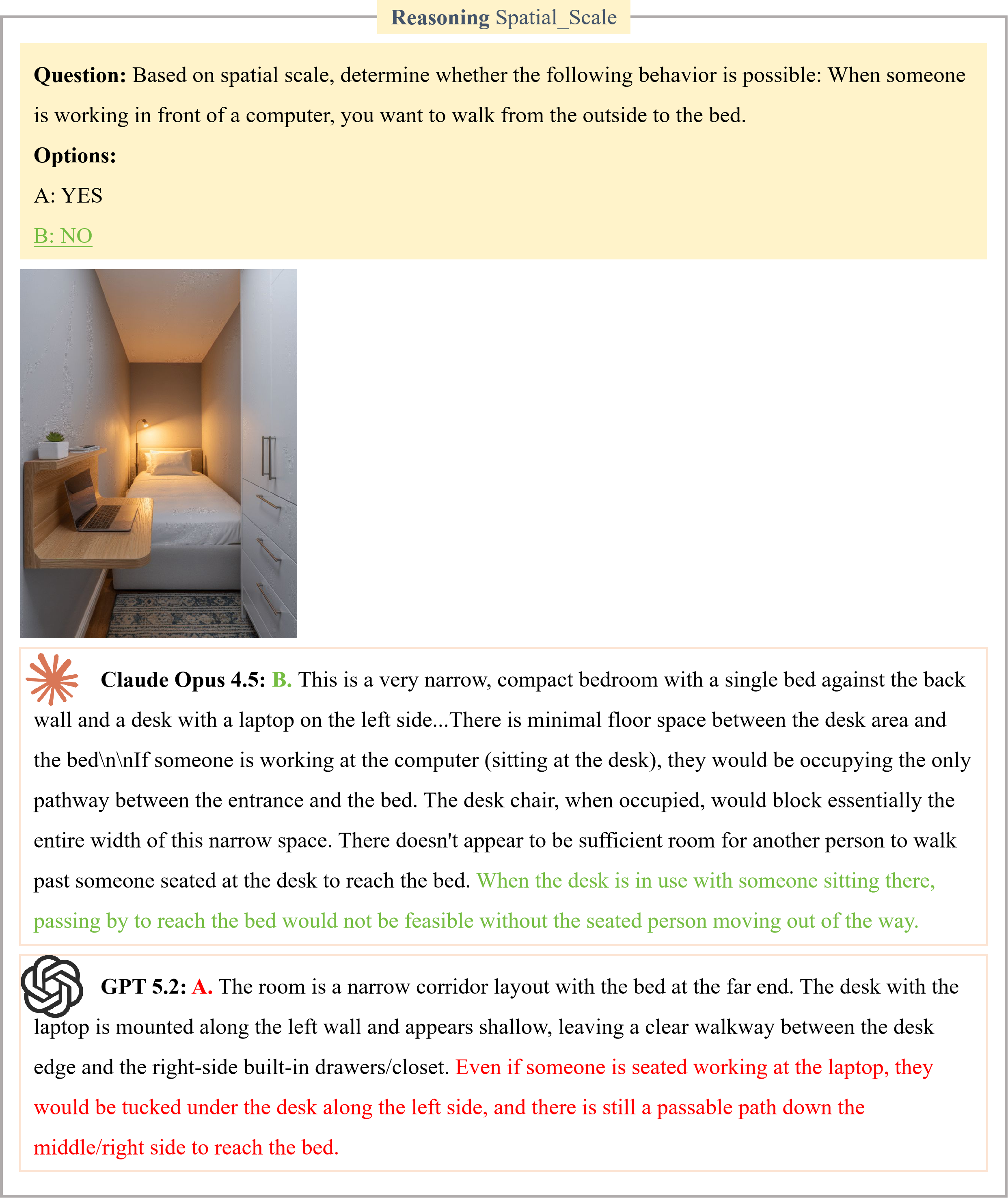}%
  {Case 15}%
  {Embodied Scale Reasoning Error and Logical Reasoning Error: GPT-5.2 failed to jointly reason over furniture dimensions, human scale, and passage width when determining navigability in narrow spaces, resulting in an incorrect judgment of passability. Additionally, ``be tucked under the desk'' is not a physically plausible or behaviorally valid form of traversal.}%
  {fig:case_study_15}

\casestudyfigure{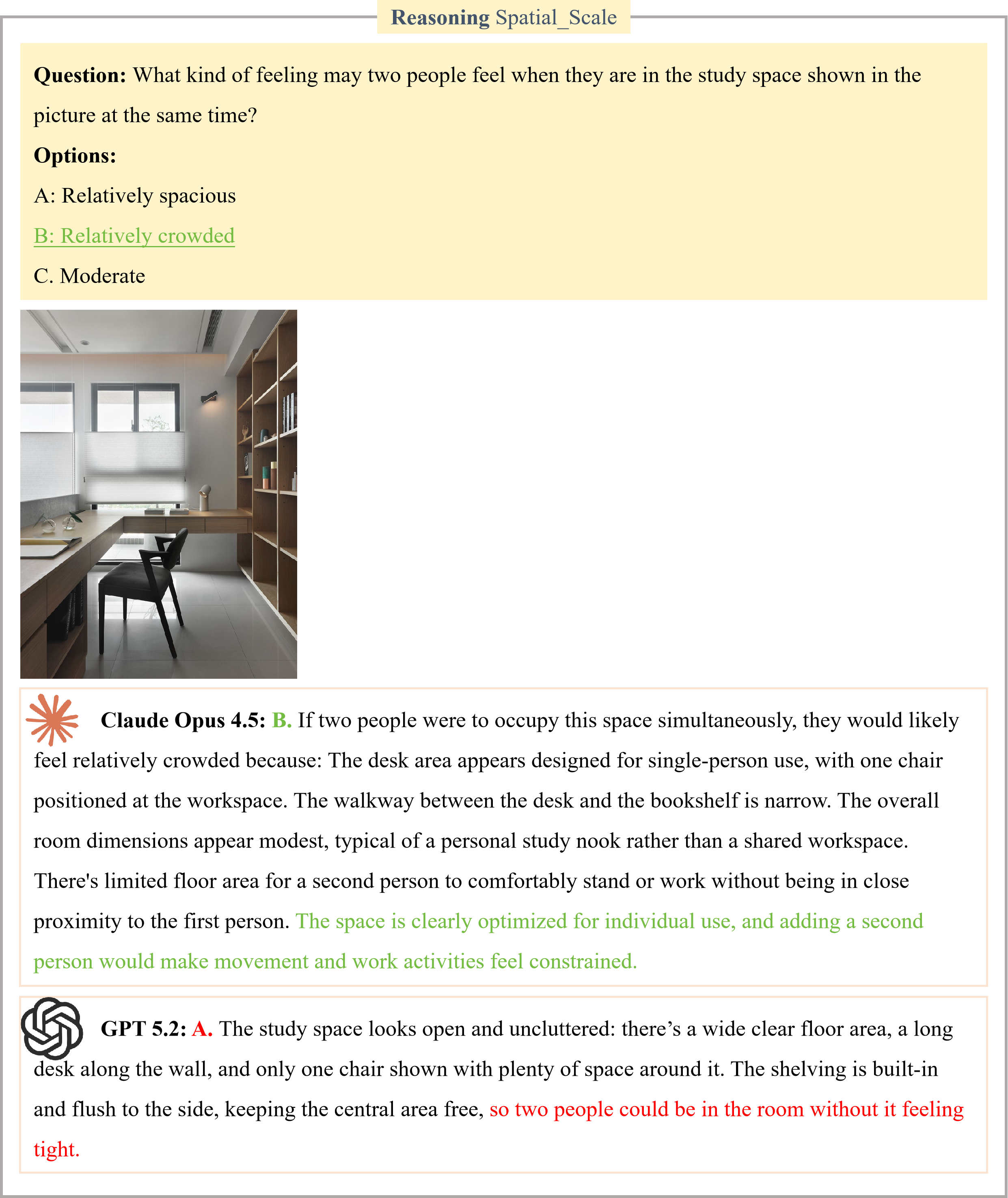}%
  {Case 16}%
  {Embodied Scale Reasoning Error and Logical Reasoning Error: GPT-5.2 failed to jointly reason over per-capita activity space, furniture density, and spatial openness when estimating the ideal occupancy of the space.}%
  {fig:case_study_16}

\casestudyfigure{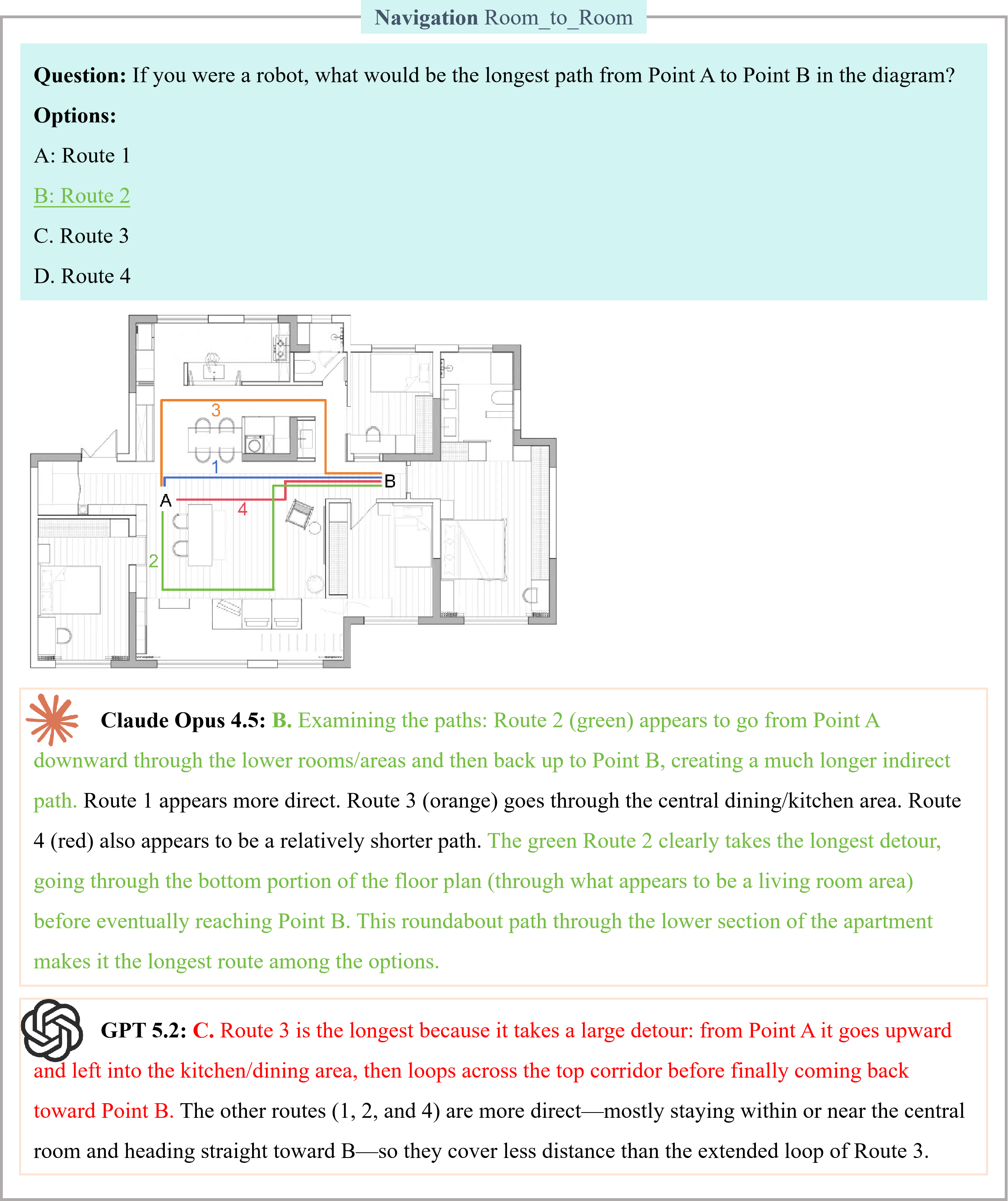}%
  {Case 17}%
  {Relational Reasoning Error: GPT-5.2 failed to correctly aggregate and compare the actual traversable path lengths from point A to point B.}%
  {fig:case_study_17}

\casestudyfigure{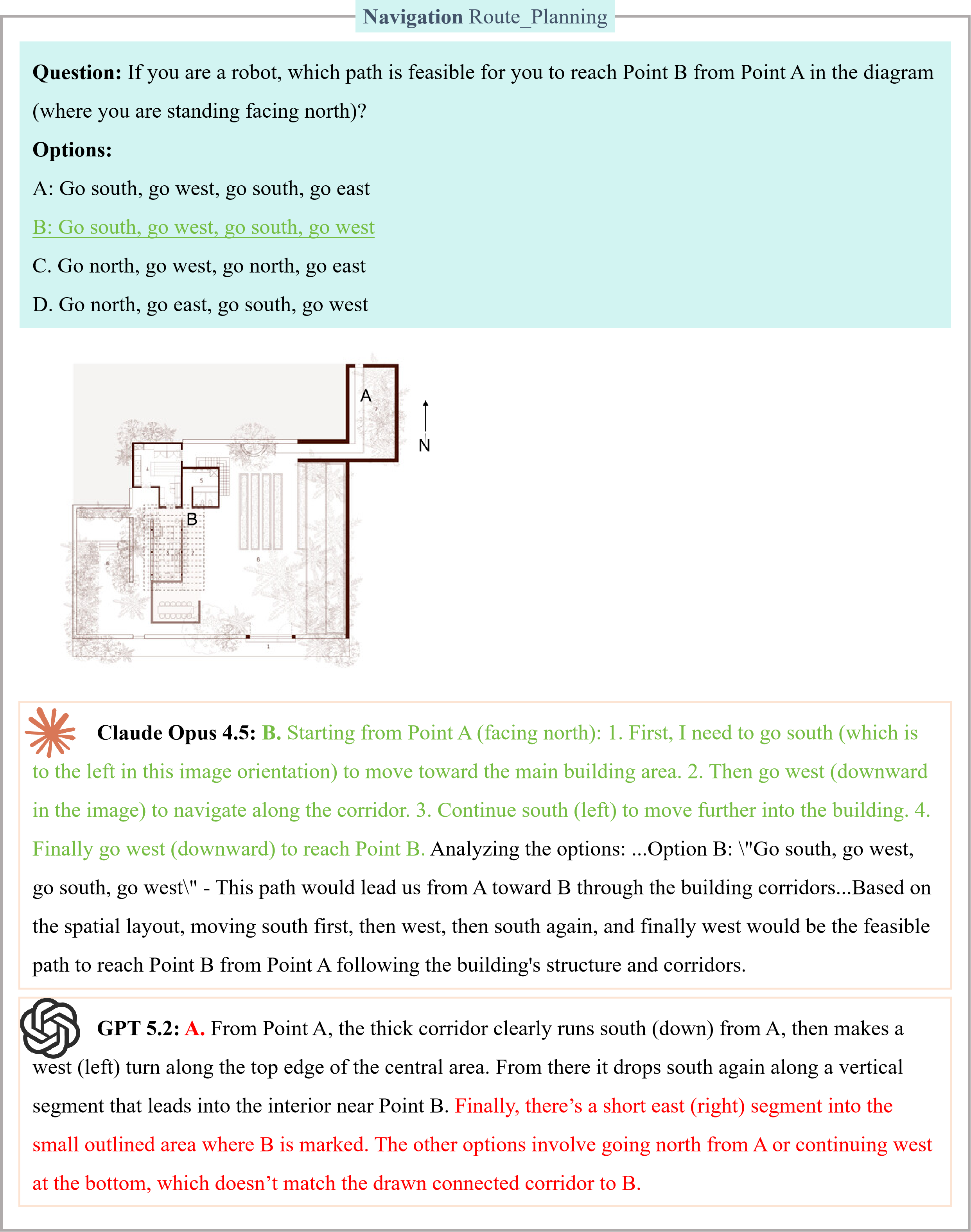}%
  {Case 18}%
  {Relational Reasoning Error and Semantic Perception Error: GPT-5.2 failed to correctly establish a local coordinate system in the final segment of the path, resulting in reversed left-right (or east-west) orientation judgments. In addition, the model did not accurately interpret the semantics of alternative paths.}%
  {fig:case_study_18}

\casestudyfigure{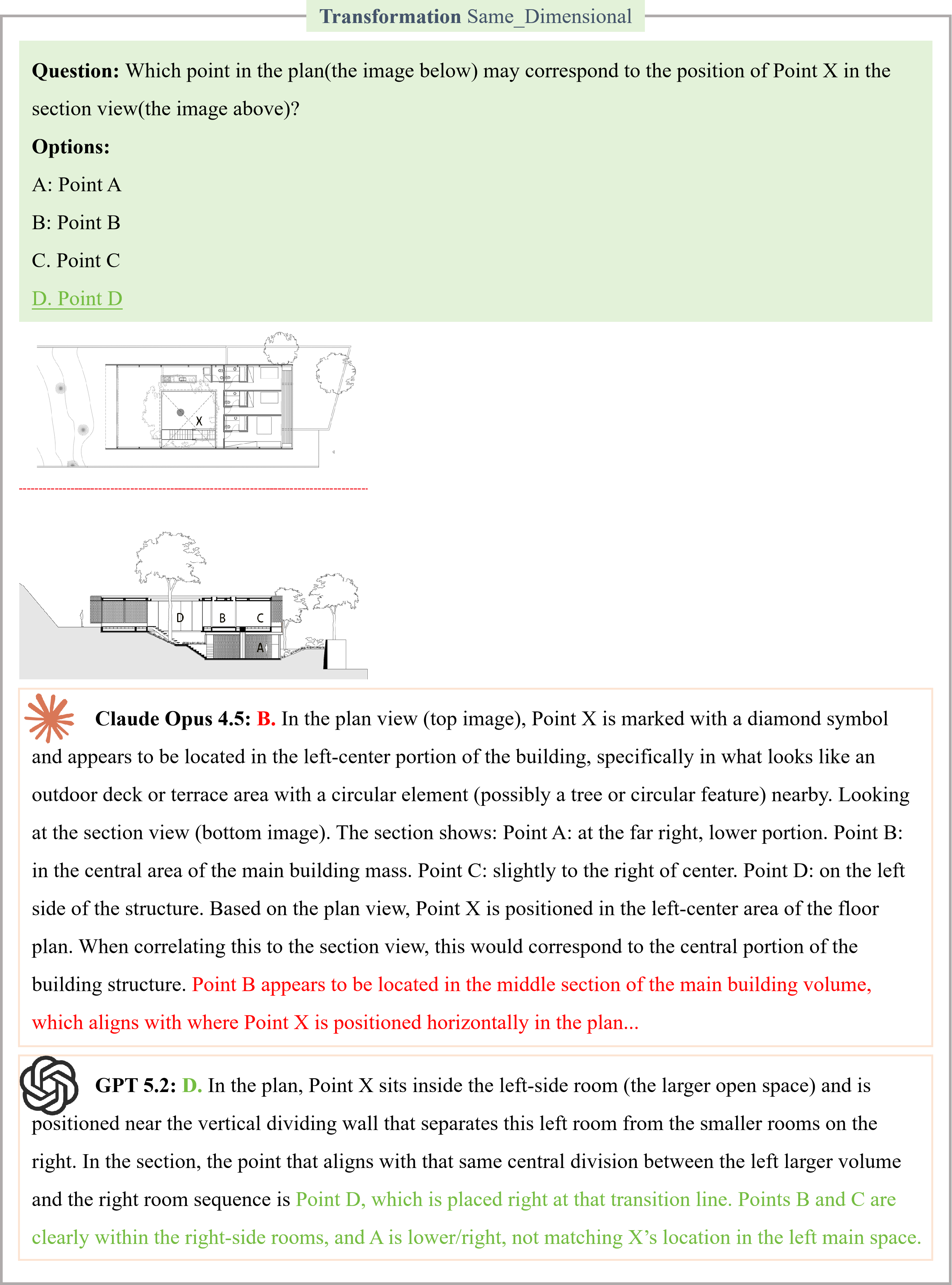}%
  {Case 19}%
  {Viewpoint Transformation Error and Relational Reasoning Error: Claude-Opus-4.5 failed to correctly align point X between the plan and the section view; even when attempting alignment via partition lines, it could not establish consistent cross-view positional correspondence between the two 2D representations.}%
  {fig:case_study_19}

\casestudyfigure{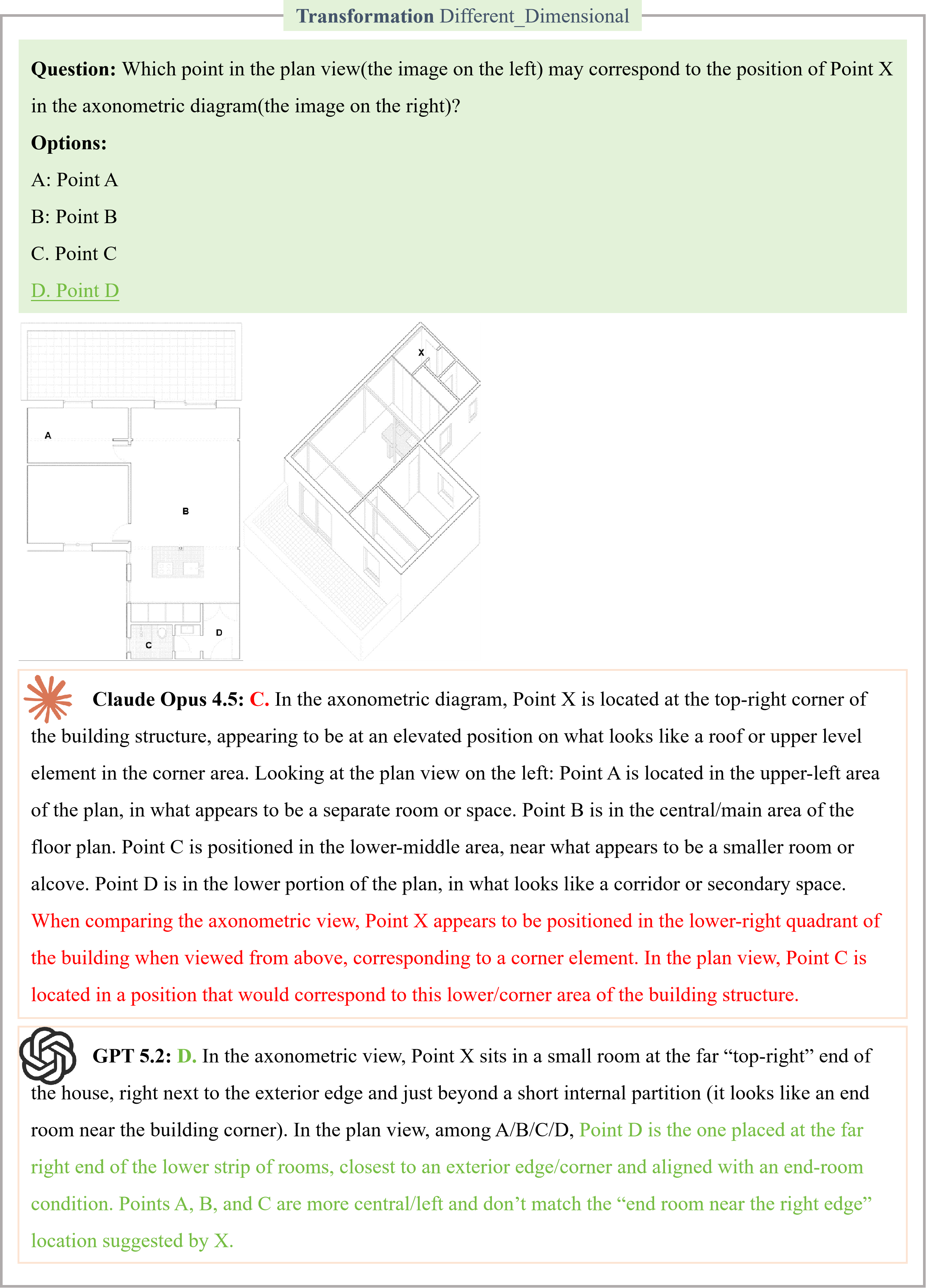}%
  {Case 20}%
  {Viewpoint Transformation Error and Relational Reasoning Error: Claude-Opus-4.5 correctly inferred the approximate location of point X when mapping from the axonometric diagram to the plan, but failed to accurately determine its precise position.}%
  {fig:case_study_20}

\casestudyfigure{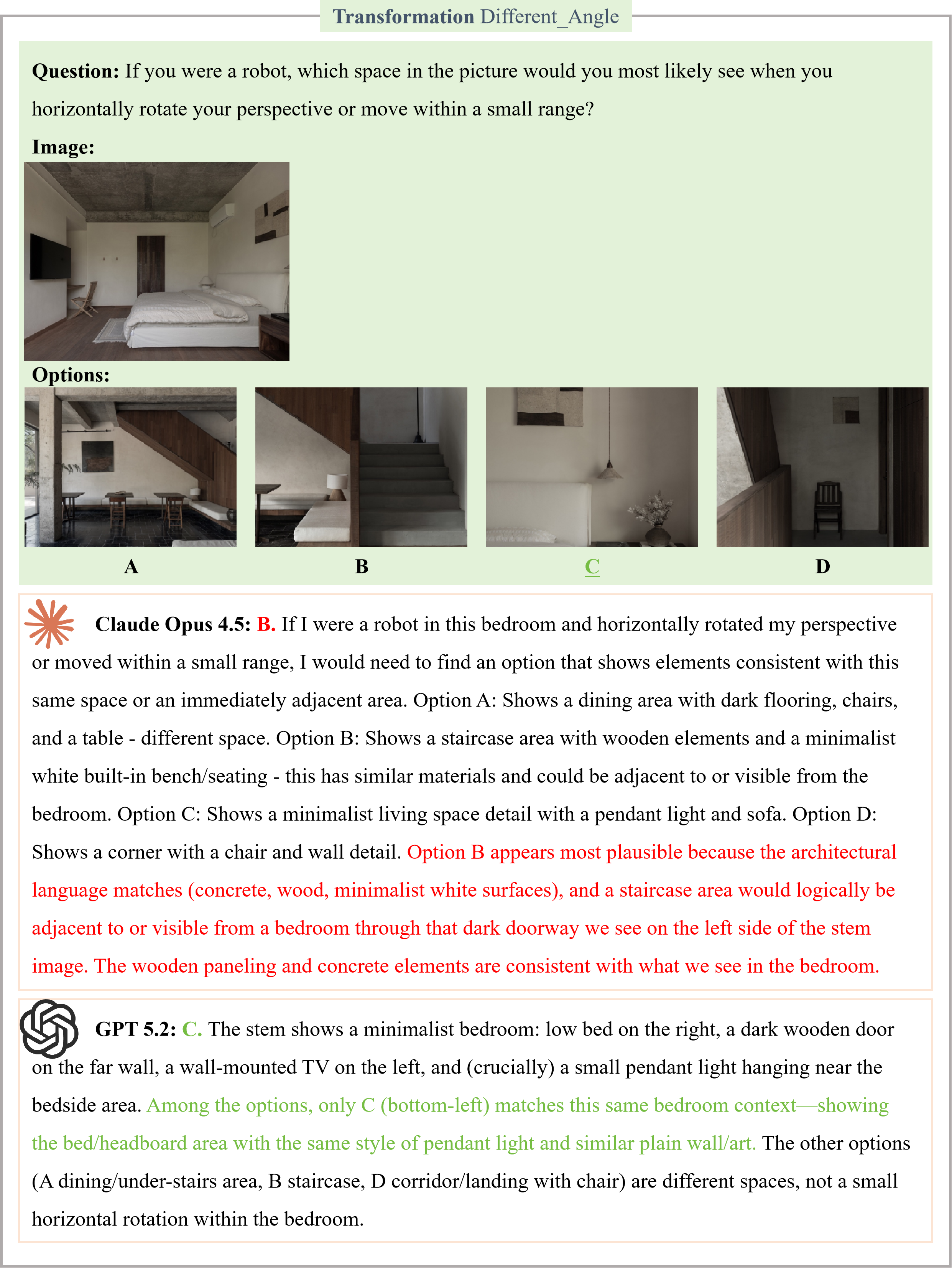}%
  {Case 21}%
  {Viewpoint Transformation Error: Claude-Opus-4.5 incorrectly predicted visual continuity of adjacent spaces when simulating small-scale in-room rotations or translations, and consequently selected alternative spatial representations that were not semantically the most consistent.}%
  {fig:case_study_21}

\casestudyfigure{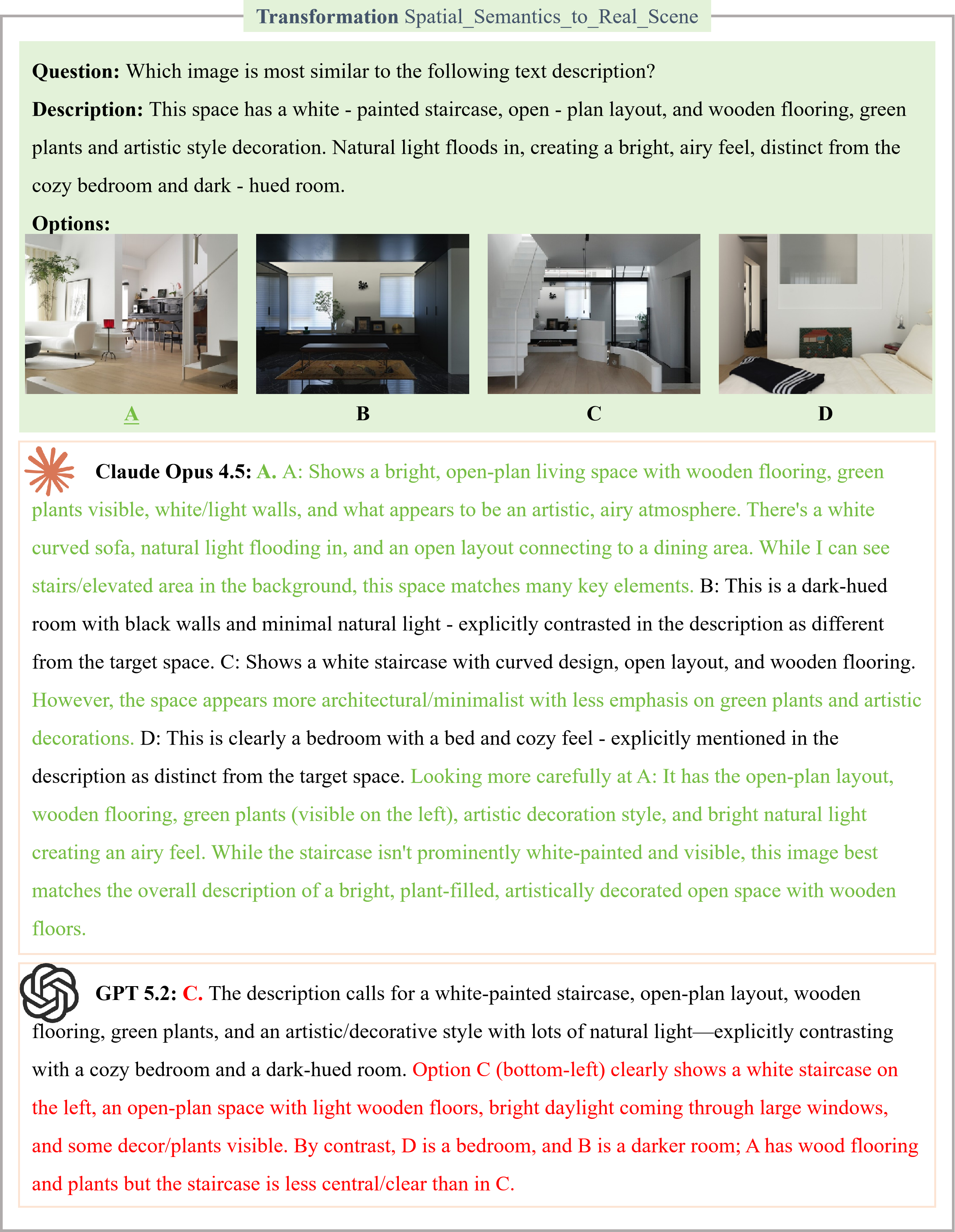}%
  {Case 22}%
  {Semantic Perception Error: GPT-5.2 incorrectly aligned spatial semantics from the textual description (e.g., white staircase, open-plan layout) with candidate images, misallocating attention to key semantic features and being misled by secondary decorative elements.}%
  {fig:case_study_22}

\casestudyfigure{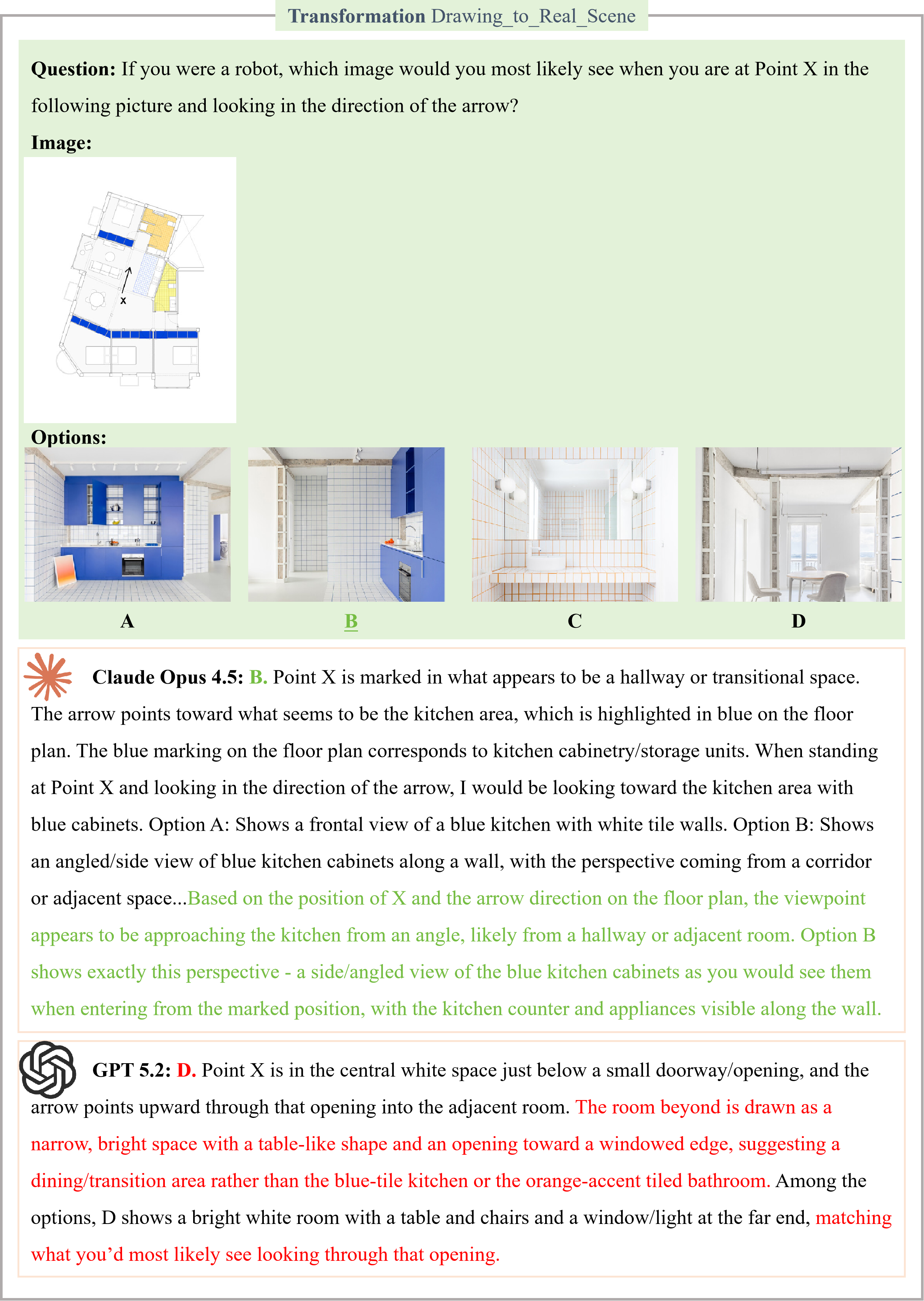}%
  {Case 23}%
  {Architectural Element Understanding Error: GPT-5.2 misinterpreted the spatial semantics indicated by the arrow on the floor plan, and consequently failed to correctly project viewpoint X and its directional cue onto real-scene images, resulting in an inaccurate prediction of the corresponding visual perspective.}%
  {fig:case_study_23}

\casestudyfigure{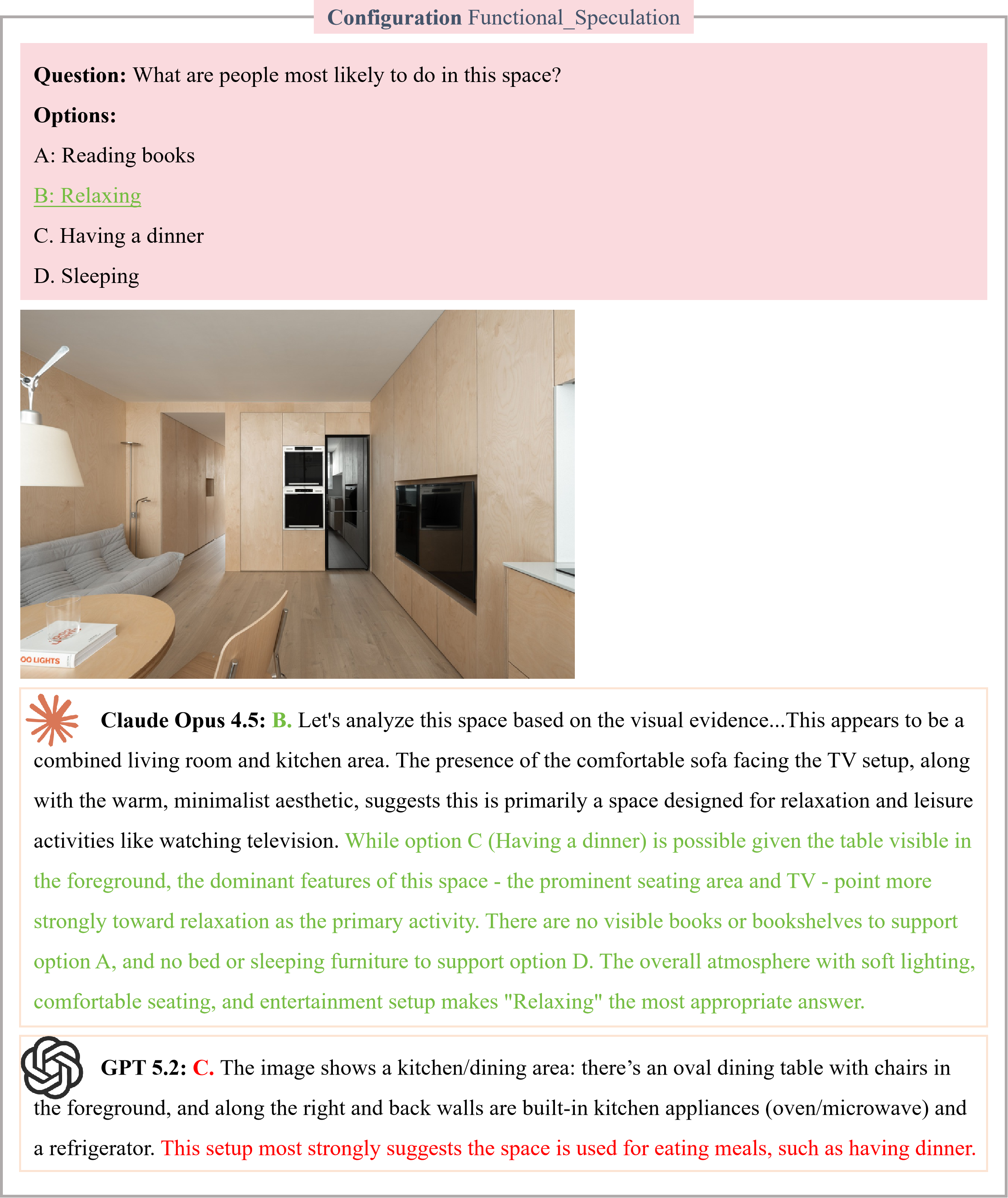}%
  {Case 24}%
  {Logical Reasoning Error: GPT-5.2 incorrectly inferred the primary functional use of the space, reflecting insufficient reasoning about the relationships among furniture arrangements, equipment configurations, and associated behavioral patterns.}%
  {fig:case_study_24}

\casestudyfigure{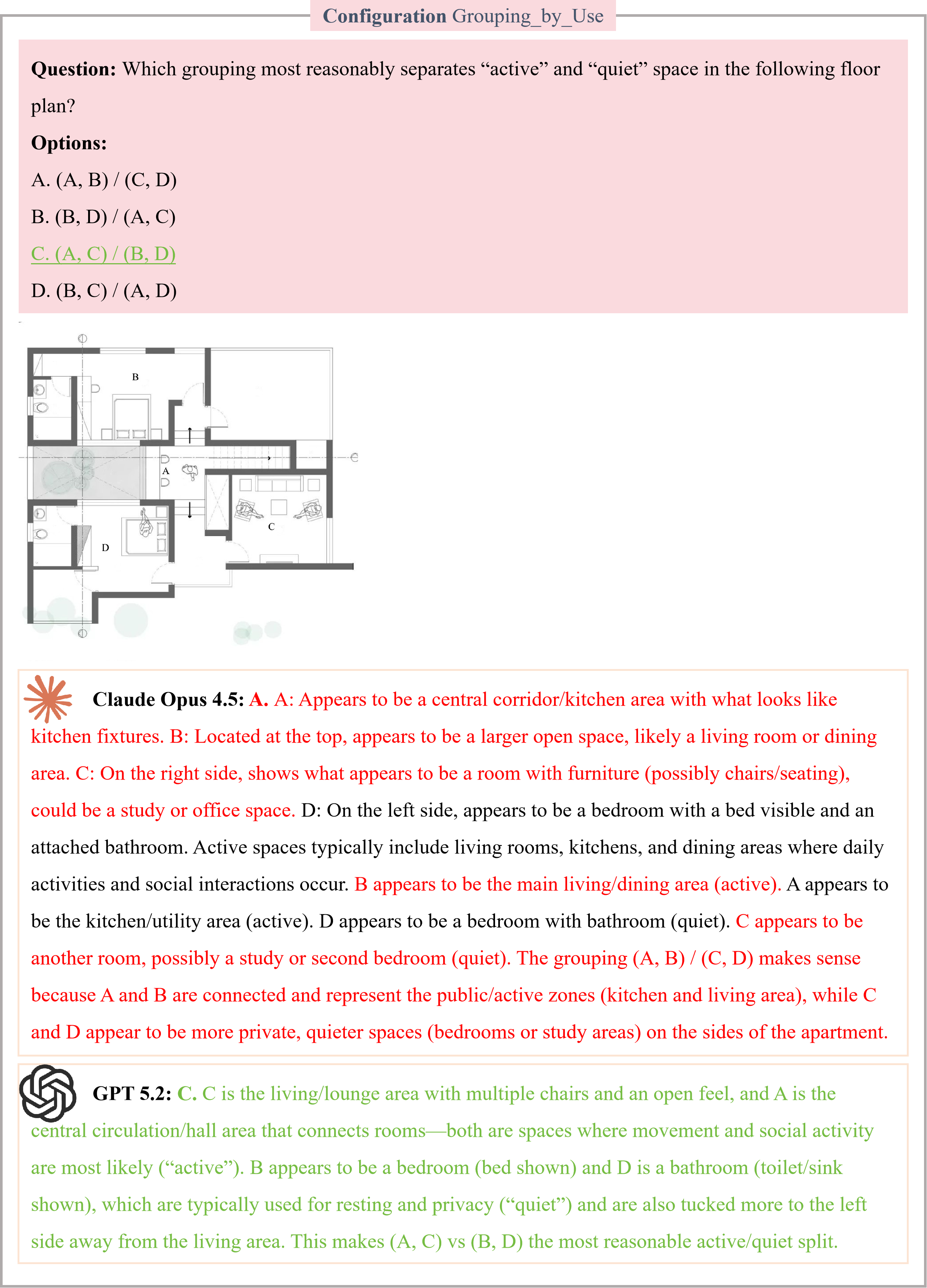}%
  {Case 25}%
  {Architectural Element Understanding Error: Claude-Opus-4.5 incorrectly identified the furniture types and room attributes in Rooms A, B, and C, leading to an erroneous inference of the functional zoning and grouping strategy.}%
  {fig:case_study_25}

\casestudyfigure{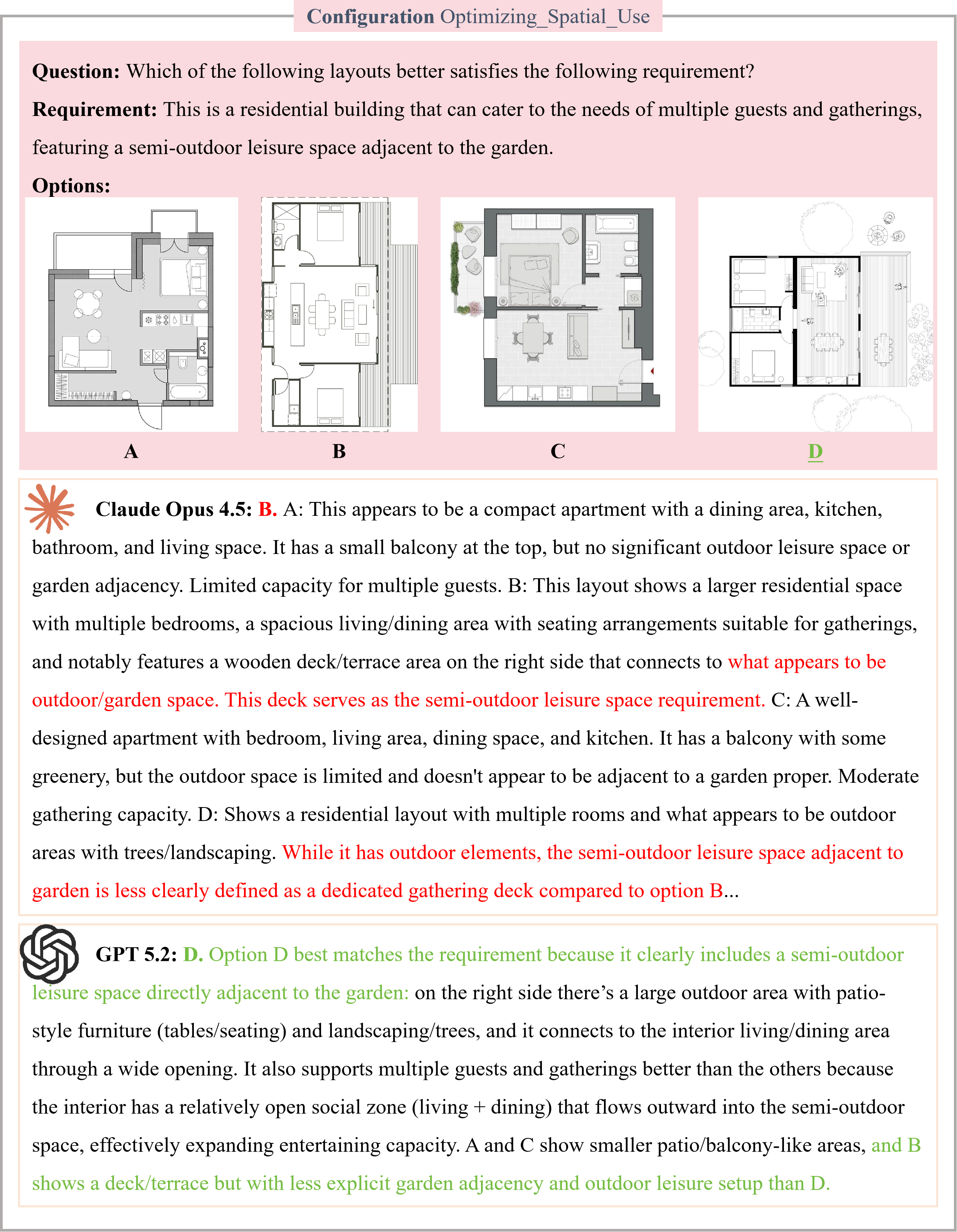}%
  {Case 26}%
  {Semantic Perception Error and Architectural Element Understanding Error: Claude-Opus-4.5 failed to correctly interpret the semantics of the ``a semi-outdoor leisure space adjacent to the garden'', misallocating weights to key semantic attributes and neglecting the ``adjacent to the garden'' constraint.}%
  {fig:case_study_26}

\casestudyfigure{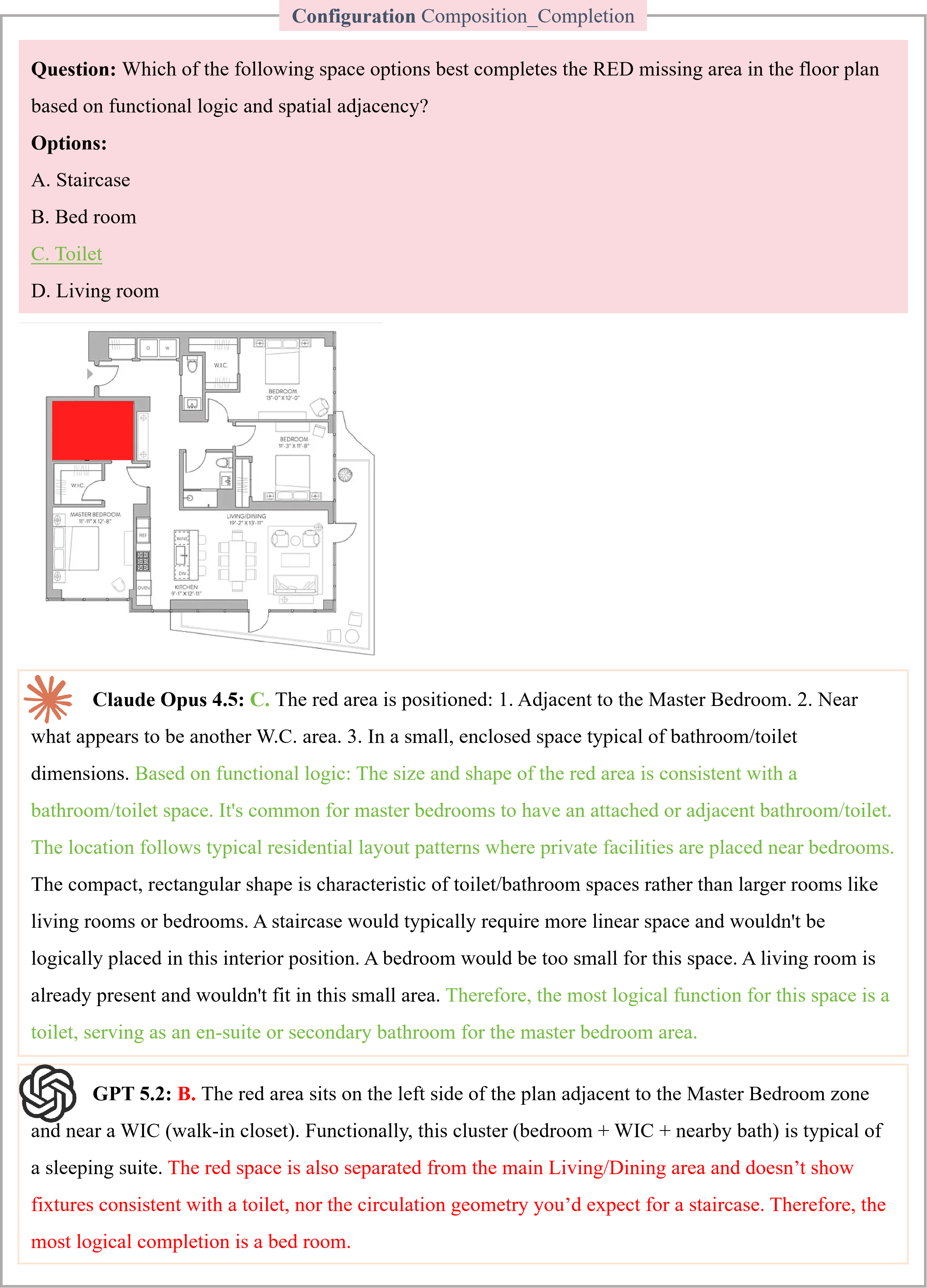}%
  {Case 27}%
  {Logical Reasoning Error: GPT-5.2 misjudged the logical relationships between adjacent spaces; in typical residential layouts, a master suite sequence such as bedroom-WIC-bathroom is a common configuration, whereas two bedrooms sharing a single access door and being treated as a unified circulation zone is atypical and functionally inconvenient.}%
  {fig:case_study_27}

\casestudyfigure{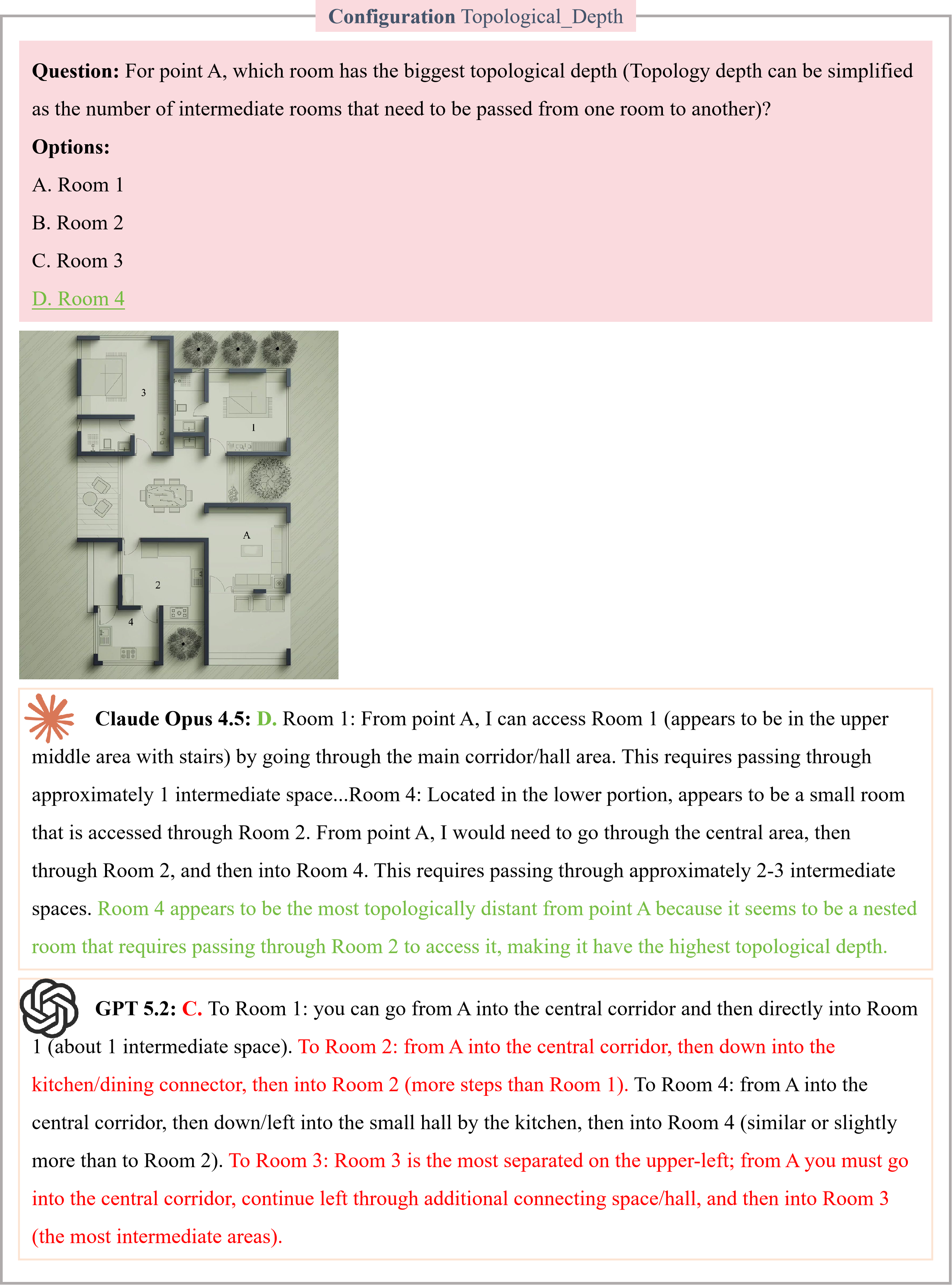}%
  {Case 28}%
  {Architectural Element Understanding Error and Relational Reasoning Error: GPT-5.2 exhibited inconsistent rules or failed to correctly identify transitional spaces (e.g., corridors and foyers) when computing topological depth, leading to erroneous judgments.}%
  {fig:case_study_28}


\end{document}